\documentclass{article}
\usepackage[utf8]{inputenc}

\usepackage{bbold}
\usepackage{graphicx}
\usepackage{booktabs} 
\usepackage{microtype}
\usepackage{xspace}
\usepackage{authblk}
\usepackage{array}
\usepackage{dsfont}
\usepackage{pifont}
\usepackage{blindtext}
\usepackage{helvet}
\usepackage{courier}
\usepackage{color}
\usepackage{colortbl}
\usepackage{mathrsfs}
\usepackage{amscd}
\usepackage{stmaryrd}
\usepackage{amssymb}
\usepackage{amsmath}
\usepackage{graphicx}
\usepackage{tikz}
\usepackage{algorithm} 
\usepackage[noend]{algpseudocode}
\algrenewcommand\algorithmicindent{0.750em}%
\usepackage{subfig}
\usepackage{multirow}
\usepackage{pdflscape}
\usepackage{makecell}
\usepackage{hyperref}
\usepackage{longtable}
\usepackage{ltablex}
\usepackage{xcolor}
\usepackage{cite}

\usepackage{soul}
\newcommand{\added}[1]{{#1}}
\newcommand{\deleted}[1]{}
\newcommand{\substituted}[2]{{#2}}

\newcommand{\commentout}[1]{}

\title{A Survey of Safety and Trustworthiness of Deep Neural Networks: Verification, Testing, Adversarial Attack and Defence, and Interpretability \thanks{This work is supported by the UK EPSRC projects on Offshore Robotics for Certification of Assets (ORCA) [EP/R026173/1] and End-to-End Conceptual Guarding of Neural Architectures [EP/T026995/1], and ORCA Partnership Resource Fund (PRF) on Towards the Accountable and Explainable Learning-enabled Autonomous Robotic Systems, as well as the UK Dstl projects on Test Coverage Metrics for Artificial Intelligence.}}
\author[1]{Xiaowei Huang}
\author[2]{Daniel Kroening}
\author[3]{Wenjie~Ruan}
\author[4]{James Sharp}
\author[5]{Youcheng Sun}
\author[1]{Emese Thamo}
\author[2]{Min~Wu}
\author[1]{Xinping Yi}
\date{
}
\affil[1]{University of Liverpool, UK\\
\{xiaowei.huang,~emese.thamo,~xinping.yi\}@liverpool.ac.uk}
\affil[2]{University of Oxford, UK\\
\{daniel.kroening,
min.wu\}@cs.ox.ac.uk}
\affil[3]{Lancaster University, UK\\
wenjie.ruan@lancaster.ac.uk}
\affil[4]{Defence Science and Technology Laboratory (Dstl), UK\\
jsharp1@dstl.gov.uk}
\affil[5]{Queen's University Belfast, UK\\
youcheng.sun@qub.ac.uk}

\begin{document}

\newcommand{\naturenumber}{\mathbb{N}}
\newcommand{\realnumber}{\mathbb{R}}
\newcommand{\lossfunction}{\ell}
\newcommand{\regulariser}{{\cal G}}
\newcommand{\real}{\mathds{R}}

\newcommand{\layers}{\mathbb{S}}
\newcommand{\layerConnections}{\mathbb{T}}
\newcommand{\defeq}{\triangleq}
\newcommand{\domains}{D}
\newcommand{\labels}{\mathcal{L}}
\newcommand{\networks}{\mathcal{N}}
\newcommand{\requirements}{\mathcal{R}}
\newcommand{\testsuites}{\mathcal{T}}
\newcommand{\neuronpairs}{\mathsf{O}}
\newcommand{\distance}[2]{ ||#1||_{#2}}
\newcommand{\valuefunction}{g}
\newcommand{\distancefunction}{h}
\newcommand{\neuronpair}{\alpha}
\newcommand{\setofcoveringmethods}{\mathsf{F}}
\newcommand{\coveringmethod}{cov}
\newcommand{\testobjectives}{obj}
\newcommand{\covered}[3]{{#1}^{#2}#3}
\newcommand{\explanations}{{\cal{E}}}
\newcommand{\explain}{{\tt expl}}
\newcommand{\encoding}{{\tt enc}}
\newcommand{\decoding}{{\tt dec}}
\newcommand{\humanoracle}{{\cal{H}}}
\newcommand{\coverage}[3]{C_{#1}^{#2}#3}
\newcommand{\metric}{M}
\newcommand{\inputDistribution}{{\cal X}}
\newcommand{\expectation}{\mathop{\mathbb{E}}}

\newtheorem{example}{Example}
\newtheorem{definition}{Definition}

\newcommand\scalemath[2]{\scalebox{#1}{\mbox{\ensuremath{\displaystyle #2}}}}

\newcommand{\feature}{\psi}
\newcommand{\features}{\Psi}
\newcommand{\setoffeatures}{\Psi}
\newcommand{\network}{N}
\newcommand{\inputPoint}{x}
\newcommand{\region}{\eta}
\newcommand{\manipulation}{\Delta}
\newcommand{\satisfy}{\models}

\newcommand{\reach}{Reach}
\newcommand{\interval}{Interval}
\newcommand{\robust}{Robust}
\newcommand{\globalmetric}{Lips}
\newcommand{\localmetric}{M_{\mathcal{L}}}

\newcommand{\propertyexpression}{\varphi}

\newcommand{\trainingDataset}{X}

\newcounter{bar}
\newcommand{\foo}{%
	\stepcounter{bar}}

\newcommand{\comment}[2]{\foo\paragraph{\underline{Comment \thebar: #1}} \emph{#2}}
\newcommand{\answer}[1]{\paragraph{A:} #1}

\setcounter{page}{0}

\maketitle

In the past few years, significant progress has been made on deep neural networks (DNNs) in achieving human-level performance on several long-standing tasks. With the broader deployment of DNNs on various applications, the concerns over their safety and trustworthiness have been raised in public, especially after the widely reported fatal incidents involving self-driving cars. Research to address these concerns is particularly active, with a significant number of papers released in the past few years. This survey paper conducts a review of the current research effort into making DNNs safe and trustworthy, by focusing on four aspects: verification, testing, adversarial attack and defence, and interpretability. In total, we survey 202 papers, most of which were published after 2017.

\tableofcontents

\newpage 

\setcounter{section}{0}

\section*{Symbols and Acronyms}

\subsection*{List of Symbols}

We provide an incomplete list of symbols that will be used in the survey. 

\bigskip

\begin{tabular}{lp{1.0\textwidth}}
    $\networks$ & a neural network \\
    $f$ & function represented by a neural network \\
    $W$ & weight \\
    $b$ & bias \\
    $n_{k,l}$ & $l$-th neuron on the $k$-th layer\\
    $v_{k,l}$ & activation value of the $l$-th neuron on the $k$-th layer\\
    $\lossfunction$ & loss function \\
    $x$ & input \\
    $y$ & output \\
    $\region$ & a region around a point \\
    $\manipulation$ & a set of manipulations \\
    $\requirements$ & a set of test conditions \\
    $\testsuites$ & test suite \\
    $\regulariser$ & regularisation term \\
    $\inputDistribution$ & the ground truth distribution of the inputs \\
    $\epsilon$ & error tolerance bound \\ 
    $L_0$(-norm) & L0 norm distance metric\\
    $L_1$(-norm) & L1 norm distance metric\\
    $L_2$(-norm) & L2 norm distance metric\\
    $L_\infty$(-norm) & L infinity norm distance metric\\
    $\expectation$ & probabilistic expectation \\
\end{tabular}

\newpage

\subsection*{List of Acronyms}

\begin{tabular}{lp{1.0\textwidth}}
    DNN & Deep Neural Network \\ 
    AI & Artificial Intelligence \\
    DL & Deep Learning \\
    MILP & Mixed Integer Linear Programming \\
    SMT & Satisfiability Modulo Theory \\
    MC/DC & Modified Condition/Decision Coverage \\
    B\&B & Branch and Bound \\
    ReLU &  Rectified Linear Unit \\
\end{tabular}
\newpage

\section{Introduction}

In the past few years, significant progress has been made in the development of deep neural networks (DNNs), which now 
outperform humans 
in several difficult tasks, such as image classification~\cite{ILSVRC15}, natural language processing~\cite{CWBKKK2011}, and two-player games~\cite{alphaGoZero}. Given the prospect of a broad deployment of DNNs in a wide range of applications, concerns regarding the safety and trustworthiness of this approach have been raised~\cite{TeslaIncident, UberIncident}. There is significant research that aims to address these concerns, with many publications appearing in the past few years.

\begin{figure}[tb]
    \centering
    \includegraphics[width=0.7\textwidth]{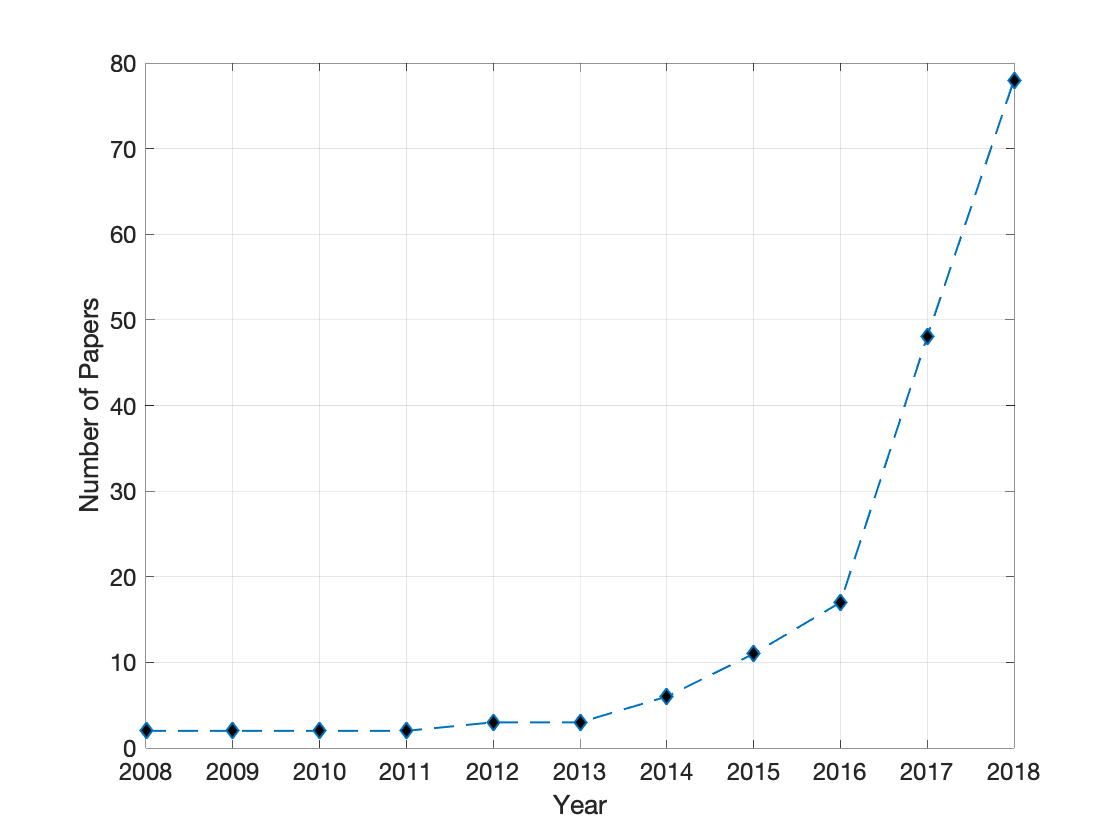}
    \caption{Number of publications, by publication year, surveyed }
    \label{fig:publicationyear}
\end{figure}

Whilst it is difficult to cover all related research activities, 
we strive to survey the past and current research efforts on making the application of DNNs safe and trustworthy. Figure~\ref{fig:publicationyear} visualises the rapid growth of this area: it gives the number of surveyed papers per calendar year, starting from 2008 to 2018. In total, we surveyed 202 papers.

Trust, or trustworthiness, is a general term and its definition varies in different contexts. Our definition is based on the practice that has been widely adopted in established industries, e.g., automotive and avionics. Specifically, trustworthiness is addressed predominantly within two processes: a certification process and an explanation process. 
The \textbf{certification} process is held  before the deployment of the product to make sure that it functions correctly (and safely). 
During the certification process, the manufacturer needs to demonstrate to the relevant certification authority, e.g., the European Aviation Safety Agency or the Vehicle Certification Agency, that the product behaves correctly with respect to a set of high-level requirements. The \textbf{explanation} process is held whenever needed during the lifetime of the product. The user manual explains a set of  expected behaviours of the product that its user may frequently experience. More importantly, an investigation can be conducted, with a formal report produced, to understand any unexpected behaviour of the product. We believe that a similar  practice should be carried out when working with data-driven deep learning systems. That is, in this survey, we  address trustworthiness based on the following concept: 
$$
\text{Trustworthiness = Certification + Explanation }
$$
In other words, a user is able to trust a system if the system has been certified by a certification authority and any of its behaviour can be well explained. Moreover, we will discuss briefly in Section~\ref{sec:humanintheloop} our view on the impact of human-machine interactions on trust. 

In this survey, we will consider the advancement of enabling techniques for both the certification and the explanation processes of DNNs. Both processes are challenging, owing to the black-box nature of DNNs and the lack of rigorous foundations.

\subsection{Certification}

For certification, an important low-level requirement for DNNs is that they are robustness against input perturbations. DNNs have been shown to suffer from a lack of robustness because of their susceptibility to \emph{adversarial examples}~\cite{szegedy2014intriguing} such that small modifications to an input, sometimes imperceptible to humans, can make the network unstable. 
Significant efforts have been  taken in order to achieve robust machine learning, including e.g., attack and defence techniques for adversarial examples. Attack techniques aim to find adversarial examples that exploit a DNN e.g., it classifies the adversarial inputs with high probability to wrong classes; meanwhile defence techniques aim to enhance the DNN so that they can identify or eliminate adversarial attack. These techniques cannot be directly applied to certify a DNN, due to their inability to provide clear assurance evidence with their results. Nevertheless, we review some prominent methods since they provide useful insights to certification techniques.

The certification techniques covered in this survey include verification and testing, both of which have been demonstrated as useful in checking the dependability of real world software and hardware systems. However, traditional techniques developed in these two areas, see e.g., \cite{AO2008,clarkebook}, cannot be be directly applied to deep learning systems, as DNNs exhibit complex internal behaviours not commonly seen within traditional verification and testing. 

DNN verification techniques determine whether a property, e.g., the local robustness for a given input $x$, holds for a DNN; if it holds, it may be feasible to supplement the answer with a mathematical proof. Otherwise, such verification techniques will return a counterexample. If a deterministic answer is hard to achieve, an answer with certain error tolerance bounds may suffice in many practical scenarios. 
Whilst DNN verification techniques are promising, they suffer from a scalability problem, due to the high computational complexity and the large size of DNNs. Up to now, DNN verification techniques either work with small scale DNNs or with approximate methods with convergence guarantees on the bounds. 

DNN testing techniques arise as a complement to the DNN verification techniques. Instead of providing mathematical proofs to the satisfiability of a property over the system, testing techniques aim to either find bugs (i.e., counterexamples to a property) or provide assurance cases \cite{Rushby2015}, by exercising the system with a large set of test cases. These testing techniques are computationally less expensive and therefore are able to work with state-of-the-art DNNs. In particular, coverage-guided testing generates test cases according to pre-specified coverage criteria. Intuitively, a high coverage suggests that more of a DNN's behaviour has been tested and therefore the DNN has a lower chance of containing undetected bugs. 

\subsection{Explanation}

The goal of Explainable AI \cite{explainableAI} is to overcome the interpretability problem of AI \cite{interpretability}; that is, to explain why the AI outputs a specific decision, say to in determining whether to give a loan someone. The EU General Data Protection Regulation (GDPR) \cite{GDRP} mandates a ``right to explanation'', meaning that an explanation of how the model reached its decision can be requested. While this ``explainability” request is definitely beneficial to the end consumer, obtaining such information from a DNN is challenging for the very developers of the DNNs.

\subsection{Organisation of This Survey}

The structure of this survey is summarised as follows. In Section~\ref{sec:preliminaries}, we will present preliminaries on the DNNs, and a few key concepts such as verification, testing, interpretability, and distance metrics. This is followed by Section~\ref{sec:safetyproblem}, which discusses safety problems and safety properties. In Section~\ref{sec:verification} and Section~\ref{sec:testing}, we review DNN verification and DNN testing techniques, respectively. The attack and defence techniques are reviewed in  Section~\ref{sec:attackndefence}; and is followed by a review of a set of interpretability techniques for DNNs in Section~\ref{sec:interpretability}. Finally, we discuss future challenges in Section~\ref{sec:challenges} and conclude this survey in Section~\ref{sec:concl}. 

\begin{figure}[h]
    \centering
    \includegraphics[scale=0.6]{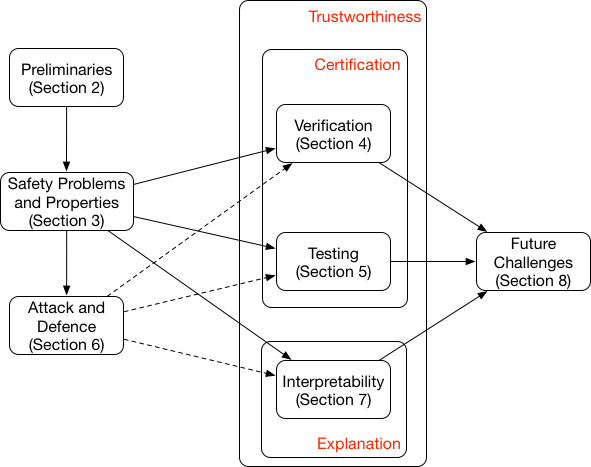}
    \caption{Relationship between sections}
    \label{fig:section_diagram}
\end{figure}

Figure~\ref{fig:section_diagram} outlines the causality relation between the sections in this paper. We use dashed arrows from attack and defence techniques (Section~\ref{sec:attackndefence}) to several sections because of their potential application in certification and explanation techniques.

\newpage

\cleardoublepage

\section{Preliminaries}
\label{sec:preliminaries}

In the following, we provide preliminaries on deep  neural networks, automated verification, software testing, interpretability, and distance metrics. 

\subsection{Deep  Neural Networks}

A (deep and feedforward) neural network, or DNN, is a tuple $\networks=(\layers,
\layerConnections, \Phi)$, where $\layers=\{\layers_k~|~k\in\{1..K\}\}$ is a set of layers,
$\layerConnections\subseteq \layers\times \layers$ is a set of connections between layers and 
$\Phi=\{\phi_k~|~k\in\{2..K\}\}$ is a set of functions, one for each non-input layer.
%
%
In a DNN, $\layers_1$ is the \emph{input} layer, $\layers_{K}$ is the \emph{output} layer,
and layers other than input and output layers are called \emph{hidden layers}.
Each layer $\layers_k$ consists of $s_k$ 
\emph{neurons} (or nodes).
The $l$-th node of layer $k$ is denoted by $n_{k,l}$.

Each node $n_{k,l}$ for $2 \leq k\leq K$ and  $1\leq l\leq s_k$ is associated with two variables $u_{k,l}$ and $v_{k,l}$, to record  its values before and after an activation function,
respectively.
The Rectified Linear Unit (ReLU) \cite{relu} is one of the most popular 
activation functions for DNNs, according to which the \emph{activation 
value} of each node of hidden layers is defined as
\begin{equation}
    \label{eq:relu}
    v_{k,l}=ReLU(u_{k,l})=
    \begin{cases}
        u_{k,l} &\mbox{  if } u_{k,l}\geq 0 \\
            0 & \mbox{  otherwise}
    \end{cases}
\end{equation}

%
Each input node $n_{1,l}$ for $1\leq l\leq s_1$ is associated with a
variable $v_{1,l}$ and each output node $n_{K,l}$ for $1\leq l\leq s_K$ is
associated with a variable $u_{K,l}$, because no activation function is
applied on them. Other popular activation functions beside ReLU include: sigmoid, tanh, and softmax. 

Except for the nodes at the input layer, every node is connected to nodes in the
preceding layer by pre-trained parameters such that for all $k$ and $l$ with
$2 \leq k\leq K$ and  $1\leq l\leq s_k$
\begin{equation}
  \label{eq:sum}
  u_{k,l}=b_{k,l}+\sum_{1\leq h \leq s_{k-1}} w_{k-1, h, l}\cdot v_{k-1,h}
\end{equation}
where $w_{k-1,h,l}$ is the weight for the connection between
$n_{k-1,h}$ (i.e., the $h$-th node of layer $k-1$) and $n_{k,l}$
(i.e., the $l$-th node of layer $k$), and $b_{k,l}$ the
so-called \emph{bias} for node $n_{k,l}$.  We note that this
definition can express both fully-connected functions and
convolutional functions\footnote{Many of the surveyed techniques can work with other types of functional layers such as max-pooling, batch-normalisation, etc. Here  for simplicity, we omit their expressions.}.
The function $\phi_k$ is the composition of Equation (\ref{eq:relu}) and (\ref{eq:sum}) by having $u_{k,l}$ for $1\leq l\leq s_k$ as the intermediate variables. Owing to the use of the ReLU as in \eqref{eq:relu}, the behavior of a neural
network is highly non-linear. 

Let $\real$ be the set of real numbers. We let $\domains_{k} = \real^{s_k}$ be the vector space
associated with layer $\layers_k$, one dimension for each variable $v_{k,l}$. 
Notably, every point $x\in \domains_{1}$ is an input. Without loss of generality, the dimensions of an input are normalised as real values in $[0,1]$, i.e., $D_1=[0,1]^{s_1}$. 
%
A DNN $\networks$ can alternatively be expressed as a function $f: \domains_{1}\rightarrow \domains_{K}$ such that 
\begin{equation}
f(x) = \phi_{K}(\phi_{K-1}(...\phi_2(x)))
\end{equation}
Finally, for any input, the DNN $\networks$ assigns a \emph{label}, that is,
the index of the node of output layer with the largest value:
\begin{equation}
\mathit{label}=\mathrm{argmax}_{1\leq l\leq s_K}u_{K,l}
\end{equation}
Moreover, we let $\labels=\{1..s_K\}$ be the set of labels. 
\begin{example}
Figure \ref{fig:nn} is a simple DNN with four layers. 
The input space is $\domains_{1}=[0,1]^2$, the two hidden vector spaces are $D_2=D_3=\real^3$, and the set of labels is $\labels=\{1,2\}$.

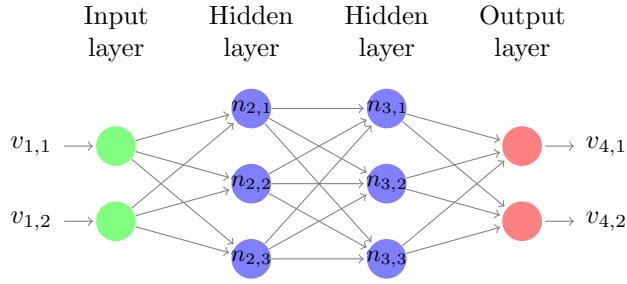
\begin{figure}[htp!]
\centering

\def\layersep{1.8cm}

\scalebox{1}{
\begin{tikzpicture}[shorten >=1pt,->,draw=black!50, node distance=\layersep]
    \tikzstyle{every pin edge}=[<-,shorten <=1pt]
    \tikzstyle{neuron}=[circle,fill=black!25,minimum size=15pt,inner sep=0pt]
    \tikzstyle{input neuron}=[neuron, fill=green!50];
    \tikzstyle{output neuron}=[neuron, fill=red!50];
    \tikzstyle{hidden neuron}=[neuron, fill=blue!50];
    \tikzstyle{annot} = [text width=4em, text centered]

    \foreach \name / \y in {1,...,2}
        \node[input neuron, pin=left:$v_{1,\y}$] (I-\name) at (0,-\y) {};

    \foreach \name / \y in {1,...,3}
        \path[yshift=0.5cm]
            node[hidden neuron] (H1-\name) at (\layersep,-\y cm) {};

    \foreach \name / \y in {1,...,3}
        \path[yshift=0.5cm]
            node[hidden neuron] (H2-\name) at (\layersep*2,-\y cm) {};

    \node[output neuron,pin={[pin edge={->}]right:$v_{4,1}$}, right of=H2-2, yshift=0.5cm] (O1) {};
    \node[output neuron,pin={[pin edge={->}]right:$v_{4,2}$}, right of=H2-2, yshift=-0.5cm] (O2) {};

    \foreach \source in {1,...,2}
        \foreach \dest in {1,...,3}
            \path (I-\source) edge (H1-\dest);

    \foreach \source in {1,...,3}
        \foreach \dest in {1,...,3}
            \path (H1-\source) edge (H2-\dest);

    \foreach \source in {1,...,3}
         \path (H2-\source) edge (O1);

    \foreach \source in {1,...,3}
         \path (H2-\source) edge (O2);

    \node[annot,above of=H1-1, node distance=1cm] (hl1) {Hidden layer};
    \node[annot,above of=H2-1, node distance=1cm] (hl2) {Hidden layer};
    \node[annot,left of=hl1] {Input layer};
    \node[annot,right of=hl2] {Output layer};

    \node[annot, right of=H1-1, node distance=0.0cm] (hl1) {\small $n_{2,1}$};
    \node[annot, right of=H1-2, node distance=0.0cm] (hl1) {\small $n_{2,2}$};
    \node[annot, right of=H1-3, node distance=0.0cm] (hl1) {\small $n_{2,3}$};
    \node[annot, right of=H2-1, node distance=0.0cm] (hl1) {\small $n_{3,1}$};
    \node[annot, right of=H2-2, node distance=0.0cm] (hl1) {\small $n_{3,2}$};
    \node[annot, right of=H2-3, node distance=0.0cm] (hl1) {\small $n_{3,3}$};
\end{tikzpicture}
}
  \caption{A simple neural network}
  \label{fig:nn}
\end{figure}

\end{example}
\bigskip


Given one particular input $x$, 
the DNN $\networks$ is
\emph{instantiated} and we use $\networks[x]$ to denote this instance of the
network.  In $\networks[x]$, for each node $n_{k,l}$, the values of the variables $u_{k,l}$ and $v_{k,l}$ are fixed and denoted as $u_{k,l}[x]$ and $v_{k,l}[x]$, respectively. 
%
Thus, the activation
or deactivation of each ReLU operation in the network is similarly determined.  
We define
  \begin{equation}
    \label{eq:sign}
    \mathit{sign}_\networks(n_{k,l},x)=
    \begin{cases}
      +1 &\mbox{  if } u_{k,l}[x] = v_{k,l}[x] \\
      -1 & \mbox{  otherwise}
    \end{cases}
  \end{equation}
The subscript $\networks$ will be omitted when clear from the context. 
The classification label of $x$ is denoted as $\networks[x].\mathit{label}$.

\begin{example}\label{example:weights}
Let $\networks$ be a DNN whose architecture is given in Figure \ref{fig:nn}.  
Assume that the weights for the first three layers are as follows:
$$
W_{1}=\scalemath{0.8}{
\begin{bmatrix}
  4 & 0 & -1\\
  1 & -2 & 1
\end{bmatrix}},\,\,
W_{2}=\scalemath{0.8}{
\begin{bmatrix}
  2 & 3 & -1\\
  -7 & 6 & 4 \\
  1 & -5 & 9
\end{bmatrix}}
$$
and that all biases are 0. When given an input 
$x=[0, 1]$, we get $\mathit{sign}(n_{2,1},x)=+1$, since
$u_{2,1}[x]=v_{2,1}[x]=1$, and $\mathit{sign}(n_{2,2},x)=-1$,
since $u_{2,2}[x] = -2 \neq 0 = v_{2,2}[x]$. 
\end{example}

\subsection{Verification}

Given a DNN $\networks$ and a property $C$, verification is considered a set of techniques to check whether the property $C$ holds on $\networks$.  
Different from other techniques such as testing and adversarial attack, a verification technique needs to provide provable guarantees for its results. A  guarantee can be either a Boolean guarantee, an approximate guarantee, an anytime approximate guarantee, or a statistical guarantee. A Boolean guarantee means that the verification technique is able to affirm when the property holds, or otherwise return a counterexample. An approximate guarantee provides either an over-approximation or an under-approximation to a property. An anytime approximate guarantee has both an over-approximation and an under-approximation, and the approximations can be continuously improved until converged. All the above guarantees need to be supported with a mathematical proof.
When a mathematical proof is hard to achieve, a statistical guarantee provides a quantitative error tolerance bound  on the resulting claim.  

We will formally define safety properties for DNNs in Section~\ref{sec:safetyproblem}, together with their associated verification problems and provable guarantees.



\subsection{Testing}


Verification problems usually have  high computational complexity, such as being NP-hard, when the properties are simple input-output constraints \cite{katz2017reluplex,RHK2018}. This, compounded with the high-dimensionality and the high non-linearity of DNNs, makes the existing verification techniques hard to work with for industrial scale DNNs. This computational intensity can be partially alleviated by considering testing techniques, at the price of provable guarantees. Instead, assurance cases are pursued in line with thouse of existing safety critical systems~\cite{Rushby2015}. 

The goal of testing DNNs is to generate a set of test cases, that can demonstrate confidence in a DNN's performance, when passed, such that they can support an assurance case.  
Usually, the generation of test cases is guided by coverage metrics. 

Let $\mathsf{N}$ be a set of DNNs,
$\mathsf{R}$ a set of test objective sets, and $\mathsf{T}$ a set of test
suites. We use $\networks,\requirements,\testsuites$ to range over $\mathsf{N},\mathsf{R}$ and $\mathsf{T}$, respectively. Note that, normally both $\requirements$ and $\testsuites$ contain a set of elements by themselves. The following is an adaption of a definition in \cite{ZHM1997} for software testing:

\begin{definition}[Test Accuracy Criterion/Test Coverage Metric]\label{def:generalcriteria}
A test adequacy criterion, or a test coverage metric, is a function $\metric:\mathsf{N}\times \mathsf{R}\times
\mathsf{T}\rightarrow [0,1]$.
\end{definition}

Intuitively, $\metric(\networks,\requirements,\testsuites)$ quantifies the
degree of adequacy to which a DNN $\networks$ is tested by a test
suite $\testsuites$ with respect to a set $\requirements$ of test objectives.
Usually, the greater the number
$\metric(\networks,\requirements,\testsuites)$, the more adequate the
testing\footnote{We may use criterion and metric
interchangeably.}.  
We will elaborate on the test criteria in Section~\ref{sec:testing}.
 Moreover, a testing oracle is a mechanism that determines whether the DNN behaves correctly for a test case. It is dependent on the properties to be tested, and  will be discussed further in Section~\ref{sec:safetyproblem}.

\subsection{Interpretability}

Interpretability is an issue arising as a result of the black-box nature of DNNs. Intuitively, it should provide a human-understandable explanation for the behaviour of a DNN. An explanation procedure can be separated into two steps: an extraction step and an exhibition step. The \emph{extraction step} obtains an intermediate representation, and the \emph{exhibition step} presents the obtained intermediate representation in a way easy for human users to understand. Given the fact that DNNs are usually high-dimensional, and the information should be made accessible for understanding by the human users, the intermediate representation needs to be at a lower dimensionality. Since the exhibition step is closely related to the intermediate representation, and is usually conducted by e.g., visualising the representation, we will focus on the extraction step in this survey. 

Depending on the requirements, the explanation can be either an instance-wise explanation, a model explanation, or an information-flow explanation. In the following, we give their general definitions, trying to cover as many techniques to be reviewed as possible. 

\begin{definition}[Instance-Wise Explanation]\label{def:instanceExplanation}
Given a function $f:\real^{s_1}\rightarrow \real^{s_K}$, which represents a DNN $\networks$, and an input $x\in \real^{s_1}$, an instance-wise explanation $\explain(f,x)\in \real^{t}$ is another representation of $x$ such that $t \leq s_1$.  
\end{definition}

Intuitively, for instance-wise explanation, the goal is to find another representation of an input $x$ (with respect to the function $f$ associated to the DNN $\networks$), with the expectation that the representation carries simple, yet essential, information that can help the user understand the decision $f(x)$. Most of the techniques surveyed in Section~\ref{sec:visualisation}, Section~\ref{sec:ranking}, and Section~\ref{sec:saliencymaps} fit with this definition. 

\begin{definition}[Model Explanation]\label{def:modelExplanation}
Given a function $f:\real^{s_1}\rightarrow \real^{s_K}$, which represents a DNN $\networks$, a model explanation $\explain(f)$ includes two functions $g_1:\real^{a_1}\rightarrow \real^{a_2}$, which is a  representation of $f$ such that $a_1 \leq s_1$ and $a_2\leq s_K$, and $g_2:\real^{s_1}\rightarrow \real^{a_1}$, which maps original inputs to valid inputs of the function $g_1$.  
\end{definition}

Intuitively, the point of model explanation is to find a simpler model which can not only be used for prediction by applying $g_1(g_2(x))$ (with certain loss) but also  be comprehended by the user. Most of the techniques surveyed in Section~\ref{sec:simplermodels} fit with this definition. There are other model explanations such as the influence function based approach reviewed in Section~\ref{sec:influencefunction}, which provides explanation by comparing different learned parameters by e.g., up-weighting some training samples. 


Besides the above two deterministic methods for the explanation of data and models, there is another stochastic explanation method for the explanation of information flow in the DNN training process. 

\begin{definition}[Information-Flow Explanation]\label{def:inforExplanation}
Given a function family $\mathcal{F}$, which represents a stochastic DNN, an information-flow explanation $\explain(\mathcal{F})$ includes a stochastic encoder $g_1(T_k|X)$, which maps the input $X$ to a representation $T_k$ at layer $k$, and a stochastic decoder $g_2(Y|T_k)$, which maps the representation $T_k$ to the output $Y$.
\end{definition}

The aim of the information-flow explanation is to find the optimal information representation of the output at each layer when information (data) flow goes through, and understand why and how a function $f \in \mathcal{F}$ is chosen as the training outcome given the training dataset $(X,Y)$.
The information is transparent to data and models, and its representations can be described by some quantities in information theory, such as entropy and mutual information.
This is an emerging research avenue for interpretability, and a few information theoretical approaches will be reviewed in Section~\ref{sec:informationtheory}, which aim to provide a theoretical explanation to the training procedure.

\subsection{Distance Metric and $d$-Neighbourhood}

Usually, a distance function is employed to compare inputs. Ideally, such a distance should reflect perceptual similarity between inputs, comparable to e.g., human perception for image classification networks. A distance metric should satisfy a few axioms which are usually needed for defining a metric space: 
\begin{itemize}
\item $||x||\geq 0$ (non-negativity),
\item $||x-y||=0$ implies that $x = y$ (identity of indiscernibles),
\item $||x-y|| = ||y-x||$ (symmetry),
\item $||x-y||+||y-z|| \geq ||x-z||$ (triangle inequality).
\end{itemize}

In practise, $L_p$-norm distances are used, including

\begin{itemize}
    \item $L_1$ (Manhattan distance): $||x||_1 = \sum_{i=1}^{n} |x_i|$
    \item $L_2$ (Euclidean distance): $||x||_2 = \sqrt{\sum_{i=1}^{n} x_i^2}$
    \item $L_\infty$ (Chebyshev distance): $||x||_\infty  = \max_{i} |x_i|$
\end{itemize}
Moreover, we also consider $L_0$-norm as $||x||_0 = |\{x_i~|~x_i\neq 0, i = 1..n\}|$, i.e., the number of non-zero elements. Note that, $L_0$-norm does not satisfy the triangle inequality. 
In addition to these, there exist other distance metrics such as Fréchet Inception Distance \cite{10.5555/3295222.3295408}. 

Given an input $x$ and a distance metric $L_p$, the \emph{neighbourhood} of $x$ is defined as follows.
\begin{definition}[d-Neighbourhood]\label{def:inputregion}
Given an input $x$, a distance function $L_p$, and a distance $d$, we define 
the \emph{d-neighbourhood} $\eta(x,L_p,d)$ of $x$ w.r.t. $L_p$ as 
\begin{equation}
\eta(x,L_p,d)=\{\hat{x} ~|~ \distance{\hat{x}-x}{p} \leq d\},
\end{equation}
the set of inputs whose distance to $x$ is no greater than $d$ with respect to $L_p$.  
\end{definition}
\cleardoublepage

\section{Safety Problems and Safety Properties} \label{sec:safetyproblem}

\begin{figure}[thb]
  \centering
  \includegraphics[width=1\linewidth]{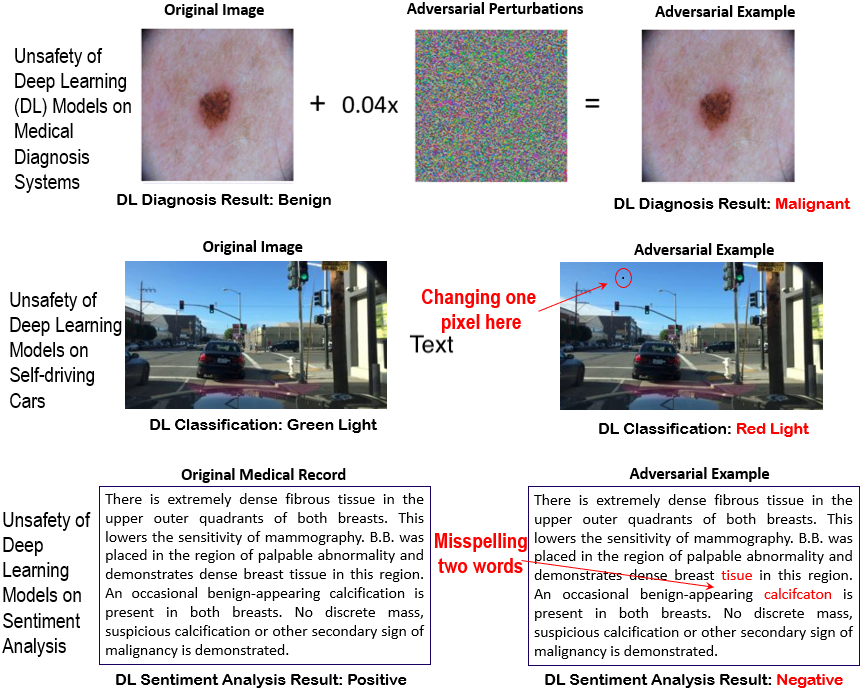}
  \caption{Examples of erroneous behaviour on deep learning models. {\em Top Row~\cite{finlayson2018adversarial}: In a medical diagnosis system, a ``Benign'' tumour is misclassified as ``Malignant'' after adding a small amount of human-imperceptible perturbations; Second Row~\cite{wu2019game}: By just changing one pixel in a ``Green-Light'' image, a state-of-the-art DNN misclassifies it as ``Red-Light''; Bottom Row~\cite{ebrahimi2018hotflip}: In a sentiment analysis task for medical records, with two misspelt words, a well-trained deep learning model classifies a ``Positive'' medical record as ``Negative''.} }
  \label{fig-e2}
\end{figure}

Despite the success of deep learning (DL) in many areas, serious concerns have been raised in applying DNNs to real-world safety-critical systems such as self-driving cars, automatic medical diagnosis, etc.
In this section, we will discuss the key safety problem of DNNs, and present a set of safety features that analysis techniques are employing. 

For any $f(x)$ whose value is a vector of scalar numbers, we use $f_j(x)$ to denote its $j$-th element. 

\begin{definition}[Erroneous Behavior of DNNs]
Given a (trained) deep neural network $f: \real^{s_1} \to \real^{s_K}$, a human decision oracle $\humanoracle: \real^{s_1} \to \real^{s_K}$, and a legitimate input $x\in \real^{s_1}$, an erroneous behavior of the DNN is such that 
\begin{equation}
\arg\max_j f_j(x) \neq \arg \max_j \humanoracle_j(x)
\end{equation}
\end{definition}

Intuitively, an erroneous behaviour is witnessed by the existence of an input $x$ on which the DNN and a human user have different perception.

\subsection{Adversarial Examples}

Erroneous behaviours include those training and test samples which are classified incorrectly by the model and have safety implications. Adversarial examples \cite{szegedy2014intriguing} represent another class of erroneous behaviours that also introduce safety implications. Here, we take the name ``adversarial example'' due to historical reasons. Actually, as suggested in the below definition, it represents a mis-match of the decisions made by a human and by a neural network, and does not necessarily involve an adversary.

\begin{definition}[Adversarial Example]\label{def:adversarialexample}
Given a (trained) deep neural network $f: \real^{s_1} \to \real^{s_K}$, a human decision oracle $\humanoracle: \real^{s_1} \to \real^{s_K}$, and a legitimate input $x\in \real^{s_1}$ with $\arg\max_j f_j(x) = \arg \max_j \humanoracle_j(x)$, an adversarial example to a DNN is defined as:
\begin{equation}
\begin{array}{ll}
    \exists	\hat x: & \arg \max_j \humanoracle_j(\hat x) = \arg \max_j \humanoracle_j(x) \\ & \land ~ || x- \hat{x} ||_p \leq d\\
    & \land~\arg \max_j f_j(\hat x) \neq \arg \max_j f_j(x) 
\end{array}
\end{equation}
where $p\in \naturenumber$,  $p\geq 1$, $d\in\realnumber$, and $d>0$. 
\end{definition}

Intuitively, $x$ is an input on which the DNN and an human user have the same classification and, based on this, an adversarial example is another input $\hat x$ that is classified differently than $x$ by network $f$ (i.e., $\arg \max_j f_j(\hat x) \neq \arg \max_j f_j(x)$), even when  they are geographically close in the input space
(i.e., $|| x- \hat{x} ||_p \leq d$) and the human user believes that they should be the same (i.e., $\arg \max_j \humanoracle_j(\hat x) = \arg \max_j \humanoracle_j(x)$). We do not consider the labelling error introduced by human operators, because it is part of the Bayes error which is irreducible for a given classification problem. On the other hand, the approaches we review in the paper are for the estimation error, which measures how far the learned network $\networks$ is from the best network of the same architecture. 

Figure~\ref{fig-e2} shows three concrete examples of the safety concerns brought by the potential use of DNNs in safety-critical application scenarios including: medical diagnosis systems, self-driving cars and automated sentiment analysis of medical records.

\begin{example}
As shown in the top row of Figure~\ref{fig-e2}, for an fMRI image, a human-invisible perturbation will turn a DL-enabled diagnosis decision of ``malignant tumour" into ``benign tumour". In this case, the human oracle is the medical expert~\cite{finlayson2018adversarial}.
\end{example}

\begin{example}
As shown in the second row of Figure~\ref{fig-e2}, in classification tasks, by adding a small amount of adversarial perturbation (w.r.t. $L_p$-norm distance), the DNNs will misclassify an image of traffic sign ``red light" into ``green light"~\cite{wicker2018feature,wu2019game}. In this case, the human decision oracle $\humanoracle$ is approximated by stating that two inputs within a very small $L_p$-norm distance are the same.
\end{example}

\begin{example}
In a DL-enabled end-to-end controller deployed in autonomous vehicles, by adding some natural transformations such as ``rain", the controller will output an erroneous decision, ``turning left", instead of a righteous decision, ``turning right"~\cite{deeproad}. However,  it is clear that, from the human driver's point of view, adding ``rain" should not change the driving decision of a car. 
\end{example}

\begin{example}
As shown in the bottom row of Figure~\ref{fig-e2}, for medical record, some minor misspellings -- which happen very often in the medical records --  will lead to significant mis-classification on the diagnosis result, from ``Positive'' to ``Negative''. 
\end{example}


As we can see, these unsafe, or erroneous, phenomenon acting on DNNs are essentially caused by the inconsistency of the decision boundaries from DL models (that are learned from  training datasets) and human oracles. This inevitably raises significant concerns on whether DL models can be safely applied in safety-critical domains. 

In the following, we review a few safety properties that have been studied in the literature. 


\subsection{Local Robustness Property}

Robustness requires that the decision of a DNN is invariant against small perturbations. The following definition is adapted from that of \cite{HKWW2017}.

\begin{definition}[Local Robustness]
	Given a DNN $\networks$ with its associated function $f$, and an input region $\eta\subseteq [0,1]^{s_1}$, the (un-targeted) local robustness of $f$ on $\eta$ is defined as 
\begin{equation}
\robust(f,\eta) \defeq \forall x\in \eta, \exists~ l\in [1..s_K], \forall j\in [1..s_K]: f_l(x) \geq f_{j}(x)
\end{equation}
For targeted local robustness of a label $j$, it is defined as 
\begin{equation}
\robust_j(f,\eta) \defeq\forall x\in \eta, \exists~ l\in [1..s_K]: f_l(x) > f_{j}(x)
\end{equation}
\end{definition}

Intuitively, local robustness states that all inputs in the region $\eta$ have the same class label. More specifically, there exists a label $l$ such that, for all inputs $x$ in region $\eta$, and other labels $j$, the DNN believes that $x$ is more possible to be in class $l$ than in any class $j$. Moreover, targeted local robustness means that a specific label $j$ cannot be perturbed for all inputs in $\eta$; specifically, all inputs $x$ in $\eta$ have a class $l \neq j$, which the DNN believes is more possible than the class $j$.  Usually, the region $\eta$ is defined with respect to an input $x$ and a norm $L_p$, as in Definition~\ref{def:inputregion}. If so, it means that all inputs in $\eta$ have the same class as input $x$. For targeted local robustness, it is required that none of the inputs in the region $\eta$ is classified as a given label $j$.

In the following, we define a test oracle for the local robustness property. Note that, all existing testing approaches surveyed relate to local robustness, and therefore we only provide the test oracle for local robustness. 

\begin{definition}[Test Oracle of Local Robustness Property] Let $D$ be a set of correctly-labelled inputs.  
Given a norm distance $L_p$ and a real number $d$, a test case $(x_1,...,x_k) \in \testsuites$ passes the test oracle's local robustness property, or oracle for simplicity, if 
\begin{equation}
\forall 1\leq i\leq k\exists~ x_0\in D: x_i\in \eta(x_0,L_p,d)
\end{equation}

\end{definition}

Intuitively, a test case $(x_1,...,x_k)$ passes the  oracle if all of its components $x_i$ are close to one of the correctly-labelled inputs, with respect to $L_p$ and $d$.  Recall that, we define $\eta(x_0,L_p,d)$ in Definition~\ref{def:inputregion}.

\subsection{Output Reachability Property}

\newcommand{\of}{w}

Output reachability computes the set of outputs with respect to a given set of inputs. We follow the naming convention from \cite{xiang2017output,RHK2018,ruan2018reachability}. Formally, we have the following definition.  

\begin{definition}[Output Reachability]\label{def:outputreachability}
	Given a DNN $\networks$ with its associated function $f$, and an input region $\eta\subseteq [0,1]^{s_1}$, the output reachable set of $f$ and $\eta$ is a set $\reach(f,\eta)$ such that  
\begin{equation}\label{equ:reachability}
\reach(f,\eta)\defeq \{f(x)~|~ x\in \eta\}
\end{equation}
\end{definition}

The region $\eta$ includes a large, and potentially infinite, number of inputs, so that no practical algorithm that can exhaustively check their classifications. 
The output reachability problem is  \emph{highly non-trivial} for this reason, and that 
$f$ is highly non-linear (or black-box). Based on this, we can define the following verification problem. 
\begin{definition}[Verification of Output Reachability]\label{def:verifyreachability}
	Given a DNN $\networks$ with its associated function $f$,  an input region $\eta\subseteq [0,1]^{s_1}$, and an output region ${\cal Y}$, the verification of output reachability on $f$, $\eta$, and  ${\cal Y}$ is to determine if 
\begin{equation}
\reach(f,\eta)={\cal Y}.
\end{equation} 
\end{definition}

Thus, the verification of reachability determines whether all inputs in $\eta$ are mapped onto a given set ${\cal Y}$ of outputs, and whether all outputs in ${\cal Y}$ have a corresponding $x$ in $\eta$. 

As a simpler question, we might be interested in  computing for a specific dimension of the set $\reach(f,\eta)$, its greatest or smallest value; for example, the greatest classification confidence of a specific label. We call such a problem a reachability problem. 

\subsection{Interval Property}

The Interval property computes a convex over-approximation of the output reachable set. We follow the naming convention from interval-based approaches, which are a typical class of methods for computing this property. Formally, we have the following definition. 

\begin{definition}[Interval Property]\label{def:intervalproperty}
	Given a DNN $\networks$ with its associated function $f$, and an input region $\eta\subseteq [0,1]^{s_1}$, the interval property of $f$ and $\eta$ is a convex set  $\interval(f,\eta)$ such that 
\begin{equation}\label{equ:interval}
\interval(f,\eta)\supseteq\{f(x)~|~x\in \eta\} 
\end{equation}
Ideally, we expect this set to be a convex hull of points in $\{f(x)~|~x\in \eta\}$. A convex hull of a set of points is the smallest convex set that contains the points. 
\end{definition}

While the computation of such a set can be trivial since $[0,1]^{s_1}\supseteq\{f(x)~|~x\in \eta\}$, it is expected that $\interval(f,\eta)$ is as close as possible to $\{f(x)~|~x\in \eta\}$, i.e., ideally it is a convex hull.  Intuitively, an interval is an over-approximation of the output reachability. Based on this, we can define the following verification problem.

\begin{definition}[Verification of Interval Property]\label{def:verifyinterval}
	Given a DNN $\networks$ with its associated function $f$,  an input region $\eta\subseteq [0,1]^{s_1}$, and an output region ${\cal Y}$ represented as a convex set, the verification of the interval property on $f$, $\eta$, and ${\cal Y}$ is to determine if
\begin{equation}\label{equ:interval2}
{\cal Y} \supseteq\{f(x)~|~x\in \eta\}
\end{equation} 
In other words, it is to determine whether the given ${\cal Y}$ is an interval satisfying Expression (\ref{equ:interval}). 
\end{definition}

Intuitively, the verification of the interval property determine whether all inputs in $\eta$ are mapped onto ${\cal Y}$. Similar to the reachability property, we might also be interested in simpler problems e.g., determining  whether a given real number $d$ is a valid upper bound for a specific dimension of $\{f(x)~|~x\in \eta\}$.


\subsection{Lipschitzian Property}


The Lipschitzian property, inspired by the Lipschitz continuity (see textbooks such as \cite{OSearcoid2006}), monitors the changes of the output with respect to small changes of the inputs. 

\begin{definition}[Lipschitzian Property]\label{def:globalmetric}
	Given a DNN $\networks$ with its associated function $f$, an input region $\eta\subseteq [0,1]^{s_1}$, and the $L_p$-norm, 
	\begin{equation}
		\globalmetric(f,\eta,L_p) \equiv \sup_{x_1,x_2 \in \eta}\dfrac{|f(x_1)-f(x_2)|}{||x_1-x_2||_p}
	\end{equation}
	is a Lipschitzian metric of $f$, $\eta$, and $L_p$.
\end{definition}

Intuitively, the value of this metric is the best Lipschitz constant.
Therefore, we have the following verification problem. 

\begin{definition}[Verification of Lipschitzian Property]\label{def:verifyLipschitz}
Given a Lipschitzian metric $\globalmetric(f,\eta,L_p)$ and a real value $d\in \real$,  it must be determined whether \begin{equation}
\globalmetric(f,\eta,L_p) \leq d. 
\end{equation}
\end{definition}

\subsection{Relationship between Properties}\label{sec:propertyrelation}

\begin{figure}[h]
    \centering
    \includegraphics[scale=0.5]{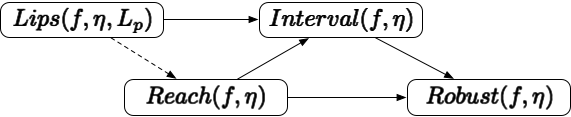}
    \caption{Relationship between properties. An arrow from a value $A$ to another value $B$ represents the existence of a simple computation to enable the computation of $B$ based on $A$. The dashed arrow  between $\globalmetric(f,\eta,L_p)$ and $\reach(f,\eta)$ means that the computation is more involved.}
    \label{fig:property_diagram}
\end{figure}

Figure~\ref{fig:property_diagram} gives the relationship between the four properties discussed above. An arrow from a value $A$ to another value $B$ represents the existence of a simple computation to enable the computation of $B$ based on $A$. For example, given a Lipschitzian metric $\globalmetric(f,\eta,L_p)$ and $\eta=\eta(x,L_p,d)$, we can compute an interval %
\begin{equation}
\interval(f,\eta)=[f(x)-\globalmetric(f,\eta,L_p)\cdot d, f(x)+\globalmetric(f,\eta,L_p)\cdot d]
\end{equation}
It can be verified that $\interval(f,\eta) \supseteq\{f(x)~|~x\in \eta\}
$. 
Given an interval $\interval(f,\eta)$ or a reachable set $\reach(f,\eta)$, we can check their respective robustness by determining the following expressions: 
\begin{equation}
\interval(f,\eta) \subseteq {\cal Y}_l = \{y\in\real^{s_K}~|~\forall j\neq l: y_l \geq y_j\} \text{, for some }l
\end{equation}
\begin{equation}
\reach(f,\eta) \subseteq {\cal Y}_l = \{y\in\real^{s_K}~|~\forall j\neq l: y_l \geq y_j\} \text{, for some }l
\end{equation}
where $y_l$ is the $l$-entry of the output vector $y$. The relation between $\reach(f,\eta)$ and $\interval(f,\eta)$ is  an implication relation, which can be seen from their definitions. Actually, if we can compute precisely the set $\reach(f,\eta)$, the inclusion of the set in another convex set $\interval(f,\eta)$ can be computed easily. 


Moreover, we use a dashed arrow  between $\globalmetric(f,\eta,L_p)$ and $\reach(f,\eta)$, as the computation is more involved by e.g., algorithms from \cite{RHK2018,wicker2018feature,WZCYSGHD2018}. 

\subsection{Instancewise Interpretability} 

First, we need to have a ranking among explanations for a given input.

\begin{definition}[Human Ranking of Explanations]
Let $\networks$ be a network with associated function $f$ and  $\explanations\subseteq \real^t$ be the set of possible explanations. We define an evaluation function $eval_{{\cal H}}: \real^{s_1}\times \explanations \rightarrow [0,1]$, which assigns for each input $x\in \real^{s_1}$ and each explanation $e\in \explanations$ a probability value $eval_{{\cal H}}(x,e)$ in $[0,1]$ such that a higher value suggests a better explanation of $e$ over $x$.  
\end{definition} 

Intuitively, $eval_{{\cal H}}(x,e)$ is a ranking of the explanation $e$ by human users, when given an input $x$. For example, given an image and an explanation algorithm which highlights part of an image, human users are able to rank all the highlighted images. 
While this ranking can be seen as the ground truth for the instance-wise interpretability, similar to using  distance metrics to measure human perception, it is hard to approximate. Based on this, we have the following definition. 

\begin{definition}[Validity of Explanation]\label{def:instance-wiseProperty}
	Let $f$ be the associated function of a DNN $\networks$, $x$ an input, and $\epsilon>0$ a real number, $\explain(f,x)\in \explanations\subseteq \real^t$ is a valid instance-wise explanation  if \begin{equation}
	  eval_{{\cal H}}(x,\explain(f,x)) > 1-\epsilon.
	\end{equation} 
\end{definition}

Intuitively, an explanation is valid if it should be among the set of explanations that are ranked sufficiently high by human users.  
\cleardoublepage
\newcommand{\DLV}{\mathsf{DLV}}
\newcommand{\DeepGame}{\mathsf{DeepGame}}
\newcommand{\TRE}{\mathsf{DeepTRE}}
\newcommand{\Reluplex}{\mathsf{Reluplex}}
\newcommand{\Planet}{\mathsf{Planet}}
\newcommand{\Sherlock}{\mathsf{Sherlock}}
\newcommand{\DeepGO}{\mathsf{DeepGO}}
\newcommand{\HySAT}{\mathsf{HySAT}}
\newcommand{\ReluVal}{\mathsf{ReluVal}}
\newcommand{\CLEVER}{\mathsf{CLEVER}}
\newcommand{\FastLin}{\mathsf{FastLin}}
\newcommand{\FastLip}{\mathsf{FastLip}}
\newcommand{\Crown}{\mathsf{Crown}}
\newcommand{\RecurJac}{\mathsf{RecurJac}}
\newcommand{\AIsquared}{\mathsf{AI^2}}

\newcommand{\ReLU}{\mathrm{ReLU}}

\section{Verification}
\label{sec:verification}

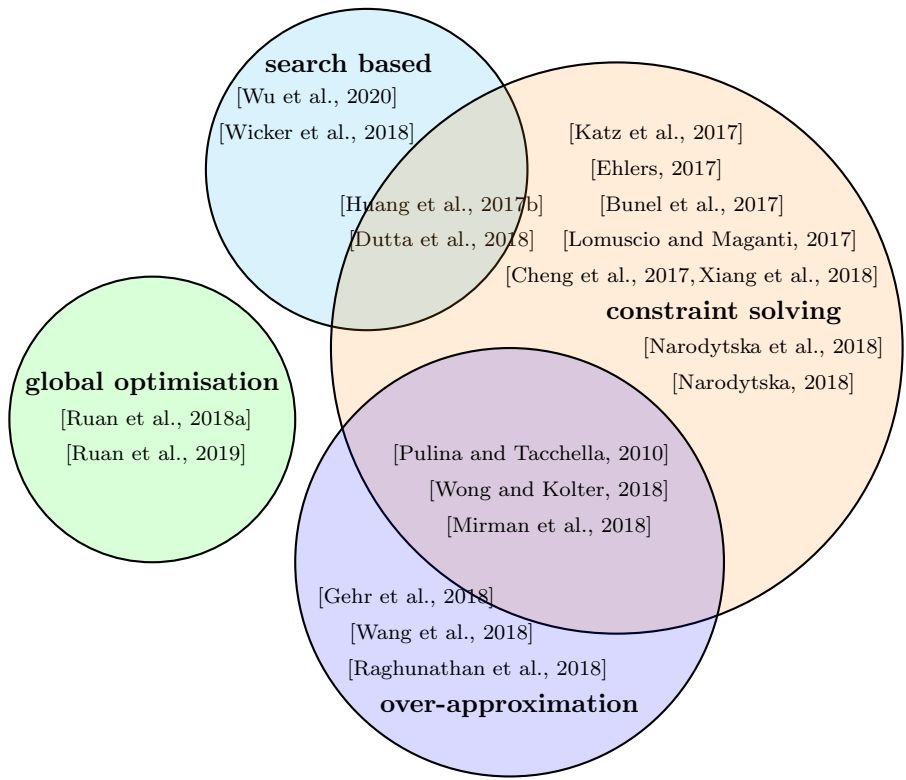
\begin{figure}[!htp]
    \centering
\begin{tikzpicture}[thick,scale=.95]
    \draw [fill=cyan, fill opacity=0.15] (0,0.5) circle (2.25cm);
    \node            (a) at (-0.25,2)       {\textbf{search based}};
    \node            (a) at (-0.75,1.5)     {\footnotesize\cite{wu2019game}};
    \node            (a) at (-0.75,1)       {\footnotesize\cite{wicker2018feature}};
    \node            (a) at (1,0)        {\footnotesize\cite{HKWW2017}};
    \node            (a) at (1,-0.5)         {\footnotesize\cite{dutta2017output}};
    
    \draw [fill=green, fill opacity=0.15] (-3,-3) circle (2cm);
    \node            (a) at (-3,-2.5)       {\textbf{global optimisation}};
    \node            (a) at (-3,-3)         {\footnotesize\cite{RHK2018}};
    \node            (a) at (-3,-3.5)       {\footnotesize\cite{ruan2019global}};

    \draw [fill=orange, fill opacity=0.15] (3.5,-2) circle (4cm);
    \node            (a) at (5,-1.5)        {\textbf{constraint solving}};
    \node            (a) at (4,1)         {\footnotesize\cite{katz2017reluplex}};
    \node            (a) at (4,0.5)         {\footnotesize\cite{Rudiger2017}};
    \node            (a) at (4.5,0)         {\footnotesize\cite{bunel2017piecewise}};
    \node            (a) at (4.75,-0.5)     {\footnotesize\cite{LM2017}};
    \node            (a) at (4.5,-1)       {\footnotesize\cite{CNR2017,xiang2017output}};
    \node            (a) at (5.5,-2)        {\footnotesize\cite{NKPSW2017}};
    \node            (a) at (5.5,-2.5)      {\footnotesize\cite{narodytska2018formal}};
    
    \draw [fill=blue, fill opacity=0.15] (2,-5) circle (3cm);
    \node            (a) at (2,-7)          {\textbf{over-approximation}};
    \node            (a) at (2.25,-3.5)     {\footnotesize\cite{PT2010}};
    \node            (a) at (2.5,-4)        {\footnotesize\cite{wong2018provable}};
    \node            (a) at (2.5,-4.5)      {\footnotesize\cite{mirman2018differentiable}};
    \node            (a) at (0.5,-5.5)        {\footnotesize\cite{GMDTCV2018}};
    \node            (a) at (1,-6)          {\footnotesize\cite{reluval}};
    \node            (a) at (1.5,-6.5)      {\footnotesize\cite{raghunathan2018certified}};
\end{tikzpicture}
    \caption{A taxonomy of verification approaches for neural networks. The classification is based on the (surveyed) core algorithms for solving the verification problem. \emph{Search based approaches} suggest that the verification algorithm is based on exhaustive search. \emph{Constraint solving} methods suggest that the neural network is encoded into a set of constraints, with the verification problem reduced into a constraint solving problem. The \emph{over-approximation} methods can guarantee the soundness of the result, but not the completeness. \emph{Global optimisation} methods suggest that the core verification algorithms are inspired by global optimisation techniques.}
    \label{fig:taxonomy-verification}
\end{figure}

In this section, we review verification techniques for neural networks. 
%
Existing approaches on the verification of networks largely fall into the following categories: \emph{constraint solving}, \emph{search based approach}, \emph{global optimisation}, and \emph{over-approximation}; note,  the separation between them may not be strict.
Figure~\ref{fig:taxonomy-verification} classifies some of the 
approaches surveyed in this paper into these categories.

In this survey, we take a different approach by classifying verification techniques with respect to the type of guarantees they can provide; these guarantee types can be:
\begin{itemize}
    \itemsep0em
    \item \emph{Exact deterministic} guarantee, which states exactly whether a property holds. We will omit the word \emph{exact} and call it deterministic guarantee in the remainder of the paper.  
    \item \emph{One-sided} guarantee, which provides either a lower bound or an upper bound to a variable, and thus can serve as a sufficient condition for a property to hold -- the variable can denote, e.g., the greatest value of some dimension in $\reach(f,\eta)$. 
    \item Guarantee with \emph{converging} lower and upper bounds to a variable.
    \item \emph{Statistical} guarantee, which quantifies the probability that a property holds.  
\end{itemize}
Note that, algorithms with one-sided guarantee and bound-converging guarantee are used to compute the real values, e.g., output reachability property (Definition~\ref{def:outputreachability}), interval property (Definition~\ref{def:intervalproperty}), or Lipschitzian property (Definition~\ref{def:globalmetric}). Their respective verification problems are based on these values, see Definitions~\ref{def:verifyreachability}, \ref{def:verifyinterval}, and \ref{def:verifyLipschitz}. 


\subsection{Approaches with Deterministic Guarantees}

\emph{Deterministic} guarantees are achieved by transforming a verification problem into a set of constraints (with or without optimisation objectives) so that they can be solved with a constraint solver. The name ``deterministic'' comes from the fact that solvers often return a deterministic answer to a query, i.e., either satisfiable or unsatisfiable. This is based on the current success of various constraint solvers such as SAT solvers, linear programming (LP) solvers, mixed integer linear programming (MILP) solvers, Satisfiability Modulo Theories (SMT) solvers.  

\subsubsection{SMT/SAT}

The Boolean satisfiability problem (SAT) determines if, given a Boolean formula, there exists an assignment to the Boolean variables such that the formula is satisfiable. Based on SAT, the satisfiability modulo theories (SMT) problem determines the satisfiability of logical formulas with respect to combinations of background theories expressed in classical first-order logic with equality. The theories we consider in the context of neural networks include the theory of real numbers and the theory of integers. For both SAT and SMT problems, there are sophisticated, open-source solvers that can automatically answer the queries about the satisfiability of the formulas.

\paragraph{An abstraction-refinement approach based on SMT solving.}

A solution to the verification of the interval property (which can be easily extended to work with the reachability property for ReLU activation functions) is proposed in \\ \cite{PT2010} by abstracting a network into a set of Boolean combinations of linear arithmetic constraints. Basically, the linear transformations between layers can be encoded directly, and the non-linear activation functions such as Sigmoid are approximated -- with both lower and upper bounds -- by piece-wise linear functions.  It is shown that whenever the abstracted model is declared to be safe, the same holds for the concrete model. Spurious counterexamples, on the other hand, trigger refinements and can be leveraged to automate the correction of misbehaviour. This approach is validated on neural networks with fewer than 10 neurons, with logistic activation function. 

\paragraph{SMT solvers for neural networks.}
Two SMT solvers $\Reluplex$~\cite{katz2017reluplex} and $\Planet$~\cite{Rudiger2017} were put forward to verify neural networks on properties expressible with SMT constraints.  SMT solvers often have good performance on problems that can be represented as a Boolean combination of constraints over other variable types. Typically, an SMT solver combines a SAT solver with specialised decision procedures for other theories. In the verification of networks, they adapt linear arithmetic over real numbers, in which an atom (i.e., the most basic expression) is of the form $\sum_{i=1}^n a_i x_i \le b$, where $a_i$ and $b$ are real numbers.



In both $\Reluplex$ and $\Planet$, they use the architecture of the Davis-Putnam-Logemann-Loveland (DPLL) algorithm in splitting cases and ruling out conflict clauses, while they differ slightly in dealing with the intersection. For $\Reluplex$, the approach inherits rules in the algorithm of Simplex and adds some rules for the $\ReLU$ operation. Through the classical pivot operation, it first looks for a solution for the linear constraints, and then applies the rules for $\ReLU$ to satisfy the $\ReLU$ relation for every node. Conversely, $\Planet$ uses linear approximation to over-approximate the neural network, and manages the condition of $\ReLU$ and max-pooling nodes with a logic formula.

\paragraph{SAT approach.}

Narodytska et al.~\cite{NKPSW2017,narodytska2018formal} propose to  verify properties of a class of neural networks (i.e., binarised neural networks) in which both weights and activations are binary, by reduction to the well-known Boolean satisfiability. Using this Boolean encoding, they leverage the power of modern SAT solvers, along with a proposed counterexample-guided search procedure, to verify various properties of these networks. A particular focus is on the robustness to adversarial perturbations. The experimental results demonstrate that this approach scales to medium-size DNN used in image classification tasks. 

\subsubsection{Mixed Integer Linear Programming (MILP)}\label{sec:milp}

Linear programming (LP) is a technique for optimising a linear objective function, subject to linear equality and linear inequality constraints. All the variables in an LP are real, if some of the variables are integers, the problem becomes a mixed integer linear programming (MILP) problem. It is noted that, while LP can be solved in polynomial time, MILP is NP-hard. 

\paragraph{MILP formulation for neural networks.}
\cite{LM2017} encodes the behaviours of fully connected neural networks with MILP. For instance, a hidden layer $z_{i+1}=\ReLU (W_i z_i+b_i)$ can be described with the following MILP:
\begin{align*}
    z_{i+1} &\ge W_i z_i + b_i,  \\
    z_{i+1} &\le W_i z_i + b_i + Mt_{i+1}, \\
    z_{i+1} &\ge 0, \\
    z_{i+1} &\le M(1-t_{i+1}),
\end{align*}
where $t_{i+1}$ has value $0$ or $1$ in its entries and has the same dimension as $z_{i+1}$, and $M>0$ is a large constant which can be treated as $\infty$. Here each integer variable in $t_{i+1}$ expresses the possibility that a neuron is activated or not. The optimisation objective can be used to express properties related to the bounds, such as $\distance{z_1 - x}{\infty}$, which expresses the $L_\infty$ distance of $z_1$ to some given input $x$. This approach can work with both the reachability property and the interval property. 

However, it is not efficient to simply use MILP to verify networks, or to compute the output range. 
In \cite{CNR2017}, a number of MILP encoding heuristics are developed to speed up the solving process, and moreover, parallelisation of MILP-solvers is used, resulting in an almost linear speed-up in the number (up to a certain limit) of computing cores in experiments.
In \cite{dutta2017output}, $\Sherlock$ alternately conducts a local and global search to efficiently calculate the output range. In a local search phase, $\Sherlock$ uses gradient descent method to find a local maximum (or minimum), while in a global search phase, it encodes the problem with MILP to check whether the local maximum (or minimum) is the global output range.

Additionally, \cite{bunel2017piecewise} presents a branch and bound (B\&B) algorithm and claims that both SAT/SMT-based and MILP-based approaches can be regarded as its special cases.

 \subsection{Approaches to Compute an Approximate Bound}


The approaches surveyed in this subsection consider the computation of a \emph{lower} (or by duality, an \emph{upper}) bound, and are able to claim the sufficiency of achieving properties. 
Whilst these approaches can only have a bounded estimation to the value of some variable, they are able to work with larger models, e.g., up to 10,000 hidden neurons. Another advantage is their potential to avoid floating point issues in existing constraint solver implementations.  
Actually, most state-of-the-art constraint solvers implementing floating-point arithmetic only give approximate solutions, which may not be the actual optimal solution or may even lie outside the feasible space~\cite{NS04}. Indeed, it may happen that a solver 
wrongly claims the satisfiability or un-satisfiability of a property. For example, \cite{dutta2017output}
reports several false positive results in $\Reluplex$, and mentions that this may come from an unsound floating point implementation.

\subsubsection{Abstract Interpretation}\label{sec:abstractInterpretation}
Abstract interpretation is a theory of sound approximation of the semantics of computer programs~\cite{CC1977}. It has been used in static analysis to verify properties of a program without it actually being run.
The basic idea of abstract interpretation is to use abstract domains (represented as e.g., boxes, zonotopes, and polyhedra) to over-approximate the computation of a set of inputs; its application has been explored in a few approaches, including $\AIsquared$~\cite{GMDTCV2018,mirman2018differentiable} and \cite{LLYCH2018}. 

Generally, on the input layer, a concrete domain ${\mathcal C}$ is defined such that the set of inputs $\eta$ is one of its elements. To enable an efficient computation, a comparatively simple domain, i.e., abstract domain ${\mathcal A}$, which over-approximates the range and relation of variables in ${\mathcal C}$, is chosen. There is a partial order $\le$ on ${\mathcal C}$ as well as ${\mathcal A}$, which is the subset relation $\subseteq$.

\begin{definition}
    A pair of functions $\alpha:{\mathcal C} \to {\mathcal A}$ and $\gamma:{\mathcal A} \to {\mathcal C}$ is a \emph{Galois connection}, if for any $a \in {\mathcal A}$ and $c \in {\mathcal C}$, we have $\alpha(c) \le a \Leftrightarrow c \le \gamma(a)$.
\end{definition}
Intuitively, a Galois connection $(\alpha,\gamma)$ expresses abstraction and concretisation relations between domains, respectively.
A Galois connection is chosen because it preserves the order of elements in two domains. Note that, $a \in \mathcal A$ is a sound abstraction of $c \in \mathcal C$ if and only if $\alpha(c) \le a$.

In abstract interpretation, it is important to choose a suitable abstract domain because it determines the efficiency and precision of the abstract interpretation.
In practice, a certain type of special shapes is used as the abstraction elements.
Formally, an abstract domain consists of shapes expressible as a set of logical constraints.
The most popular abstract domains for the Euclidean space abstraction include Interval, Zonotope, and Polyhedron; these are detailed below.

\begin{itemize}
    \itemsep0em
    \item \textbf{Interval.} An \textit{interval} $I$ contains logical constraints in the form of $a \le x_i \le b$, and for each variable $x_i$, $I$ contains at most one constraint with $x_i$.
    \item \textbf{Zonotope.} A \textit{zonotope} $Z$ consists of constraints in the form of $z_i=a_i+\sum_{j=1}^m b_{ij} \epsilon_j$, where $a_i,b_{ij}$ are real constants and $\epsilon_j \in [l_j,u_j]$. The conjunction of these constraints expresses a centre-symmetric polyhedron in the Euclidean space.
    \item \textbf{Polyhedron.} A \textit{polyhedron} $P$ has constraints in the form of linear inequalities, i.e., $\sum_{i=1}^n a_ix_i \le b$, and it gives a closed convex polyhedron in the Euclidean space.
\end{itemize}
\begin{example}\label{example:domains}
Let $\bar x \in \realnumber^2$. Assume that the range of $\bar x$ is a discrete set 
$X=\{(1,0),(0,2),(1,2),(2,1)\}$.
We can have abstraction of the input $X$ with Interval, Zonotope, and Polyhedron as follows.

\begin{itemize}
    \itemsep0em
    \item Interval: $[0,2]\times[0,2]$.
    \item Zonotope: $\{ x_1 = 1 - \frac 12 \epsilon_1 - \frac 12 \epsilon_2, \enspace x_2 = 1 + \frac 12 \epsilon_1 + \frac 12 \epsilon_3 \}$, where $\epsilon_1,\epsilon_2,\epsilon_3 \in [-1,1]$.
    \item Polyhedron: $\{ x_2 \le 2, \enspace x_2 \le - x_1 + 3, \enspace x_2 \ge x_1 - 1, \enspace x_2 \ge -2x_1 + 2 \}$.
\end{itemize}
\end{example}

The abstract interpretation based  approaches can verify interval property, but cannot verify reachability property.

\subsubsection{Convex Optimisation based Methods}

Convex optimization is to minimise convex functions over convex sets. Most neural network functions are not convex (even if it is convex, it can be a very complicated), and hence an approximation is needed for the computation based on it. 
A method is proposed in \cite{wong2018provable} to learn deep $\ReLU$-based classifiers that are provably robust against norm-bounded adversarial perturbations on the training data. The approach works with interval property, but not reachability property.
It
may flag some non-adversarial examples as adversarial examples. 
The basic idea is to consider a convex outer 
over-approximation of the set of activations reachable through a norm-bounded perturbation, and then develop a robust optimisation procedure that minimises the worst case loss over this outer region (via a linear program). Crucially, 
it is shown that the dual problem to this linear program can be represented itself as a deep network similar to the back-propagation network, leading to very efficient optimisation approaches that produce guaranteed bounds on the robust loss. 
The approach is illustrated  on a number of tasks with robust adversarial guarantees. For example, for MNIST, they produce a convolutional classifier that provably has less than 5.8\% test error for any adversarial attack with bounded $L_{\infty}$ norm less than $\epsilon=0.1$.

Moreover, \cite{DSGMK2018} works by taking a different formulation of the dual problem, i.e., applying Lagrangian relaxation on the optimisation. This is to avoid working with constrained non-convex optimisation problem.   

\subsubsection{Interval Analysis}

In \cite{reluval}, the interval arithmetic is leveraged to compute rigorous bounds on the DNN outputs, i.e., interval property. The key idea is that, given the ranges of operands, an over-estimated range of the output can be computed by using only the lower and upper bounds of the operands. Starting from the first hidden layer, this computation can be conducted through to the output layer. 
Beyond this explicit computation, symbolic interval analysis along with several other optimisations are also developed to minimise over-estimations of output bounds.  
These methods are implemented in $\ReluVal$, a system for formally checking security properties of $\ReLU$-based DNNs. 
An advantage of this approach, comparing to constraint-solving based approaches, is that it can be easily parallelisable. In general, interval analysis is close to the interval-based abstract interpretation, which we explained in Section~\ref{sec:abstractInterpretation}. 

In \cite{PRGS2017}, lower bounds of adversarial perturbations needed to alter the classification of the neural networks are derived by utilising the layer functions. The proposed bounds have theoretical guarantee that no adversarial manipulation could be any smaller, and in this case, can be computed efficiently - at most linear time in the number of (hyper)parameters of a given model and any input, which makes them applicable for choosing classifiers based on robustness.

\subsubsection{Output Reachable Set Estimation}
In \cite{xiang2017output}, the output reachable set estimation is addressed. 
Given a DNN $\networks$ with its associated function $f$, and a set of inputs $\eta$, the output reachable set is $\reach(f,\eta)$ as in Definition~\ref{def:outputreachability}.
The problem is to either compute a close estimation $\mathcal{Y}'$ such that $\reach(f,\eta)\subseteq \mathcal{Y}'$, or to determine whether $\reach(f,\eta) \cap \neg \mathcal{S} = \emptyset$ for a safety specification $\mathcal{S}$, where $\mathcal{S}$ is also expressed with a set similar as the one in Equation (\ref{equ:reachability}). Therefore, it is actually to compute the interval property. 
First, a concept called maximum sensitivity is introduced and, for a class of multi-layer perceptrons whose activation functions are monotonic functions, the maximum sensitivity can be reduced to a MILP problem as that of Section~\ref{sec:milp}. 
Then, using a simulation-based method, the output reachable set estimation problem for neural networks is formulated into a chain of optimisation problems. Finally, an automated safety verification is developed based on the output reachable set estimation result. 
The approach is applied to the safety verification for a robotic arm model with two joints.


\subsubsection{Linear Approximation of ReLU Networks}

$\FastLin$/$\FastLip$~\cite{WZCSHBDD2018} analyses the $\ReLU$ networks on both interval property and Lipschitzian property. For interval property, they consider linear approximation over those $\ReLU$ neurons that are uncertain on their status of being activated or deactivated. For Lipschitzian property, they use the gradient computation for the approximation computation. $\Crown$~\cite{zhang2018crown}  generalises the interval property computation algorithm in $\FastLin$/$\FastLip$ by allowing the linear expressions for the upper and lower bounds to be different and enabling its working with other activation functions such as tanh, sigmoid
and arctan. The Lipschitzian property computation is improved in $\RecurJac$~\cite{zhang2018recurjac}.

\subsection{Approaches to Compute Converging  Bounds}\label{sec:covergingbounds}

While the above approaches can work with small networks (up to a few thousands hidden neurons), state-of-the-art DNNs usually contain at least multi-million hidden neurons. It is necessary that other approaches are developed to work with real-world systems. In Section~\ref{sec:covergingbounds} and Section~\ref{sec:statisticalguarantee}, the approaches are able to work with large-scale networks, although they might have other restrictions or limitations. Since the approaches surveyed in this subsection compute converging upper and lower bounds, they can work with both output reachability  property and interval property. 

\subsubsection{Layer-by-Layer Refinement}
\cite{HKWW2017} develops an automated verification framework for feedforward multi-layer neural networks based on Satisfiability Modulo Theory (SMT). The key features of this framework are that it \emph{guarantees} a misclassification being found if it exists, and that it propagates the analysis \emph{layer-by-layer}, i.e., from the input layer to, in particular, the hidden layers, and to the output layer.  

In this work, \emph{safety} for an individual classification decision, i.e., pointwise (or local) robustness, is defined as the invariance of a classifier's outcome to perturbations within a small neighbourhood of an original input. Formally, $$ \networks, \region_k, \manipulation_k \satisfy \inputPoint $$ where $\inputPoint$ denotes an input, $\networks$ a neural network, $\region_k$ a region surrounding the corresponding activation of the input $x$ at layer $k$, and $\manipulation_k$ a set of manipulations at layer $k$. Later, in \cite{wicker2018feature,wu2019game}, it is shown that the minimality of the manipulations in $\manipulation$ can be guaranteed with the existence of Lipschitz constant. 

To be more specific, its verification algorithm uses single-/multi-path search to exhaustively explore a finite region of the vector spaces associated with the input layer or the hidden layers, and a layer-by-layer refinement is implemented using the Z3 solver to ensure that the local robustness of a deeper layer implies the robustness of a shallower layer.
The methodology is implemented in the software tool~$\DLV$, and evaluated on image benchmarks such as MNIST, CIFAR10, GTSRB, and ImageNet. Though the complexity is high, it scales to work with state-of-the-art networks such as VGG16. 
Furthermore, in \cite{wicker2018feature,wu2019game}, the search problem is alleviated by Monte-Carlo tree search.

\subsubsection{Reduction to A Two-Player Turn-based Game}

\newcommand{\maximumsaferadius}{{\tt MSR}}
\newcommand{\featurerobustness}{{\tt FR}}
\newcommand{\playerOne}{{\tt I}}
\newcommand{\playerTwo}{{\tt II}}

In $\DeepGame$~\cite{wu2019game}, two variants of pointwise robustness are studied: 
\begin{itemize}
    \itemsep0em    
    \item the \emph{maximum safe radius} ($\maximumsaferadius$) problem, which for a given input sample computes the minimum distance to an adversarial example, and
    \item the \emph{feature robustness} ($\featurerobustness$) problem, which aims to quantify the robustness of individual features to adversarial perturbations.
\end{itemize}

It demonstrates that, under the assumption of Lipschitz continuity, both problems can be approximated using finite optimisation by discretising the input space, and the approximation has provable guarantees, i.e., the error is bounded. 
It subsequently reduces the resulting optimisation problems to the solution of a two-player turn-based game, where Player~$\playerOne$ selects features and Player~$\playerTwo$ perturbs the image within the feature. While Player~$\playerTwo$ aims to minimise the distance to an adversarial example, depending on the optimisation objective Player~$\playerOne$ can be \emph{cooperative} or \emph{competitive}. An anytime approach is employed to solve the games, in the sense of approximating the value of a game by monotonically improving its upper and lower bounds. The Monte-Carlo tree search algorithm is applied to compute upper bounds for both games, and the Admissible A* and the Alpha-Beta Pruning algorithms are, respectively, used to compute lower bounds for the $\maximumsaferadius$ and $\featurerobustness$ games.

\subsubsection{Global Optimisation Based Approaches}


$\DeepGO$~\cite{RHK2018} shows that most known layers of DNNs are Lipschitz continuous, and presents a verification approach based on global optimisation. For a single dimension, an algorithm is presented to always compute the lower bounds (by utilising the Lipschitz constant) and eventually converge to the optimal value. Based on this single-dimensional algorithm, the algorithm for multiple dimensions is to exhaustively search for the best combinations. The algorithm is able to work with state-of-the-art DNNs, but is restricted by the number of dimensions to be perturbed.  

In $\TRE$~\cite{ruan2019global}, the authors focus on the Hamming distance, and study the problem of quantifying the global robustness of a trained DNN, where global robustness is defined as the expectation of the maximum safe radius over a testing dataset. They propose an approach to iteratively generate lower and upper bounds on the network's robustness. The approach is anytime, i.e., it returns intermediate bounds and robustness estimates that are gradually, but strictly, improved as the computation proceeds; tensor-based, i.e., the computation is conducted over a set of inputs simultaneously, instead of one by one, to enable efficient GPU computation; and has provable guarantees, i.e., both the bounds and the robustness estimates can converge to their optimal values. 


\subsection{Approaches with Statistical Guarantees}\label{sec:statisticalguarantee}

This subsection reviews a few approaches aiming to achieve statistical guarantees on their results, by claiming e.g., the satisfiability of a property, or a value is a lower bound of another value, etc., with certain probability.

\subsubsection{Lipschitz Constant Estimation by Extreme Value Theory}

\cite{WZCYSGHD2018} proposes a metric, called $\CLEVER$, to estimate the Lipschitz constant, i.e., the approach works with Lipschitzian property. It estimates the robustness lower bound by sampling the norm of gradients and fitting a limit distribution using extreme value theory. However, as argued by \cite{Goodfellow2018}, their evaluation approach can only find statistical approximation of the lower bound, i.e., their approach has a soundness problem. 

\subsubsection{Robustness Estimation} 
\cite{BILVNC2016} proposes two statistics of robustness to measure the frequency and the severity of adversarial examples, respectively. Both statistics are based on a parameter $\epsilon$, which is the maximum radius within which no adversarial examples exist. 
The computation of these statistics is based on the local linearity assumption which holds when $\epsilon$ is  small enough. Except for the application of the $\ReLU$ activation function which is piece-wise linear, this assumption can be satisfied by the existence of the Lipschitz constant as shown in \cite{RHK2018}.


\subsection{Computational Complexity of Verification} 

There are two ways to measure the complexity of conducting formal verification. The first, appeared in \cite{katz2017reluplex}, measures the complexity with respect to the number of hidden neurons. This is due to the fact that their approach is to encode the DNN into a set of constraints, and in the constraints every hidden neuron is associated with two variables. On the other hand, in \cite{RHK2018}, the complexity is measured with respect to the number of input dimensions. This is due to the fact that their approach is to manipulate the input. 
For both cases, the complexity is shown NP-complete, although it is understandable that the number of hidden neurons can be larger than the number of input dimensions.

\subsection{Summary}

We summarise some existing approaches to the verification of DNNs in Table~\ref{tab:verification}, from the aspects of the type of achievable guarantees, underlying algorithms, and objective properties, i.e., robustness, reachability, interval, and Lipschitzian.

\renewcommand{\arraystretch}{1.5}

\begin{table}[h]
    \caption{Comparison between the verification approaches of deep neural networks}
    \label{tab:verification}
\scalebox{0.56}{
    \begin{tabular}{|c||c|c|c|c|c|c|c|}
        \toprule
        \multirow{2}{*}{} & \multirow{2}{*}{\textbf{Guarantees}} & \multicolumn{2}{c|}{\multirow{2}{*}{\textbf{Algorithm}}} & \multicolumn{4}{c|}{\textbf{Property}} \\ \cline{5-8}
         & & \multicolumn{1}{c}{} & & \emph{Robustness} & \emph{Reachability} & \emph{Interval} & \emph{Lipschitzian} \\ \hline

        \cite{PT2010} & \multirow{8}{*}{\makecell{Deterministic\\Guarantees}} & \multirow{4}{*}{\makecell{Constraints\\Solving}} & \multirow{3}{*}{SMT} & \checkmark & \checkmark & \checkmark & \\
        \cite{katz2017reluplex} & & & & \checkmark & \checkmark & \checkmark & \\
        \cite{Rudiger2017} & & & & \checkmark & \checkmark & \checkmark & \\ \cline{4-8}
        \makecell{\cite{NKPSW2017}\\\cite{narodytska2018formal}} & & & SAT & \checkmark & \checkmark & \checkmark & \\ \cline{3-8}
        \cite{LM2017} & & \multicolumn{2}{c|}{\multirow{4}{*}{MILP}} & \checkmark & \checkmark & \checkmark & \\
        \cite{CNR2017} & & \multicolumn{1}{c}{} & & \checkmark & \checkmark & \checkmark &  \\
        \cite{dutta2017output} & & \multicolumn{1}{c}{} & & \checkmark & \checkmark & \checkmark & \\
        \cite{bunel2017piecewise} & & \multicolumn{1}{c}{} & & \checkmark & \checkmark & \checkmark & \\
        \midrule
        \cite{GMDTCV2018} & \multirow{9}{*}{\makecell{Lower/Upper\\Bound}} & \multicolumn{2}{c|}{\multirow{3}{*}{\makecell{Abstract\\Interpretation}}} & \checkmark & & \checkmark & \\
        \cite{mirman2018differentiable} & & \multicolumn{1}{c}{} &  & \checkmark & & \checkmark & \\
        \cite{LLYCH2018} & & \multicolumn{1}{c}{} & & \checkmark & & \checkmark & \\ \cline{3-8}
        \cite{wong2018provable} & & \multicolumn{2}{c|}{\makecell{Convex Optimisation}} & \checkmark & & \checkmark & \\ \cline{3-8}
        \cite{reluval} & & \multicolumn{2}{c|}{\multirow{2}{*}{\makecell{Interval Analysis}}} & \checkmark & & \checkmark & \\
        \cite{PRGS2017} & & \multicolumn{1}{c}{} & & \checkmark & & \checkmark & \\ \cline{3-8}
        \cite{xiang2017output} & & \multicolumn{2}{c|}{\makecell{Set Estimation}} & \checkmark & & \checkmark & \\ \cline{3-8}
        \cite{WZCSHBDD2018} & & \multicolumn{2}{c|}{\makecell{Linear Approximation}} & \checkmark & & \checkmark & \checkmark \\
        \midrule
        \cite{HKWW2017} & \multirow{5}{*}{\makecell{Converging\\Bounds}} & \multirow{3}{*}{\makecell{Search\\Based}} & \makecell{Layer-by-Layer\\Refinement} & \checkmark & \checkmark & \checkmark & \\ \cline{4-8} 
        \cite{wicker2018feature} &  & & \multirow{2}{*}{\makecell{Two-Player\\Turn-based Game}} & \checkmark & \checkmark & \checkmark & \checkmark \\
        \cite{wu2019game} & & & & \checkmark & \checkmark & \checkmark & \checkmark \\ \cline{3-8}
        \cite{RHK2018} & & \multicolumn{2}{c|}{\multirow{2}{*}{\makecell{Global\\Optisimation}}} & \checkmark & \checkmark & \checkmark & \checkmark \\
        \cite{ruan2019global} & & \multicolumn{1}{c}{} & & \checkmark & \checkmark &  \checkmark &  \\
        \midrule
        \cite{WZCYSGHD2018} & \multirow{2}{*}{\makecell{Statistical\\Guarantees}} & \multicolumn{2}{c|}{\makecell{Extreme Value Theory}} & & & & \checkmark \\ \cline{3-8}
        \cite{BILVNC2016} & & \multicolumn{2}{c|}{\makecell{Robustness Estimation}} & \checkmark & & & \\ 
        \bottomrule
        \end{tabular}
}
\end{table}


\cleardoublepage

\newenvironment{proof}{\noindent \textbf{Proof:}}{\hfill  $\boxempty$}

\newcommand\true{\textsf{true}\xspace}
\newcommand\false{\textsf{false}\xspace}

\newtheorem{theorem}{Theorem}
\newtheorem{proposition}{Proposition}
\newtheorem{lemma}{Lemma}

\section{Testing}\label{sec:testing}

Similar to traditional software testing against software verification, DNN testing provides a certification methodology with a balance between completeness  and efficiency. In established industries, e.g., avionics and automotive, the needs for software testing have been settled in various standards such as DO-178C and MISRA. However, due to the lack of logical structures and system specification, it is still unclear how to extend such standards to work with systems with DNN components. In the following, we survey testing techniques from three aspects: coverage criteria (Section~\ref{sec:test-criteria}), test case generation (Section~\ref{sec:test-gen}), and model-level mutation testing (Section~\ref{sec:mutationTesting}). The first two do not alter the structure of the DNN,  while the mutation testing involves the change to the structure and parameters of the DNN. 

\subsection{Coverage Criteria for DNNs}
\label{sec:test-criteria}

Research in software engineering has resulted in a broad range of approaches
to testing software. Please refer to \cite{ZHM1997,JH2011,SWMPHS2017}  for comprehensive reviews.
In white-box testing, the structure of a
program is exploited to (perhaps automatically) generate test cases.
Structural coverage criteria (or metrics) define a set of test objectives to be covered, guiding the generation of test cases and evaluating the completeness of a test suite. 
E.g., a test suite with 100\% statement coverage exercises all statements of the program at
least once.  While it is arguable whether this ensures functional correctness,
high coverage is able to increase users' confidence (or trust) in the testing results~\cite{ZHM1997}.  Structural coverage analysis and testing are also
used as a means of assessment in a number of safety-critical scenarios, and
criteria such as statement and modified condition/decision coverage (MC/DC) are
applicable measures with respect to different criticality levels.  MC/DC was developed by NASA\cite{HVCR2001} and has been widely adopted.  It is used in avionics software development guidance to ensure adequate testing of applications with the highest criticality \cite{do178}.

We let $\setofcoveringmethods$ be a set of covering methods, and $\requirements=\neuronpairs(\networks)$ be the set of test objectives to be covered.
In \cite{sun2018testing-b} $\neuronpairs(\networks)$ is instantiated as the set of causal relationships between feature pairs, while
in \cite{PCYJ2017} $\neuronpairs(\networks)$ is instantiated as the set of statuses of hidden neurons.

 

\begin{definition}[Test Suite]
Given a DNN $\networks$, a test suite $\testsuites$ is a finite set of input vectors, i.e., $\testsuites\subseteq D_{1}\times D_{1}\times ... \times D_{1}$. Each vector is called a test case. 
\end{definition}

Usually, a test case is a single input $\testsuites\subseteq D_{1}$, e.g., in \cite{PCYJ2017}, or a pair of inputs $\testsuites\subseteq D_{1}\times D_{1}$, e.g., in \cite{sun2018testing-b}. Ideally, given the set of test objectives
$\requirements$ according to some covering method $\coveringmethod$, 
we run a test case generation algorithm to find a test suite $\testsuites$ such that %
  \begin{equation}
  \forall \neuronpair\in \requirements \exists (x_1,x_2,...,x_k) \in \testsuites: \coveringmethod(\neuronpair,(x_1,x_2,...,x_k))
 \end{equation} 
where $\coveringmethod(\neuronpair,(x_1,x_2,...,x_k))$ intuitively means that the test objective $\neuronpair$ is satisfied under the test case $(x_1,x_2,...,x_k)$. 
In practice, we might want to
compute the degree to which the test objectives are satisfied by a generated test suite $\testsuites$.

\begin{definition}[Test Criterion]
Given a DNN $\networks$ with its associated function $f$, a covering method $\coveringmethod$, test objectives in $\requirements$, 
and a test suite $\testsuites$,  the test criterion $\metric_{\coveringmethod}(\requirements,\testsuites)$ is as follows: 
\begin{equation}
  \label{eq:madc}
  \metric_{\coveringmethod}(\requirements,\testsuites)=\frac{|\{\neuronpair \in \requirements | \exists  (x_1,x_2,...,x_k) \in \testsuites: \coveringmethod(\neuronpair,(x_1,x_2,...,x_k))\}|}{|\requirements|}
\end{equation}
\end{definition}
 
Intuitively, it computes the percentage of the test objectives that are
covered by test cases in $\testsuites$ w.r.t. the covering method~$\coveringmethod$. 

\subsubsection{Neuron Coverage}

Neuron coverage \cite{PCYJ2017} 
can be seen as the statement coverage variant for DNN testing. \added{Note that, the neuron coverage is primarily designed for ReLU networks, although an easy adaptation can be applied to make it work with other activation functions.}

\begin{definition} \cite{PCYJ2017}
A node $n_{k,i}$ is neuron covered by a test case $x$, denoted as $\covered{N}{}(n_{k,i},x)$, if $\mathit{sign}(n_{k,i},x)= +1$. 
\end{definition}

The set of objectives to be covered is  $\neuronpairs(\networks)=\{n_{k,i}~|~2\leq k\leq K-1, 1\leq i\leq s_k\}$. Each test case is a single input, i.e., $\testsuites \subseteq D_{L_1}$. The covering method is as follows:  $\coveringmethod(n_{k,i},x')$ if and only if $\covered{N}{}(n_{k,i},x')$. 


A search algorithm DeepXplore \cite{PCYJ2017} is developed
to generate test cases for neuron coverage. It takes multiple DNNs $\{f^k~|~k\in 1..n\}$ as the input and maximises over both the number of observed differential behaviours and the neuron coverage while preserving domain-specific constraints provided by the users. Let $f^k_c(x)$ be the class probability that $f^k$ predicts $x$ to be $c$. The optimisation objective for a neuron $n_{j,i}$ is as follows.
\begin{equation}
obj(x,n_{j,i})=(\sum_{k\neq j}f^k_c(x)-\lambda_1 f^j_c(x)) + \lambda_2 v_{j,i}(x)
\end{equation}
where $\lambda_1$ and $\lambda_2$ are a tunable parameters, $\sum_{k\neq j}f^k_c(x)-\lambda_1 f^j_c(x)$ denotes differential behaviours, and $v_{j,i}(x)$ is the activation value of neuron $n_{j,i}$ on $x$. 

Moreover,  the greedy search combined with image transformations is used in \cite{tian2017deeptest}
to increase neuron coverage, and is  applied to DNNs for autonomous driving. In \cite{sun2018concolic}, the conclic testing algorithm -- a software analysis technique combining symbolic execution and concrete execution -- can work with a set of coverage metrics including neuron coverage and some other coverage metrics to be surveyed in the following.

\subsubsection{Safety Coverage}
In \cite{wicker2018feature}, the input space is discretised with
a set of hyper-rectangles, and then one test case is generated for each
hyper-rectangle. 

\begin{definition}
\label{def:safety-cover}
Let each hyper-rectangle $rec$ contain those inputs with the same pattern
of ReLU, i.e., for all $x_1,x_2\in rec$, $2\leq k\leq K-1$ and $1\leq l\leq s_k$, we have
$\mathit{sign}(n_{k,l},x_1)=\mathit{sign}(n_{k,l},x_2)$.  A~hyper-rectangle $rec$ is safe covered by a test
case $x$, denoted as $\covered{S}{}(rec,x)$, if $x\in rec$.
\end{definition}

Let $Rec(\networks)$ be the set of hyper-rectangles. 
%
The set of objectives to be covered is  $\neuronpairs(\networks)= Rec(\networks)$. Each test case is a single input, i.e., $\testsuites \subseteq D_{L_1}$. The covering method is as follows:  $\coveringmethod(rec,x)$ if and only if $\covered{S}{}(rec,x)$. 

Moreover, there are different ways to define the set of hyper-rectangles. For example, the ``boxing clever'' method in \cite{box-clever}, initially proposed for designing
training datasets, divides the input space into a series of representative boxes. 
When the hyper-rectangle is sufficiently fine-grained with respect to Lipschitz constant of the DNN, the method in \cite{wicker2018feature}
becomes exhaustive search and has provable guarantee on its result. In terms of the test case generation algorithm, it uses Monte Carlo tree search to exhaustively enumerate for each hyper-rectangle a test case. 
%


\subsubsection{Extensions of Neuron Coverage}

In \cite{ma2018deepgauge}, several coverage criteria are proposed, following similar
rationale as neuron coverage and focusing on individual neurons' activation values.  

\begin{definition}  \cite{ma2018deepgauge}
A node $n_{k,i}$ is neuron boundary covered by a test case $x$, denoted as $\covered{NB}{}(n_{k,i},x)$, if $v_{k,i}[x] > v_{k,i}^u$. 
\end{definition}

Let $rank(n_{k,i},x)$ represent the rank of the value $v_{k,i}[x]$ among those values $\{v_{k,j}[x]~|~1\leq j\leq s_k\}$ of the nodes at the same layer, i.e., there are $rank(n_{k,i},x)-1$ elements in $\{v_{k,j}[x]~|~1\leq j\leq s_k\}$ which are greater than $v_{k,i}[x]$. 
\begin{definition}  \cite{ma2018deepgauge}
For $1\leq m\leq s_k$, a node $n_{k,i}$ is top-$m$ neuron covered by  $x$, denoted as $\covered{TN}{m}(n_{k,i},x)$, if $rank(n_{k,i},x) \leq m$. 
\end{definition}

Let $v_{k,i}^l=\min_{x\in X}v_{k,i}[x]$ and $v_{k,i}^u=\max_{x\in X}v_{k,i}[x]$ for some input $x$. We can split the interval $I_{k,i} = [v_{k,i}^l,v_{k,i}^u]$ into $m$ equal sections, and let $I_{k,i}^j$ be the $j$th section. 

\begin{definition}  \cite{ma2018deepgauge}
Given $m\geq 1$, a node $n_{k,i}$ is $m$-multisection neuron covered by  a test suite $\testsuites$, denoted as $\covered{MN}{m}(n_{k,i},\testsuites)$, if $\forall 1\leq j\leq m\exists x\in \testsuites: v_{k,i}[x] \in I_{k,i}^j$, i.e., all sections are covered by some test cases. 
\end{definition}

Each test case is a single input, i.e., $\testsuites \subseteq D_{L_1}$. We omit the definition of test objectives  $\neuronpairs$ and covering methods $\coveringmethod$,
which are similar to the original neuron coverage case.




No particular algorithm is developed
in \cite{ma2018deepgauge} for generating test cases for the criteria proposed; instead, they  apply
adversarial attack methods (e.g., \cite{FGSM}) to generate an extra set of new inputs that is shown to
increase the coverage. Following \cite{ma2018deepgauge}, an exploratory study on combinatorial testing
is conducted in \cite{ma2018combinatorial} to cover combinations of neurons' activations at the same layer. 


\subsubsection{Modified Condition/Decision Coverage (MC/DC)}

\emph{Modified Condition/Decision Coverage} (MC/DC)~\cite{HVCR2001} is a
method of ensuring adequate testing for safety-critical software.  At its
core is the idea that if a choice can be made, all the possible factors
(conditions) that contribute to that choice (decision) must be tested.  For
traditional software, both conditions and the decision are usually Boolean
variables or Boolean expressions. 
\begin{example}\label{example:mcdc}
The decision
\begin{equation}\label{eq:mcdc}
d \iff ((a > 3) \lor (b = 0) ) \land (c \not= 4)
\end{equation}
contains the three conditions $(a>3)$, $(b=0)$ and $(c\not=4)$.  The
following  test cases provide 100\% MC/DC coverage:
\begin{enumerate}
\item $(a>3)$=\false, $(b=0)$=\true,  $(c \not= 4)$=\false
\item $(a>3)$=\true,  $(b=0)$=\false, $(c \not= 4)$=\true
\item $(a>3)$=\false, $(b=0)$=\false, $(c \not= 4)$=\true
\item $(a>3)$=\false, $(b=0)$=\true,  $(c \not= 4)$=\true
\end{enumerate}
The first two test cases already satisfy both  \emph{condition coverage}
(i.e., all possibilities of the conditions  are exploited) and 
\emph{decision coverage} (i.e., all possibilities of the decision  are
exploited).  The other two cases are needed because, for MC/DC each
condition should evaluate to \true and \false at least once, and should
independently affect the decision outcome
\end{example}


Motivated by the MC/DC testing for traditional software, an MC/DC variant for DNNs are initially
proposed in \cite{sun2018testing,sun2018testingjournal}, which is further refined in \cite{sun2018testing-b}. Different
from the criteria in \cite{PCYJ2017,ma2018deepgauge} that only consider individual neurons' activations,
the criteria in \cite{sun2018testing,sun2018testingjournal,sun2018testing-b} take into account the causal relation between
features in DNNs: \emph{the core idea is to ensure that not only the presence of a feature needs to be
tested but also the effects of less complex features on a more complex feature must be tested}.

We let $\setoffeatures_k$ be a collection of subsets of nodes at layer $k$. 
Without loss of generality, each element of $\setoffeatures_k$, i.e., a subset of nodes in the $k$-th layer, represents a \emph{feature} learned at layer $k$.

At first, different from the Boolean case, where changes of conditions and decisions are straightforwardly switches of \true/\false values,
the change observed on a feature can be either a sign change or a value change.

\begin{definition}[Sign Change]\label{def:sc}
Given a feature $\feature_{k,l}$ and two test cases $x_1$ and $x_2$, 
the
sign change of $\feature_{k,l}$ is exploited by $x_1$ and $x_2$, denoted as
$sc(\feature_{k,l},x_1,x_2)$, if
\begin{itemize}
    \item $\mathit{sign}(n_{k,j},x_1) \neq
\mathit{sign}(n_{k,j},x_2)$ for all $n_{k,j}\in\feature_{k,l}$.
\end{itemize}  
Moreover, we~write $ nsc(\feature_{k,l},x_1,x_2)$ if
\begin{itemize}
    \item $\mathit{sign}(n_{k,j},x_1) =
\mathit{sign}(n_{k,j},x_2)$ for all $n_{k,j}\in\feature_{k,l}$.
\end{itemize}
\end{definition}

Note that $nsc(\feature_{k,l},x_1,x_2) \neq \neg sc(\feature_{k,l},x_1,x_2)$. When the ReLU activation function is assumed,
the sign change of a feature represents switch of the two cases, in which neuron activations of this feature are and are not propagated
to the next layer.

A feature's sign change is sometimes too restrictive and its value change compensates this. We can denote a value function as $g: \setoffeatures_k \times D_{L_1}\times D_{L_1}\rightarrow \{\true,\false\}$. Simply speaking, it  expresses the DNN developer's intuition (or knowledge) on what contributes as a significant change on the feature $\feature_{k,l}$, by specifying the difference between two vectors $\feature_{k,l}[x_1]$ and $\feature_{k,l}[x_2]$.  
%
The following are a few examples. 

\begin{example}
For a singleton set $\feature_{k,l}=\{n_{k,j}\}$, the  function  $\valuefunction(\feature_{k,l}, x_1, x_2) $ can express 
e.g., $|u_{k,j}[x_1] - u_{k,j}[x_2]| \geq d$ (absolute change), or
$\frac{u_{k,j}[x_1]}{u_{k,j}[x_2]} > d \lor
\frac{u_{k,j}[x_1]}{u_{k,j}[x_2]} < 1/d$ (relative change), etc.
It~can also express the constraint on one of the values $u_{k,j}[x_2]$ such
as $u_{k,j}[x_2] > d $ (upper boundary).
\end{example}
\begin{example}
For the general case, the  function $\valuefunction(\feature_{k,l}, x_1, x_2) $ can express the distance between two vectors  $\feature_{k,l}[x_1]$ and $\feature_{k,l}[x_2]$ by e.g., norm-based distances $\distance{\feature_{k,l}[x_1] -
\feature_{k,l}[x_2]}{p}  \leq d$ for a real number $d$ and a distance measure
$L^p$, or structural similarity distances such as SSIM \cite{WSB2003}. It can also express
constraints between nodes of the same layer such as 
$\bigwedge_{j\neq i}v_{k,i}[x_1] \geq v_{k,j}[x_1]$. 
\end{example}

Consequently, the value change of a feature is defined as follows. 

\begin{definition}[Value Change]\label{def:vc}
Given a feature $\feature_{k,l}$, two test cases $x_1$ and $x_2$, and a value function $\valuefunction$,  
the value change of $\feature_{k,l}$ is exploited by $x_1$ and $x_2$ with respect to 
$\valuefunction$, denoted as
$vc(\valuefunction,\feature_{k,l},x_1,x_2)$, if 
\begin{itemize}
    \item $\valuefunction(\feature_{k,l}, x_1,x_2)$=\true.
\end{itemize}
Moreover, we write $\neg
vc(\valuefunction,\feature_{k,l},x_1,x_2)$ when the condition is not satisfied.
\end{definition}

Based on the concept of sign changes and value changes, a family of four coverage criteria are
proposed in \cite{sun2018testing,sun2018testingjournal,sun2018testing-b}, i.e., the MC/DC variant for DNNs, to exploit
the causal relationship between the changes of features at consecutive layers of the neural network.

\begin{definition}[Sign-Sign Coverage, or SS Coverage]
  \label{def:ssc}
  A feature pair $\neuronpair=(\feature_{k,i},\feature_{k+1,j})$ is SS-covered by two test cases
  $x_1, x_2$, denoted as $\covered{SS}{}{(\neuronpair,x_1,x_2)}$, if the following
  conditions are satisfied by the  DNN instances $\networks[x_1]$ and $\networks[x_2]$: 
  \begin{itemize}
    \item $sc(\feature_{k,i},x_1,x_2)$ and  $nsc(P_k\setminus \feature_{k,i},x_1,x_2)$;
    \item $sc(\feature_{k+1,j},x_1,x_2)$.
  \end{itemize}
  where $P_k$ is the set of nodes in layer $k$.  
\end{definition}

\begin{definition}[Sign-Value Coverage, or SV Coverage]
\label{def:svc}
Given a value function $\valuefunction$, a feature pair
$\neuronpair=(\feature_{k,i},\feature_{k+1,j})$ is SV-covered by two test cases $x_1,
x_2$, denoted as $\covered{SV}{\valuefunction}{(\neuronpair,x_1,x_2)}$, if
the following conditions are satisfied by the DNN instances
$\networks[x_1]$ and $\networks[x_2]$:
  \begin{itemize}
    \item $sc(\feature_{k,i},x_1,x_2)$and  $nsc(P_k \setminus \feature_{k,i},x_1,x_2)$;
    \item $vc(\valuefunction,\feature_{k+1,j},x_1,x_2)$ and $nsc(\feature_{k+1,j},x_1,x_2)$.
  \end{itemize}
\end{definition}

\begin{definition}[Value-Sign Coverage, or VS Coverage]
  \label{def:dsc}
  Given a value function $\valuefunction$, a feature pair
  $\neuronpair=(\feature_{k,i},\feature_{k+1,j})$ is VS-covered by two test cases 
  $x_1, x_2$, denoted as $\covered{VS}{\valuefunction}{(\neuronpair,x_1,x_2)}$,
  if the following conditions are satisfied by the DNN instances $\networks[x_1]$
  and $\networks[x_2]$: 
  \begin{itemize}
    \item $vc(\valuefunction,\feature_{k,i},x_1,x_2)$ and \substituted{$nsc(L_{k},x_1,x_2)$}{$nsc(P_{k},x_1,x_2)$};
    \item $sc(\feature_{k+1,j},x_1,x_2)$.
  \end{itemize}
\end{definition}

\begin{definition}[Value-Value Coverage, or VV Coverage]
  \label{def:dvc}
 Given two value functions $\valuefunction_1$ and $\valuefunction_2$, a feature pair $\neuronpair=(\feature_{k,i},\feature_{k+1,j})$ is VV-covered by two test cases $x_1, x_2$, denoted as $\covered{VV}{\valuefunction_1,\valuefunction_2}{(\neuronpair,x_1,x_2)}$, if the following conditions are satisfied by the  DNN instances $\networks[x_1]$ and $\networks[x_2]$: 
  \begin{itemize}
    \item $vc(\valuefunction_1, \feature_{k,i},x_1,x_2)$ and \substituted{$nsc(L_{k},x_1,x_2)$}{$nsc(P_{k},x_1,x_2)$};
    \item $vc(\valuefunction_2,\feature_{k+1,j},x_1,x_2)$ and $nsc(\feature_{k+1,j},x_1,x_2)$.
  \end{itemize}
\end{definition}

For all the above, each test case is a pair of inputs, i.e., $\testsuites \subseteq D_{L_1}\times D_{L_1}$. The test objectives  $\neuronpairs$ is a set of feature pairs, provided by the user or computed automatically according to the structure of the DNN. The covering methods $\coveringmethod$ has been defined in the above definitions.


For the test case generation, \cite{sun2018testing,sun2018testingjournal} develops an algorithm based on liner programming (LP). This is complemented with an adaptive gradient descent (GD) search algorithm in \cite{sun2018testing-b} and a concolic testing algorithm in \cite{sun2018concolic}.

\subsubsection{Quantitative Projection Coverage}
In \cite{huang-atva18}, it is assumed that there exist a number of weighted criteria for describing
the operation conditions. For example, for self-driving cars, the criteria can be based on e.g., weather, landscape, partially occluding pedestrians, etc. With these criteria one can systematically partition the input domain and weight each partitioned class based
on its relative importance. Based on this, the quantitative $k$-projection is  such that the
data set, when being projected onto the $k$-hyperplane, needs to have (in each region) data points no
less than the associated weight. 
While the criteria in \cite{huang-atva18} are based on self-driving scenes,  \cite{huang18b} present
a few further criteria  that take into account DNN internal structures, 
focusing on individual neurons' or neuron sets' activation values.

In terms of the test case generation,  a method based on 0-1 Integer Linear Programming is developed.
It has been integrated into the nn-dependability-kit~\cite{cheng2018nn}.



\subsubsection{Surprise Coverage}
\cite{kim2018guiding} aims to measure the relative novelty
(i.e., surprise) of the test inputs with respect to the
training dataset, by measuring the difference of activation patterns \cite{sun2018testing,sun2018testingjournal} between inputs. 
Given a training set $\trainingDataset\subseteq D_{L_1}$, a layer $k$, and a new input $x$, one of the measurements is to compute the following value
\begin{equation}\label{equ:surprising}
-\log(\frac{1}{|\trainingDataset|}\sum_{x_i\in \trainingDataset}{\cal K}_H(v_k(x)-v_k(x_i)))
\end{equation}
where $v_k(x)$ is the vector of activation values for neurons in layer $k$ when the input is $x$, and $|\trainingDataset|$ is the number of training samples. Moreover, ${\cal K}$ is a Gaussian kernel function and $H$ is a bandwidth matrix, used in Kernel Density Estimation~\cite{WJ1994}. Based on this, the coverage is defined, similar as the m-multisection \cite{ma2018deepgauge}, to cover a few pre-specified segments within a range $(0,U]$. 
%
%
Intuitively, a good test input set for a
DNN should be systematically diversified to include
inputs ranging from those similar to training data (i.e., having lower values for Expression (\ref{equ:surprising})) to those
significantly different (i.e., having higher values for Expression (\ref{equ:surprising})). 


In terms of test case generation, \cite{kim2018guiding} utilises a few existing algorithms for adversarial attack, including FGSM \cite{FGSM}, Basic iterative method \cite{KGB2016}, JSMA \cite{JSMA}, CW attack \cite{carlini2017adversarial}.

\subsubsection{Comparison between Existing Coverage Criteria}

\begin{figure}[h]
    \centering
    \includegraphics[scale=0.3]{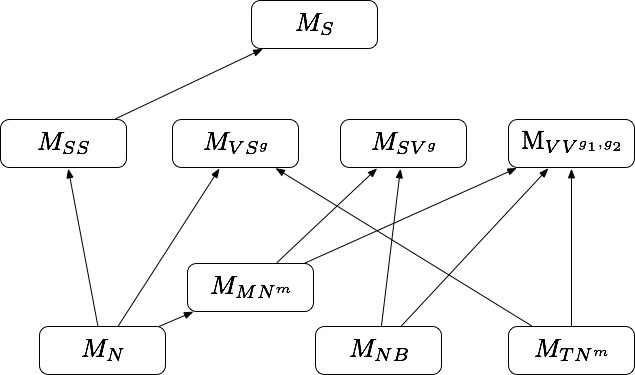}
    \caption{Relationship between test criteria}
    \label{fig:relationship}
\end{figure}

Figure~\ref{fig:relationship} gives a diagrammatic illustration of the relationship between some coverage criteria we survey above. An arrow from $A$ to $B$ denotes that the coverage metric $A$ is weaker than the  coverage metric $B$. We say that 
metric $M_{\coveringmethod_1}$ is weaker than another metric $M_{\coveringmethod_2}$, if for any given test suite $\testsuites$ on 
$\networks$, we have that $M_{\coveringmethod_1}(\testobjectives_1,\testsuites)<1 $ implies  $M_{\coveringmethod_2}(\testobjectives_2,\testsuites)<1 $, for their respective objectives $\testobjectives_1$ and $\testobjectives_2$. Particularly, as discussed in~\cite{salay2018using}, when considering the use of machine models in safety critical applications 
like automotive software, DNN structural coverage criteria can be applied in a similar manner as their traditional software counterparts (e.g., statement coverage or MC/DC)
to different Automotive Safety Integrity Levels (ASALs).

\bigskip

\subsection{Test Case Generation}
\label{sec:test-gen}

In the following, we survey the test case generation methods for DNNs that have not been covered in Section~\ref{sec:test-criteria} and that do not employ the existing adversarial attack algorithms. We will review adversarial attacks in  Section~\ref{sec:attackndefence}. 

\subsubsection{Input Mutation}

Given a set of inputs, input mutation generates new inputs (as test cases) by changing the existing input according to some predefined transformation rules or algorithms. For example, \cite{wicker2018feature} systematically mutates input dimensions with the goal of enumerating all hyper-rectangles in the input space. Moreover, aiming at testing the fairness (i.e., free of unintended bias) of DNNs, 
AEQUITAS~\cite{udeshi2018automated} essentially employs an input mutation technique to first randomly sample a set of inputs and then explore the neighbourhood of the sampled inputs by changing a subset of input dimensions, however, it has not been applied to DNN model.


\subsubsection{Fuzzing}

Fuzzing, or fuzz testing, is an automated software testing technique that efficiently generates a massive amount of
random input data (possibly invalid or unexpected) to a program, which  is then monitored for exceptions and failures. 
A fuzzer can be mutation-based that modifies existing input data. Depending on the level of awareness of the program structure,
the fuzzer can be white/grey/block-box. 
There are recent works that adopt fuzz testing to deep neural networks.

TensorFuzz~\cite{odena2018tensorfuzz} is a coverage-guided fuzzing  method for DNNs.
It randomly mutates the inputs, guided by a coverage metric over the goal of satisfying user-specified constraints. The coverage is measured by a fast approximate nearest neighbour
algorithm. TensorFuzz is validated in finding   numerical errors, generating disagreements between DNNs 
and their quantized versions, and surfacing undesirable behaviour in DNNs.
Similar to TensorFuzz, DeepHunter \cite{xie2018coverage} is another coverage-guided
grey-box DNN fuzzer, which utilises these extensions of neuron coverage from \cite{ma2018deepgauge}.
Moreover, DLFuzz~\cite{dlfuzz} is a differential fuzzing testing framework. It mutates the input to maximise the 
neuron coverage and the prediction difference between the original 
input and the mutated input.

\subsubsection{Symbolic Execution and Testing}

Though input mutation and fuzzing are good at generating a large amount of random data, there is no guarantee that certain
test objectives will be satisfied.
Symbolic execution (also symbolic evaluation) is a means of analysing a program to determine what inputs cause each part of a program to execute. It assumes symbolic values for inputs rather than obtaining actual inputs as normal execution of the program would, and thus arrives at expressions in terms of those symbols for expressions and variables in the program, and constraints in terms of those symbols for the possible outcomes of each conditional branch. 

\added{Concolic testing is a hybrid software testing technique that alternates between concrete execution,
i.e., testing on particular  inputs,  and  symbolic  execution, i.e., symbolically encoding particular execution paths.} This idea still holds for deep neural networks.
In DeepConcolic~\cite{sun2018concolic,sun2018concolicb}, coverage criteria for DNNs that have been studied
in  the  literature are first formulated using the Quantified  Linear Arithmetic over Rationals,  and  then  a 
coherent  method  for  performing concolic testing to increase test coverage is provided.
The concolic procedure starts from executing the DNN using concrete inputs. Then, for those test objectives that
have not been satisfied, they are ranked according to some heuristic. Consequently, a top ranked pair of test objective
and the corresponding concrete input are selected and symbolic analysis is thus applied to find a new input test.
The experimental results show the effectiveness of the concolic testing approach
in both achieving high coverage and finding adversarial examples.



The idea in \cite{gopinath2018symbolic} is  to  translate  a  DNN  into  an  imperative  program,  thereby
enabling program analysis to assist with DNN validation. 
It introduces  novel  techniques  for  lightweight
symbolic  analysis  of  DNNs  and  applies  them  in  the  context  of
image classification to address two challenging problems, i.e.,  identification of important pixels (for attribution and
adversarial  generation), and  creation  of  1-pixel  and  2-pixel
attacks. In \cite{agarwal2018automated}, black-box style local explanations are first called to build a decision tree, to
which the symbolic execution is then applied to detect individual discrimination in a DNN: such a discrimination exists
when two inputs, differing only in the values of some specified attributes (e.g., gender/race), get different decisions
from the neural network.

\subsubsection{Testing using Generative Adversarial Networks}
Generative adversarial networks (GANs) are a class of AI algorithms 
used in unsupervised machine learning. It is implemented by a system of two neural networks contesting with each other in a zero-sum game framework.
%
DeepRoad~\cite{deeproad} 
automatically  generate  large
amounts of accurate driving scenes to test the consistency  of  DNN-based  autonomous  driving  systems  across  different  scenes.  In  particular,  it synthesises driving scenes with various weather
conditions  (including  those  with  rather  extreme
conditions) by applying the Generative Adversarial Networks (GANs) along with the corresponding
real-world weather scenes. 
%
%
%
%
%


\subsubsection{Differential Analysis}
We have already seen differential analysis techniques in \cite{PCYJ2017} and \cite{dlfuzz} that analyse the differences between multiple DNNs
to maximise the neuron coverage.
Differential analysis of a single DNN's internal states has been also applied to debug the neural network model by \cite{mode},  in which a DNN is said to be
buggy when its test accuracy for a specific output label is lower than the ideal accuracy. Given a buggy output label, the differential analysis in \cite{mode}
builds two heat maps corresponding to its correct and wrong classifications. Intuitively, a heat map is an image
whose size equals to the number of neurons and the pixel value represents the importance of a neuron (for the output). Subsequently,
the difference between these two maps can be used to highlight these faulty neurons that are responsible for the
output bug. Then, new inputs are generated (e.g., using GAN) to re-train the DNN so to reduce the influence of the detected faulty neurons
and the buggy output.


\subsection{Model-Level Mutation Testing}\label{sec:mutationTesting}

Mutation testing is a white-box testing technique that performs by changing certain statements in the source code and checking if the test cases are able to find the errors. 
%
Once a test case fails on a mutant, the mutant is said to be killed.
Mutation testing is not used to find the bugs in software but evaluate
the quality of the test suite which is measured by the percentage of mutants that they kill.


~\cite{shenmunn} 
proposes five mutation operators, including (i) deleting one neuron in input layer, (ii) deleting one or more hidden neurons, (iii) changing one or more activation functions, (iv) changing one or more bias values, and (v) changing weight value. 
\cite{deepmutation} considers data mutations, program mutations, and model-level mutations. For data mutations, a few operations on training dataset are considered, including duplicating a small portion of data, injecting faults to the labels, removing some data points, and adding noises to the data. For program mutations, a few operations are considered, including adding or deleting a layer and remvoing activation function. Model-level mutations include changing the weights, shuffling the weights between neurons in neighboring layers, etc.  
%
%
%
Moreover, \cite{cheng2018manifesting} simulates program bugs by mutating Weka implementations of several classification algorithms, including
Naive Bayes, C4.5, k-NN, and SVM.

\subsection{Summary}

We compare the testing methods for DNNs in Table \ref{tab:testings}. Overall, search algorithms, including greedy search and gradient ascent
(GA) search, are  often used in the testing. For simple coverage criteria such as neuron coverage and its
extensions and surprise coverage, established machine learning adversarial attack algorithms (e.g., FGSM~\cite{FGSM} and JSMA \cite{JSMA}) are sufficient enough
for test case generation. In the more complex cases, Linear Programming (LP) or Integer Linear Programming (ILP) approaches can be used,
and the Monte Carlo Tree Search (MCTS) method is called for generating tests for the safety coverage. Advanced testing methods like concolic testing and fuzzing
have been also developed for DNNs. 

Sometimes, distances between test inputs need to be taken into account in DNN testing to guide the tests generation.
As in the last column of Table \ref{tab:testings}, different norm distances have been applied, however, there is no conclusion
on which one is the best. Works like \cite{PCYJ2017,dlfuzz} are in principle based on differential analysis of multiple DNNs,
thus more than one DNN inputs are expected. \added{As in \cite{nollerhydiff}, DNN adversarial testing can be formulated in the form of software differential testing.}

\begin{table*}[h]
\scriptsize
    \caption{Comparison between different DNN testing methods}
    \label{tab:testings}
    \centering
    \scalebox{1.0}{
    \def\arraystretch{1.2}
    \begin{tabular}{|p{27mm}||p{20mm}|p{25mm}|p{10mm}|p{15mm}|}
    \toprule
                                    & Test generation           & Coverage criteria                 &   DNN inputs  &  Distance metric  \\\hline
     \cite{PCYJ2017}                & dual-objective search     & neuron coverage                   &   multiple    &   $L_{1}$    \\\hline
     \cite{tian2017deeptest}        &   greedy search           & neuron coverage                   & single        &   Jaccard distance    \\\hline     
     \cite{wicker2018feature}       & MCTS                      & safety coverage                   & single        &   n/a             \\\hline
     \cite{ma2018deepgauge}         & adversarial attack    & neuron coverage extensions        & single        & n/a   \\\hline 
     \cite{sun2018testing,sun2018testingjournal,sun2018testing-b} & LP, adaptive GA search & MC/DC    & single    & $L_{\infty}$      \\\hline
     \cite{huang-atva18}            & 0-1 ILP   &   quantitative projection coverage        &   single  & n/a \\\hline
     \cite{kim2018guiding}    & adversarial attacks  &   surprise coverage   &   single  &   n/a \\\hline    
     \cite{sun2018concolic,sun2018concolicb}     &   concolic testing &   MC/DC, neuron coverage and its extensions   &   single  &   $L_0,L_{\infty}$  \\\hline
     \cite{odena2018tensorfuzz} &   fuzzing     &   fast approximate nearest neighbour  & single    &   n/a    \\\hline
     \cite{xie2018coverage}     & fuzzing       &   neuron coverage extensions  &   single  &   n/a \\\hline
     \cite{dlfuzz}              &   fuzzing     &   neuron coverage             &   multiple    &   n/a \\\hline
     \cite{gopinath2018symbolic,agarwal2018automated}    &   symbolic execution  &  n/a  &   single  & n/a \\\hline  
     \cite{deeproad}            &   GAN         &   n/a                         &   single      &   n/a \\\hline
     \cite{mode}                &   GAN         &   n/a                         &   single      &   n/a \\
     
    \bottomrule
    \end{tabular}
    }
\end{table*}


Up to now, most techniques are developed by extending the existing techniques from software testing with simple adaptations. It is necessary to  validate the developed techniques (as discussed in Section~\ref{sec:validationTesting}) and study the necessity of developing dedicated testing techniques. \added{Meanwhile, recent effort has been also made to apply surveyed DNN testing methods from feedforward networks to other deep learning models like recurrent neural networks~\cite{huang2019testrnn,huang2019test,du2019deepstellar}.}
\cleardoublepage
\section{Adversarial Attack and Defence}
\label{sec:attackndefence}
 
The goal of attack techniques is to provide evidence (i.e., adversarial examples) for the lack of robustness of a DNN without having a provable guarantee. Defence techniques are dual to attack techniques, by either improving the robustness of the DNN to immune the adversarial examples, or differentiating the adversarial examples from the correct inputs.
From Section~\ref{sec:advAttacks} to Section~\ref{sec:inputAgnosticAdvAttacks}, we review the attack techniques from different aspects. These techniques are compared in Section~\ref{sec:comparisonAdvAttacks} with a few other techniques from verification. Then, in Section~\ref{sec:defence} and Section~\ref{sec:certifiedDefence}, we review defence techniques (without a guarantee), and certified defence techniques, respectively.


\subsection{Adversarial Attacks}\label{sec:advAttacks}

Given an input, an adversarial attack (or attacker) tries to craft a perturbation or distortion to the input to make it misclassified by a well-trained DNN. Usually, it is required that the adversarial example is misclassified with high confidence. Attack techniques can be divided roughly into two groups based on the type of misclassification they try to achieve:

\begin{itemize}
    \item With a targeted perturbation, the attacker is able to control the resulting misclassification label.
    \item With an un-targeted perturbation, the attacker can enable the misclassification but cannot control its resulting misclassification label.
\end{itemize}

According to the amount of information an attacker can access, adversarial perturbations can also be classified into two categories:
\begin{itemize}
    \item White-box perturbation - the attacker needs to access the parameters and the internal structure of the trained DNN, and may also need to access the training dataset.
    \item Black-box perturbation - an attacker can only query the trained DNN with perturbed inputs, without the ability to access the internal structure and parameters of the DNN.  
\end{itemize}

Moreover, according to the norm-distances used to evaluate the difference between a perturbed input and the original input, adversarial attacks can be generally classified as $L_0$, $L_1$, $L_2$ or $L_\infty$-attacks. Please note that all perturbations can be measured with any of these norms, but an attack technique can produce adversarial examples which are better measured with a particular norm.  


Currently, most of the adversarial attacks are concentrating on computer vision models. From a technical point of view, those attacks can be categorised as using cost gradients, such as in~\cite{FGSM,moosavi2017universal,biggio2013evasion}, or using gradients of the output w.r.t neural network's input, such as JSMA attack~\cite{JSMA}, or directly formulating into optimisation problems to produce adversarial perturbations such as DeepFool~\cite{moosavi2016deepfool}and C\&W attack~\cite{CW2016}.

It also has been demonstrated that adversarial examples are transferable across different neural network models~\cite{szegedy2014intriguing,JSMA}. In~\cite{KGB2016}, the authors further demonstrate that adversarial examples can be transferred into real world scenarios. Namely, adversarial examples can be still misclassified after being printed in a paper physically. In the following, we will review a few notable works in details.

\subsubsection{Limited-Memory BFGS Algorithm (L-BFGS)}

\cite{szegedy2014intriguing} noticed the existence of adversarial examples, and described them as `blind spots' in DNNs. They found that adversarial images usually appear in the neighbourhood of correctly-classified examples, which can fool the DNNs although they are human-visually similar to the natural ones. It also empirically observes that random sampling in the neighbouring area is not efficient to generate such examples due to the sparsity of adversarial images in the high-dimensional space.
Thus, they proposed an optimization solution to efficiently search the adversarial examples. Formally, assume we have a classifier $f:\real^{s_1} \rightarrow \{1 \dots s_K\}$ that maps inputs to one of $s_K$ labels, and $x \in \real^{s_1}$ is an input, $t \in \{1 \dots s_K\}$ is a target label such that $t \neq \arg\max_l f_l(x)$. Then the adversarial perturbation $r$ can be solved by

\begin{equation}
  \begin{array}{l}
    ~~~~~~\min ||r||_2 \\
  \textit{s.t.}~~\arg\max_{l} f_l(x+r) = t \\
  ~~~~~~x+r \in \real^{s_1}
  \end{array}
\end{equation}
%
Since the exact computation is hard, an approximate algorithm based on the limited-memory Broyden–Fletcher–Goldfarb–Shanno algorithm (L-BFGS) is used instead. 
Furthermore, \cite{szegedy2014intriguing} observed that adversarial perturbations are able to transfer among different model structures and training sets, i.e., an adversarial image that aims to fool one DNN classifier also potentially deceives another neural network with different architectures or training datasets.


\subsubsection{Fast Gradient Sign Method (FGSM)}

Fast Gradient Sign Method~\cite{FGSM} is able to find adversarial perturbations 
with a fixed $L_{\infty}$-norm constraint. FGSM conducts 
a one-step modification to all pixel values so that the value of the loss function  is increased  under a certain $L_{\infty}$-norm constraint. 
The authors claim that the linearity of the neural network classifier leads to the adversarial images because the adversarial examples are found by moving linearly along the reverse direction of the gradient of the cost function. 
%
%
Based on this linear explanation, \cite{FGSM} proposes an efficient linear approach to generate adversarial images. Let $\theta$ represents the model parameters, $x,y$ denote the input and the label and $J(\theta, x, y)$ is the loss function. We can calculate adversarial perturbation $r$ by
\begin{equation}
    r = \epsilon\ \text{sign}\left(\nabla_x J(\theta, x, y)\right)
\end{equation}
A larger $\epsilon$ leads to a higher success rate of attacking,  but potentially results in a bigger human visual difference. This attacking method has since been extended to a targeted and iterative version~\cite{KGB2016}.

\subsubsection{Jacobian Saliency Map based Attack (JSMA)}

\cite{JSMA} present a $L_0$-norm based adversarial attacking method by exploring the \emph{forward derivative} of a neural network. Specifically it utilizes the Jacobian matrix of a DNN's logit output w.r.t. its input to identify those most sensitive pixels which then are perturbed to fool the neural network model effectively.
%
%
Let $c \in \labels$ denote a target class and $x \in [0,1]^{s_1}$ represent an input image. JSMA will assign each pixel in $x$ a salient weight based on the Jacobian matrix. Each salient value basically quantifies the sensitivity of the pixel to the predicted probability of class $c$.
%
%
To generate the adversarial perturbation, the pixel with the highest salient weight is firstly perturbed by a \emph{maximum distortion parameter} $\tau > 0$. If the perturbation leads to a mis-classification, then JSMA attack terminates. Otherwise, the algorithm will continue until a mis-classification is achieved. When a maximum $L_0$-norm distortion $d > 0$ is reached, the algorithm also terminates. 
This algorithm is primarily to produce adversarial images that are optimized under the $L_0$-norm distance.
JMSA is generally slower than FGSM due to the computation of the Jacobian matrix. 

\subsubsection{DeepFool: A Simple and Accurate Method to Fool Deep Neural Networks}

In DeepFool, \cite{moosavi2016deepfool} introduces an iterative approach to generate adversarial images on any $L_p, p \in [1, \infty)$ norm distance. In this work, the authors first show how to search adversarial images for an affine binary classifier, i.e., $g(x) = \text{sign}(w^T\cdot x + b )$. Given an input image $x_0$, DeepFool is able to produce an optimal adversarial image by projecting $x_0$ orthogonally on the hyper-plane $\mathcal{F} = \{ x | w^T \cdot x +b =0\}$. Then this approach is generalized for a multi-class classifier: $W \in \real^{m \times k}$ and $b \in \real^k$. Let $W_i$ and $b_i$ be the $i$-th component of $W$ and $b$, respectively. We have 
\begin{equation*}
g(x) = \underset{i \in \{1 \dots k\}}{\text{argmax}}~g_i(x) \text{ where } g_i(x) = W_i^Tx + b_i
\end{equation*}
For this case, the inupt $x_0$ is projected to the nearest face of the hyper-polyhedron $P$ to produce the optimal adversarial image, namely,
\begin{equation*}
    P(x_0) = \bigcap_{i=1}^k \{x | g_{k_0}(x) \geq g_{i}(x)  \}
\end{equation*}
where $k_0 = g(x_0)$. We can see that $P$ is the set of the inputs with the same label as $x_0$. In order to generalize DeepFool to neural networks, the authors introduce an iterative approach, namely, the adversarial perturbation is updated at each iteration by approximately linearizing the neural network and then perform the projection. Please note that, DeepFool is a heuristic algorithm for a neural network classifier that provides no guarantee to find the adversarial image with the minimum distortion, but in practise it is an very effective attacking method.

\subsubsection{Carlini \& Wagner Attack}

{\em  C\&W Attack} ~\cite{CW2016} is an optimisation based adversarial attack method which formulates finding an adversarial example as image distance minimisation problem such as $L_0, L_2$ and $L_\infty$-norm. Formally, it formalizes the adversarial attack as an optimisation problem:
\begin{equation}
\lossfunction(\upsilon) = ||\upsilon||_p + c \cdot F(x + \upsilon),
\end{equation} 
where $x+\upsilon$ is a valid input, and $F$ represents a surrogate function such as $x+\upsilon$ is able to fool the neural network when it is negative. The authors directly adopt the optimizer Adam~\cite{kingma2014adam} to slove this optimization problem. 
It is worthy to mention that C\&W attack can work on three distance including $L_2$, $L_0$ and $L_{\infty}$ norms. 
%
A smart trick in C\&W Attack lies on that it introduces a new optimisation variable to avoid box constraint (image pixel need to within $[0,1]$).  C\&W attack is shown to be a very strong attack which is more effective than JSMA~\cite{JSMA}, FGSM~\cite{FGSM} and DeepFool~\cite{moosavi2016deepfool}. It is able to find an adversarial example that has a significant smaller distortion distance, especially on $L_2$-norm metric.


\begin{figure}
    \centering
    \includegraphics[width=0.6\textwidth]{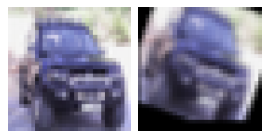}
    \caption{Rotation-Translation: Original (L) `automobile', adversarial (R) `dog' from \cite{engstrom2017rotation}. {\em The original image of an `automobile' from the CIFAR-10 dataset is rotated (by at most $30^\circ$) and translated (by at most 3 pixels) results in an image that state-of-art classifier ResNet~\cite{he2016deep} classifies as `dog'.}}
    \label{fig:rottran}
\end{figure}

\subsection{Adversarial Attacks by Natural Transformations}\label{sec:naturalAdvAttacks}

Additional to the above approaches which perform adversarial attack on pixel-level, research has been done on crafting adversarial examples by applying natural transformations.

\subsubsection{Rotation and Translation}

\cite{engstrom2017rotation} argues that most existing adversarial attacking techniques generate adversarial images which appear to be human-crafted and less likely to be `natural'.
It shows that DNNs are also vulnerable to some image transformations which are likely occurring in a natural setting. For example, translating or/and rotating an input image could significantly degrade the performance of a neural network classifier. Figure~\ref{fig:rottran} gives a few examples. 
Technically, given an allowed range of translation and rotation such as {\em $\pm3$ pixels $\times \pm 30^\circ$}, \cite{engstrom2017rotation} aims to find the minimum rotation and translation to cause a misclassification. To achieve such an purpose, in this paper, several ideas are explored including
\begin{itemize}
    \item a first-order iterative method using the gradient of the DNN's loss function,
    \item performing an exhaustive search by descretizing the parameter space,
    \item a worst-of-k method by randomly sampling $k$ possible parameter values and choosing the value that cause the DNN to perform the worst. 
\end{itemize}

\subsubsection{Spatially Transformed Adversarial Examples}

\begin{figure}[t]
    \centering
    \begin{minipage}{0.24\textwidth}
        \centering
        \includegraphics[width=\textwidth]{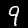}
        \text{(a) Original:  `9'}
    \end{minipage}
    \begin{minipage}{0.24\textwidth}
            \centering
        \includegraphics[width=\textwidth]{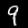}
        \text{(b) Adversarial: `8'}
    \end{minipage}
    \begin{minipage}{0.24\textwidth}
        \centering
        \includegraphics[width=\textwidth]{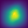}
        \text{(c) x-dimension}
    \end{minipage}
    \begin{minipage}{0.24\textwidth}
            \centering
        \includegraphics[width=\textwidth]{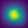}
        \text{(d) y-dimension}
    \end{minipage}
    \caption{Applying spatial transformation to MNIST image of a `9'~\cite{xiao2018spatially}. {\em the Image (a) on the left is the original MNIST example image of a `9', and image (b) is the spatially transformed adversarial version that a simple convolutional network \cite{papernot2018cleverhans} labels as `8'. Notice how minor the difference between the two images is - the `9' digit has been very slightly `bent' - but is sufficient for miss-classification. The flow-field that defines the spatial transformation is visualised in Image (c) (x-dimension) and Image (d) (y-dimension). The brighter areas indicate where the transformation is most intense - leftwards in the x-dimension and upwards in the y-dimension.}}
    \label{fig:stadvMNIST}
\end{figure}

\cite{xiao2018spatially} 
introduces 
to produce uncontrived adversarial images via mortifying the pixel's location using spatial transformations instead of directly changing its pixel value. The authors use the flow field to control the spatial transformations, which essentially quantifies the location displacement of a pixel to its new position. Figure~\ref{fig:stadvMNIST} gives a few examples. Using a bi-linear interpolation approach the generated adversarial example is differentiable w.r.t. the flow field, which then can be formulated as an optimisation problem to calculate the adversarial flow field.
Technically, 
\cite{xiao2018spatially} introduce a distance measure $L_{flow}(\cdot)$ (rather than the usual $L_p$ distance) to capture the local geometric distortion. Similar to  C\&W attack~\cite{CW2016}, the flow field is obtained by solving an optimisation problem in which the loss function is defined to balance between the $L_{flow}$ loss and adversarial loss.
Through human study, \cite{xiao2018spatially} demonstrate that adversarial examples based on such spatial transformation are more similar to original images based on human perception, comparing to those adversarial examples from  $L_p$-norm based attacks such as FGSM~\cite{FGSM} and C\&W~\cite{CW2016}.

\subsubsection{Towards Practical Verification of Machine Learning: The Case of Computer Vision Systems (VeriVis)}

\cite{pei2017towards} introduce a verificaiton framework, called VeriVis, to measure the robustness of DNNs on 
%
%
a set of 12 practical image transformations including reflection, translation, scale, shear, rotation, occlusion brightness, contrast, dilation, erosion, average smoothing, and median smoothing. Every transformation is controlled by a key parameter with a \emph{polynomial-sized} domain. Those transformations are exhaustively operated on a set of input images. Then the robustness of a neural network model can be measured.
%
\textsc{VeriVis} is applied to evaluate several state-of-the-art classification models, which empirically reveals that all classifiers show a significant number of safety violations. 

\subsection{Input-Agnostic Adversarial Attacks}\label{sec:inputAgnosticAdvAttacks}

A key characteristic of the above attacks lies on that an adversarial example is generated with respect to a specific input, and therefore cannot be applied to the other input. Thus some researchers show more interests in \emph{input-agnostic} adversarial perturbations.


\subsubsection{Universal Adversarial Perturbations}

The first method on input-agnostic adversarial perturbation was proposed by~\cite{moosavi2017universal}, called \emph{universal} adversarial perturbations (UAP), since UAP is able to fool a neural network on \emph{any} input images with high probability. 
Let $\upsilon$ the the current perturbation. UAP iteratively goes through a set $\trainingDataset$ of inputs sampled 
from the input distribution $\inputDistribution$. At the iteration for $x_i\in \trainingDataset$ it updates the perturbation $\upsilon$ as follows. First, it finds the minimal 
$\Delta \upsilon_i$ w.r.t. $L_2$-norm distance so that $x_i+\upsilon+\Delta\upsilon_i$ is incorrectly classified by neural network $\networks$. Then, it projects $\upsilon + \Delta\upsilon_i$ back into $L_p$-norm ball with a radius $d$ to enable that the generated perturbation is sufficiently small, i.e., let 
\begin{equation}
\begin{array}{l}
\upsilon = \arg\underset{\upsilon'}{\min} {||\upsilon' - (\upsilon + \Delta\upsilon_i) ||_2}\\
\text{s.t.~~} ||\upsilon'||_p \leq d.
\end{array}
\end{equation}
The algorithm will proceed until the empirical error of the sample set is sufficiently large, namely, no less than $1 - \delta$ for a pre-specified threshold $\delta$.

\subsubsection{Generative Adversarial Perturbations}

By training a universal adversarial network (UAN), \cite{hayes2018learning} generalizes the C\&W attack in~\cite{CW2016} to generates input-agnostic adversarial perturbations. Assume that we have
a maximum perturbation distance $d$ and an $L_p$ norm, a UAN $\mathcal{U}_{\theta}$ randomly samples an input $z$ from a normal distribution and generates a raw perturbation $r$. Then it is scaled through $w \in [0, \frac{d}{||r||_p}]$ to have $r'=w\cdot r$. Then, the new input $x+r'$ needs to be checked with DNN $\networks$ to see if it is an adverarial example. 
The parameters $\theta$ is optimized by adopting a gradient descent method, similar to the one used by C\&W attack~\cite{CW2016}. 


Later on, \cite{poursaeed2017generative} introduces a similar approach in\cite{hayes2018learning} to produce input-agnostic adversarial images. It first samples a random noise to input to UAN, and the output then is resized to meet an $L_p$ constraint which is further added to input data, clipped, and then is used to train a classifier. This method is different to~\cite{hayes2018learning} in two aspects. Firstly, it explores two UAN architectures including U-Net \cite{ronneberger2015u} and ResNet Generator \cite{johnson2016perceptual}, and find ResNet Generator works better in the majority of the cases. 
%
Secondly, this work also trained a UAN by adopting several different classifiers, thus the proposed UAP can explicitly fool multiple classifiers, which is obtained by the below loss function:
\begin{equation}
    l_{multi-fool}(\lambda) = \lambda_1 \cdot l_{fool_1} + \dots + \lambda_m \cdot l_{fool_m}
\end{equation}
where $l_{fool_i}$ denotes the adversarial loss for classifier $i$, $m$ is the number of the classifiers, and $\lambda_i$ is the weight to indicate the difficulty of fooling classifier $i$.

\subsection{Other Types of Universal Adversarial Perturbations}

\added{Instead of fooling neural networks on image classification tasks, recently there are other types of universal attacking methods emerged, which target to attack DNN models in a wide range of applications including semantic image segmentation~\cite{hendrik2017universal}, image retrieval~\cite{li2019universal}, speech recognition~\cite{neekhara2019universal}, and object recognition~\cite{mopuri2018generalizable}.

In the work of~\cite{hendrik2017universal}, the authors proposed two approaches to generate universal perturbations that can adversarially fool a set of images for semantic image segmentation. The first approach in their paper is able to produce a fixed target segmentation as the output, and the second method can remove a designated target class while reserving the original segmentation outcome. This work also empirically demonstrated that UAP on semantic level has a smaller perturbation space than image classification tasks. Later on, \cite{li2019universal} presented a UAP method that can attack image retrieval systems, i.e., enabling an image retrieval system to put irrelevant images at the top places of a returned ranking list for a query. Since images in a retrieval system are labeled by similarity instead of labels in classification, the authors proposed to corrupt the similarity relationship of neighborhood structures by swapping the similarity in the tuple structures. Except for those UAPs on image domain, lately \cite{neekhara2019universal} proposed a universal attacking method to fool speech recognition systems. In this work, the authors revealed the existence of audio-agnostic perturbations on deep speech recognition systems. They discovered that the generated universal perturbation is transferable across different DNN model architectures. Those universal adversarial perturbations further demonstrate that the venerability of DNNs to universal adversarial examples not only appears in image classification domain but also exists in a wide range of DNN-based systems.}

\subsection{Summary of Adversarial Attack Techniques}\label{sec:comparisonAdvAttacks}

Table~\ref{tab:attacks} summarises the main similarities and differences of a number of surveyed adversarial attacks from five aspects: distance metric, whether the attack is targeted or un-targeted, the level of accessed information required (i.e., model structures/parameters, logits, output confidences, label), dataset tested, and core method used. 
\begin{table}
\scriptsize
    \caption{Comparison between different adversarial attacks}
    \label{tab:attacks} 
    \centering
    \scalebox{1.0}{
    \def\arraystretch{1.2}
    \begin{tabular}{|p{25mm}||p{20mm}|p{13mm}|p{15mm}|p{18mm}|p{20mm}|}
    \toprule
      & Distance Metric & Targeted/ Untargeted  & Accessed Information  &  Dataset Tested & Core Method \\\hline
     {L-BFGS \cite{szegedy2014intriguing} }& $L_{2}$     & Untarget &   Model Parameters    &  MNIST & L-BFGS \\  \hline
     {FGSM \cite{FGSM} }& $L_{\infty}$     & Untarget &   Model Parameters    &  MNIST, CIFAR10 & Fast linear algorithm \\  \hline
     {DeepFool \cite{moosavi2016deepfool} }& $L_p, p \in [1, \infty)$     & Both &   Model Parameters  &  MNIST, CIFAR10 & Iterative linearization\\ \hline
     {C \& W \cite{CW2016} }& $L_0, L_2, L_\infty$ & Both &   Logits  &  MNIST, CIFAR10 & Adam Optimizer\\  \hline
     {JSMA \cite{JSMA} }& $L_0$ & Both &   Model Parameters  &  MNIST, CIFAR10 & Jacobian Saliency \\ 
           
     \hline
     {DeepGame \cite{wu2019game}}& $L_0, L_1, L_2, L_\infty$ & Untarget &   Logits  &  MNIST, CIFAR10 &  Game-based approach\\
      \hline
     {L0-TRE \cite{ruan2019global}}& $L_0$ & Untarget & Logits  &  MNIST, CIFAR10, ImageNet &  Tensor-based grid search\\
           \hline
     {DLV \cite{HKWW2017}}& $L_1, L_2$ & Untarget &   Model Parameters  &  MNIST, CIFAR10, GTSRB &  Layer-by-layer search\\
      \hline
     {SafeCV \cite{wicker2018feature}}& $L_0$ & Both & Logits  &  MNIST, CIFAR10 &  Stochastic search\\
     
     \hline
     {\cite{engstrom2017rotation}}& N/A (Natural Transformations) & Both &   Logits  &  MNIST, CIFAR10, ImageNet & Rotating and/or translating input images\\ \hline
     {\cite{xiao2018spatially}}& $L_{flow}(\cdot)$ (Measuring geometric distortion)& Both &   Logits  &  MNIST, CIFAR10, ImageNet & Minimising adversarial and $L_{flow}$ loss\\
     \hline
     {VeriVis \cite{pei2017towards}}& N/A (Natural Transformations) & Both &   Logits  &  MNIST, CIFAR10, ImageNet & A set of 12 `real-world' transformations\\
     
     \hline
     {UAP \cite{moosavi2017universal}}& $L_2$ (Universal perturbation) & Both &   Logits  &  MNIST, CIFAR10, ImageNet & Generalizing DeepFool into universal adversarial attacks\\
     \hline
     {UAN \cite{hayes2018learning}}& $L_p$ (Universal perturbation) & Both &   Logits  &  MNIST, CIFAR10, ImageNet & Generalizing C\&W into universal adversarial attacks\\
     \hline
     {\cite{poursaeed2017generative}}& $L_p$ (Universal perturbation) & Both &   Logits  &  MNIST, CIFAR10, ImageNet &  Training a generative network\\
     
       \hline
   \added{{\cite{hendrik2017universal}}}& Semantic Segmentation (Universal perturbation) & Target &   Model Parameters  &  Cityscapes,
   CamVid
   &  Generalizing UAP to segmentation tasks\\
            \hline
   \added{{\cite{li2019universal}}}& Image Retrieval (Universal perturbation) & Disturbing ranking list &   Model Parameters  &  Oxford5k,
   Paris6k, ROxford5k, RParis6k &  Corrupting neighbourhood relationships in feature space \\
     
            \hline
     \added{{\cite{neekhara2019universal}}} & Speech Recognition (Universal perturbation) & Untarget &   Model Parameter  &  Mozilla Common Voice Dataset
     &  Iterative gradient sign method to maximize the CTC-Loss\\
     
    \bottomrule
    \end{tabular}
    }
\end{table}

\subsection{Adversarial Defence}\label{sec:defence}

On the opposite side of adversarial attacks~\cite{biggio2013evasion, szegedy2014intriguing}, researchers also show huge interests in designing various defence techniques, which are to either identify or reduce adversarial examples so that the decision of the DNN can be more robust. Until now, the developments of attack and defence techniques have been seen as an ``arm-race''. For example, most defences against attacks in the white-box setting, including~\cite{papernot2016distillation,metzen2017detecting,hendrycks2016early,meng2017magnet}, have been demonstrated to be vulnerable to e.g., iterative optimisation-based attacks~\cite{carlini2017magnet,carlini2017adversarial}.

\subsubsection{Adversarial Training}


Adversarial training is one of the most notable defence methods, which was first proposed by~\cite{FGSM}. It can improve the robustness of DNNs against adversarial attacks by retraining the model on adversarial examples. Its basic idea can be expressed as below:
\begin{equation}
\theta^* = \arg \min_\theta\; \mathbb{E}_{(x, y) \in \trainingDataset} \; \lossfunction(x; y; f_\theta).
\end{equation}
where $f_\theta$ is a DNN parameterized by $\theta$, $\lossfunction(\cdot)$ is the loss function, $\trainingDataset$ is the training dataset.
This is improved in \cite{madry2017towards} by assuming that all neighbours within the $\epsilon$-ball should have the same class label, i.e., local robustness. Technically, this is done by changing the optimisation problem by requiring that   
%
for a given $\epsilon$-ball (represented as a $d$-Neighbourhood), to solve \begin{equation}
\theta^* = \arg \min_\theta\; \mathbb{E}_{(x, y) \in \trainingDataset} \left[ \max_{\delta \in [-\epsilon,\epsilon]^{s_1}} \ell(x+\delta; y; f_\theta) \right].
\end{equation} 
\cite{madry2017towards} adopted Projected Gradient Descent (PGD) to approximately solve the inner maximization problem.

Later on, to defeat the iterative attacks, \cite{na2017cascade} proposed to use a cascade adversarial method which can produce adversarial images in every mini-batch. Namely, at each batch, it performs a separate adversarial training by putting the adversarial images (produced in that batch) into the training dataset.
Moreover,~\cite{TKPGBM2017} introduces ensemble adversarial training, which augments training data with perturbations transferred from other models. 


\subsubsection{Defensive Distillation}

Distillation~\cite{HVD2015} is a training procedure which 
trains a DNN using knowledge transferred from a different
DNN. Based on this idea, 
\cite{PMWJS2016} proposes defensive distillation which keeps the same network architecture to train both the original
network as well as the distilled network. It proceeds by (i)  sampling a set $\{(x,f(x))\}$ of samples from the original network and training a new DNN $f^1$, and (ii) sampling a set $\{(x,f^1(x))\}$ of samples from the new DNN $f^1$ and training another new DNN $f^d$. It is shown that the distilled DNN $f^d$ is more robust than the original DNN.

\subsubsection{Dimensionality Reduction}

Some researchers propose to defense adversarial attacks by dimension reduction. For example, \cite{BCM2017} use Principal Component Analysis and data `anti-whitening' reduces the high dimensional inputs for improving the resilience of various machine learning models including deep neural networks. In~\cite{XEQ2017}, the authors introduce a feature squeezing approach to detect the adversarial images by “squeezing” out unnecessary input
features.

\subsubsection{Input Transformations}

A popular defence approach is to do input transformations before feeding an input into the DNN. 
\cite{MC2017} suggest that an adversarial example can be either far away from existing data or close to the decision boundary. For the former, one or more
separate detector networks are used to learn to differentiate between normal and adversarial examples
by approximating the manifold of normal examples. For the latter, a reformer network implemented by an auto-encoder moves adversarial examples
towards the manifold of normal examples so that they can be classified correctly.

On the other hand, \cite{song2017pixeldefend} observe that most of the adversarial images lie in the low probabilistic area of the distribution. So they train a PixelCNN generative model~\cite{vdOKVEGK2016} to project the potential perturbed images into the data manifold and then greedily search an image with a highest probability, which was finally fed into DNNs.
%
%
Along this line, \cite{samangouei2018defense} adopt a similar idea but apply Generative Adversarial Networks (GAN)~\cite{goodfellow2014generative} instead of the PixelCNN.

\cite{guo2017countering} exercise over five input transformations, including (i)  random image cropping and re-scaling, to altering the spatial positioning of the
adversarial perturbation; (ii) conducting bit-depth reduction 
to 
removes small (adversarial) variations in pixel
values from an image; (iii) JPEG compression
to remove small perturbations; (iv) total variance minimisation by randomly drop pixels; and (v) image quilting, which reconstructs images by replacing small (e.g., $5\times 5$)
      patches with
        patches from clean images.
        
\cite{xie2017mitigating} present a defence idea through adding several randomization layers before the neural networks.
For example, for a DNN with a $299 \times 299$ input, it first uses a randomization layer to randomly rescale the image into an image with a $[299, 331) \times [299, 331)$ size, and then takes a second randomisation layer to randomly zero-pad the image. By doing so, a number of randomised patterns can be created before an image is fed to the DNN.
The paper claims that randomization at inference time makes the network much more robust to adversarial images, especially for iterative attacks (both white-box and black box), but hardly hurts the performance on clean (non-adversarial) images.

\subsubsection{Combining Input Discretisation with Adversarial Training}

Input discretization is to separate continuous-valued
pixel inputs into a set of non-overlapping buckets, which are each mapped to a fixed binary
vector. Similar as input transformations, input discretization based approaches  apply a non-differentiable and non-linear transformation (discretization) to
the input, before passing it into the model. 
%
%
%
%
In~\cite{BRRG2018}, the authors present an idea of using thermometer encoding in adversarial training. Because the gradient-based methods such as PGD is impossible due to the discretization brought by the thermometer encoding in adversarial training, as an alternative, the authors introduce Logit-Space Projected Gradient Ascent (LS-PGA). This paper demonstrates that, the combination of thermometer encoding with adversarial training, can improve the robustness of neural networks against adversarial attacks.


\subsubsection{Activation Transformations}

\cite{dhillon2018stochastic} introduces stochastic activation pruning, by adapting the activation of hidden layers on their way to propagating to the output. The idea is that, in each layer during forward propagation, it stochastically drops out nodes, retains nodes with probabilities proportional to the magnitude of their activation, and scales up
the surviving nodes to preserve the dynamic range of the activations. 
%
Applying SAP to increase the robustness at the price of slightly decreasing clean classification accuracy.

\subsubsection{Characterisation of Adversarial Region}

Adversarial region is a connected region of the input domain in which all points subvert the DNN in a similar way. Summarised in \cite{ma2018characterizing}, they are of low probability (i.e., not naturally occurring),  span a contiguous multidimensional
space,   lie off (but are close to) the data submanifold, and  have class distributions
that differ from that of their closest data submanifold. 

\cite{FCSG2017} uses Kernel Density estimation (KDE) as a measure to identify adversarial subspaces. Given an input $x$ and its label $t$, it is to compute 
\begin{equation}
KDE(x) = \frac{1}{|\trainingDataset_t|}\sum_{\hat x\in \trainingDataset_t}\text{exp}(\frac{|f^{K-1}(x)-f^{K-1}(\hat x)|^2}{\sigma^2})
\end{equation}
where $f^{K-1}$ is the function for the first $K-1$ layers, i.e., $f^{K-1}(x)$ is the logit output of the input $x$ from $f$, and $\sigma$ is a Gaussian standard deviation. Then based on the threshold $\tau$, it can distinguish if  $x$ is an adversarial or natural image.



\cite{ma2018characterizing} uses Local Intrinsic Dimensionality (LID) \cite{Houle2017} to measure the adversarial region by considering the local distance distribution
from a reference point to its neighbours.




\subsubsection{Defence against Data Poisoning Attack}

Instead of  defencing against adversarial attacks, \cite{SKL2017} considers a data poisoning attack in which the attacker is to manipulate a percentage $\epsilon$ of the training dataset $\trainingDataset$ to have a new dataset $\trainingDataset_p$ such that $|\trainingDataset_p|=\epsilon |\trainingDataset|$. The purpose of the attacker is 
to mislead the defender who trains the model  over the set $\trainingDataset\cup \trainingDataset_p$. The success of defence or attack is defined with respect to a loss function $\lossfunction$.  

\subsection{Certified Adversarial Defence}\label{sec:certifiedDefence}

The approaches in Section~\ref{sec:defence} cannot provide guarantee over the defence results. In this subsection, we review a few principled approaches on achieving robustness. Basically, they adapt the training objective (either the loss function or the regularisation terms) to enforce some robustness constraints. 

\subsubsection{Robustness through Regularisation in Training}

Attempts have been made on achieving  robustness by applying dedicated regularisation terms in the training objective. Since training with an objective does not necessarily guarantee the achievement of the objective for all inputs, the robustness is approximate.  For example, 
\cite{HA2017} trains with a cross-entropy loss, which makes the difference
$f_c(x)-f_k(x)$ as large as possible for $c$ the class of $x$ and all $k = 1, \ldots, s_K$, and a Cross-Lipschitz regularisation term
\begin{equation}
    \regulariser(f) = \frac{1}{nK^2} \sum_{i=1}^n \sum_{l,k=1}^{s_K} ||\nabla f_c(x_i) - \nabla f_k(x_i)||_2^2 
\end{equation}
where $\{x_i\}, {i=1..n}$ is the training dataset. The goal of this regularisation term is to make the differences of the classifier functions at the data points as constant as possible.
Moreover,
in \cite{raghunathan2018certified}, DNNs with one hidden layer are studied. An upper bound on the worst-case loss is computed, based on a semi-definite relaxation. Since this upper bound is differentiable, it is taken as a regularisation term.

\subsubsection{Robustness through Training Objective}\label{sec:certifiedthroughtrainingobj}

\cite{sinha2018certifiable}  considers the problem
\begin{equation}
\min_{\theta}\sup_{\inputDistribution\in {\cal P}}\mathbb{E}_\inputDistribution[\lossfunction(x,y;\theta)]
\end{equation}
where ${\cal P}$ is a set of distributions around the data-generating distribution $\inputDistribution_0$, and $x\sim \inputDistribution_0$.
The set ${\cal P}$ includes the distributions that are close to $\inputDistribution_0$ in terms of the Wasserstein metric. 
Then, considering  
a Lagrangian penalty formulation of the Wasserstein metric, a training procedure is applied so that the model parameters can be updated with worst-case perturbations of training data. 

Moreover, \cite{Gowal_2019_ICCV} considers the following loss:  
\begin{equation}
\kappa \lossfunction(z_K,y_{true};\theta) + (1-\kappa) \lossfunction(\hat{z}_K(\epsilon),y_{true};\theta)
\end{equation}
where $\lossfunction$ is the cross-entropy loss function, $\lossfunction(z_K,y_{true};\theta)$ is the loss of fitting data such that $z_K$ is the output logit and $y_{true}$ is the correct output label, and $\lossfunction(\hat{z}_K(\epsilon),y_{true};\theta)$ is the loss of satisfying the specification such that $\hat{z}_k(\epsilon)$ is the worst-case logit where the logit of the true class is equal to its lower bound and the other logits are equal to their upper bound. The lower and upper bounds are approximated with interval bound propagation \cite{katz2017reluplex}. The hyper-parameter $\kappa$ is to control the relative weight of satisfying the specification versus fitting the data.

\subsection{Summary of Adversarial Defence Techniques}

In summary, defence techniques 
are either to identify 
or to reduce  adversarial examples. Since these do not lead to any guarantees on the robustness, the so-called ``arm race'' with attack techniques appears. Certified defence aims to improve on this situation by adding robustness constraints into the training procedure (either through regularisation term or training objective).  

Moreover, defence techniques are   closely related to the verification techniques. Every verification method can serve as a defence method, for its ability to identify the adversarial examples with guarantees. 
For an adversarial input, it may be classified wrongly if directly passing the DNN. To defence this, a DNN verifier can step in and determine if it is an adversarial example. If it is, the verification output can be utilised to alert that the classification from the DNN is not trustable.

\cleardoublepage

\section{Interpretability}
\label{sec:interpretability}

Interpretability (or explainability) has been an active area of research, due to the black-box nature of DNNs. DNNs have shown the ability of achieving high precision in their predictions. However, to gain the trust from human users, it is essential to enable them to understand the decisions a DNN has made. Moreover, it is also possible that the results in interpretability can enhance other techniques, such as verification, testing, and adversarial attacks. 

In the following, we review three categories of interpretability methods. First of all, instance-wise explanation, which aims to explain the decision made by a DNN on a given input, is presented in Section~\ref{sec:visualisation}, Section~\ref{sec:ranking}, and Section~\ref{sec:saliencymaps}. Second, model explanation, which aims to provide a better understanding on the complex model, is reviewed in Section~\ref{sec:influencefunction} and Section~\ref{sec:simplermodels}. Finally, in Section~\ref{sec:informationtheory}, we review the recent progress of using information theoretical methods to explain the training procedure.  

\subsection{Instance-wise Explanation by Visualising a Synthesised Input}\label{sec:visualisation}

The first line of research aims to understand the representation learned by DNNs through visualisation over another input generated based on the current input. It  focuses on convolutional neural networks (CNNs) with images as their inputs, and is mainly for instance-wise explanation. Technically, with respect to Definition~\ref{def:instanceExplanation}, it is to let $t=s_1$ and $\explain(f,x)\in \real^{s_1}$. 

\subsubsection{Optimising Over a Hidden Neuron}

Recall that $ u_{k,l}(x)$ is the activation of the $l$-th neuron on the $k$-th layer for the input $x$.
\cite{EBCV2009} synthesises images that cause high activations for particular neurons by treating the synthesis problem as an optimisation problem as follows. 
\begin{equation}
x^* =\arg \max_{x} u_{k,l}(x).
\end{equation}
 In general, the optimisation problem  is non-convex, and therefore a gradient ascent based algorithm is applied to find a local optimum. 
Starting with some initial input $x=x_0$, the activation $u_{k,l}(x)$  is computed, and then steps are taken in the input space along the gradient direction  ${\partial u_{k,l}(x)/{\partial x}}$ to synthesize inputs that cause higher and higher activations for the neuron $n_{k,l}$, and eventually the 
process terminates at some $x^*$ which is deemed to be a preferred input stimulus for the neuron in question. By visualising the hidden representations, it provides insights into how the neural network processes an input. 

\subsubsection{Inverting Representation}
\cite{MV2014} computes an approximate inverse of an image representation. Let  $rep:\real^{s_1} \rightarrow \real^{a_1}$ be a representation function, which maps an input to a representation of $a_1$ dimensions. Now, given a representation $T_0$ to be inverted, the task is to find an input $x^*$ such that 
\begin{equation}\label{equ:invert}
x^* = \arg\min_{x\in \real^{s_1}} \lossfunction(rep(x),T_0) + \lambda \regulariser(x)
\end{equation}
where $\lossfunction$ is a loss function measuring the difference of two representations (e.g.,  Euclidean norm), $\regulariser(x)$ is a regularisation term, and $\lambda$ is a multiplier to balance between the two terms. 
For the regulariser, they use either an image prior
\begin{equation}
\regulariser_\alpha(x) = \distance{x}{\alpha}^\alpha = \sum_{i=1}^{s_1} |x_i|^\alpha
\end{equation}
where $\alpha = 6$ is taken in their experiments, or a total variation, which encourages  images to consist of piece-wise constant
patches. 
To solve the above optimsation problem (\ref{equ:invert}), the classical gradient descent approach can be applied with the use of momentum to escape from saddle points. \cite{MV2014} argues that, while there may be no unique solution to this problem, sampling the space of possible reconstructions can be used to characterise the space of images that the representation deems to be equivalent, revealing its invariances. 

\cite{DB2015} proposes to analyse which information is preserved by a feature representation and which information is discarded. Given a feature vector, they train a DNN to predict the expected pre-image, i.e., the weighted average of all natural images which could have produced the given feature vector. Given a training set of images and their features $\{x_i,\phi_i\}$, the task is to learn the weight $w$ of a network $f(\phi,w)$ to minimise a Monte-Carlo estimate of the loss: 
\begin{equation}
\hat{w} = \arg\min_w \sum_i ||x_i - f(\phi_i,w) ||_2^2
\end{equation}

As a side note, it has been argued in  \cite{YCNFL2015} that gradient-based approaches do not produce images that resemble natural images.

\subsection{Instancewise Explanation by Ranking}\label{sec:ranking}

The next major thread of research is to  compute a ranking of a set of features. Specifically, given an input $\inputPoint$ and a set $\features$ of features, we aim to find a mapping $r_{\inputPoint}:\features\rightarrow \real$ 
such that each feature $\feature\in \features$ gets a rank $r_\inputPoint(\feature)$. Intuitively, the ranking $r_\inputPoint$ provides a total order among features on their contributions to a decision made by $\networks$ on the input $\inputPoint$. 

\subsubsection{Local Interpretable Model-agnostic Explanations (LIME)}

Local Interpretable Model-agnostic Explanations (LIME) \cite{LIME} interprets individual model prediction by locally approximating the model around a given prediction. Given an image $x$, LIME treats it as a set of superpixels. To compute the rank, it minimises the following objective function: 
\begin{equation}
\explain(f,x) = \arg\min_{e\in \explanations} \lossfunction(f,e,\pi_{x}) + \regulariser(e)
\end{equation}
where $\lossfunction(f,e,\pi_{x})$ is a loss of taking the explanation $e$ with respect to the original model $f$, under the condition of a local kernel $\pi_{x}$. The reguralisation term $\regulariser(e)$ is to penalise the complexity of the explanation $e$. 
%
For instance, if $G$ is the class of linear models with weights $w_e$, $\regulariser(e)$ can be chosen as $L_0$ norm $||w_e||_0$ to encourage sparsity, and the explainer $\explain(f,x)$ is the sparse weights  which rank the importance of dominant features.

\subsubsection{Integrated Gradients}

\cite{STY2017} suggests that a good explanation should satisfy the following two axiomatic attributes: 
\begin{itemize}
\item sensitivity - for every input and baseline that differ in one feature but have different predictions, the differing feature is given a non-zero attribution. 
\item implementation invariance - the attributions are always identical for two functionally equivalent networks.
\end{itemize}

They propose the use of integrated gradients and show that it is the only method that satisfies certain desirable axioms, including sensitivity and implementation invariance. Technically, the integrated gradient along the $i$-th dimension for an input $x$ and a baseline $x'$ is the following.
\begin{equation}
{\rm IntGrad}_i(x) = (x_i-x_i')\times \int_{0}^1 \frac{{\partial f(x'+\alpha\times(x-x'))}}{{\partial x_i}} d\alpha
\end{equation}
where $\frac{{\partial} f(x)}{{\partial x_i}}$ is the gradient of $f(x)$ along the $i$-th dimension of input $x$. 
The quantity ${\rm IntGrad}_i(x)$ is used to indicate the contribution of $x_i$ to the prediction $f(x)$ relative to a baseline input $x'$. If $f(\cdot)$ is differentiable everywhere, then
\begin{equation}
\sum_{i=1}^{s_1} {\rm IntGrad}_i(x)  = f(x) - f(x'),
\end{equation}
where $s_1$ is the input dimensions. For a given baseline $x'$ subject to $f(x') \approx 0$, an explainer can distribute the output to the individual features of the inputs.

\subsubsection{Layer-wise Relevance Propagation (LRP)}

Given an input-output mapping (for example a DNN) $f : \real^{s_1} \rightarrow \real$, the layer-wise relevance propagation (LRP) method \cite{LRP} is a concept of pixel-wise decomposition 
\begin{align}
f(x) \approx \sum_{i=1}^{s_1} R_i
\end{align}
to understand how much a single pixel $i$ contributes to the prediction $f(x)$. 
By propagating backward the prediction probability of the input through DNN and calculating the relevance scores, LRP attributes the importance scores to the pixels. Intuitively, the importance score, $R_i$, represents the local contribution of the pixel $d$ to the prediction function $f(x)$.
It is suggested in \cite{SGK2017} that LRP is equivalent to the Grad $\odot$ Input method (to be reviewed in Section~\ref{sec:gradientBasedMethod}) when the reference activations of all neurons are fixed to zero. 

\subsubsection{Deep Learning Important FeaTures (DeepLIFT)}

\cite{SGK2017} is a recursive prediction explanation method. It attributes to each input neuron $x_i$ a value $C_{\Delta x_i\Delta y}$ that represents the effect of that input neuron being set to a reference value as opposite to its original value. The reference value, chosen by the user, represents a typical uninformative background value for the feature. DeepLIFT uses a ``summation-to-delta'' property that states: 
\begin{equation}
\sum_{i=1}^{s_1} C_{\Delta x_i\Delta y} = \Delta o, 
\end{equation}
where $o=f(x)$ is the method output, $\Delta o=f(x)-f(r)$, $\Delta x_i = x_i-r_i$, and $r$ is the reference input. 
DeepLIFT improves over the canonical gradient-based methods by placing meaningful importance scores even if the gradient is zero, and avoiding discontinuities due to bias terms.

\subsubsection{Gradient-weighted Class Activation Mapping (GradCAM)}
By flowing the gradient information into the last convolutional layer of CNNs, Gradient-weighted Class Activation Mapping (GradCAM) \cite{CAM} computes a feature-importance map (i.e., a coarse localization) highlighting regions in the image corresponding to a certain concept.
Specifically, GradCAM computes the feature importance scores of a feature map $k$ for a class $c$ as follows.
\begin{align}
\alpha_k^c = \frac{1}{Z} \sum_i \sum_j \frac{\partial y^c}{\partial A_{ij}^k}
 \end{align}
where $\frac{\partial y^c}{\partial A_{ij}^k}$ is the gradients of the score via backpropagation with respect to feature maps $A^k$ of a convolutional layer, and the sums aim at global average pooling. The weights of the feature maps are used to indicate the importance of the input.
Intuitively, $\alpha_k^c $ represents a partial linearization of the deep network downstream from $A^k$, and captures the
‘importance’ of feature map $k$ for a target class $c$. 
Furthermore, Guided Grad-CAM obtains the more fine-grained feature importance scores, by multiplying the feature importance scores obtained from Grad-CAM with those from Guided Backpropagation in an elementwise manner.

\subsubsection{SHapley Additive exPlanation (SHAP)}

SHapley Additive exPlanation (SHAP) \cite{LL2017} suggests taking an additive model $$
g(z') = \phi_0 + \sum_{i=1}^M \phi_i z_i'
$$ 
as an explanation model, where $M$ is the number of simplified input features, $z'\in \{0,1\}^M$, and $\phi_i\in \real$ is a coefficient. It shows that under three properties (local accuracy, missingness, consistency), there is only one possible explanation model as follows.
\begin{equation}
\explain_i(f,x) = \sum_{z' \subseteq x'} \frac{|z'|! (M-|z'|-1)!}{M!} [f_x(z')-f_x(z'\setminus i)]
\end{equation}
where $|z'|$ is the number of non-zero entries in $z'$, and $z'\subseteq x'$ represents all $z'$ vectors where the non-zero entries are a subset of the non-zero entries in $x'$. Intuitively, $\explain_i(f,x)$ -- the Shapley values from cooperative game theory --  is an importance value to each feature $i$ that represents the effect on the model prediction of including that feature.

\subsubsection{Prediction Difference Analysis}
\cite{PDA} presents a prediction difference analysis (PDA) method to visualise the influence corresponding to a special input change or element removal. The idea is to assign a relevance value to each input feature w.r.t. a class $c$, so that the influence in terms of prediction difference can be measured by evaluating the difference between two conditional probabilities $p(c|x)$ and $p(c|x_{-i})$, where
\begin{equation}
p(c|x_{-i}) = \sum_{x_i} p(x_i|x_{-i}) p(c| x)
\end{equation}
calculates the conditional probability if $x_i$ is removed from the input $x$.

\subsubsection{Testing with Concept Activation Vector (TCAV)}

\cite{KWGCWVS2018} argues that, since most machine learning models operate on features, such as pixel values, that do not correspond to high-level concepts that humans understand, existing ranking-based approaches do not produce an explanation that can be easily accessible to humans. It then works on the idea of supplementing high level concepts (i.e., concept activation vector, CAV) into this explanation to make the  final explanation closer to human understandable explanation. The high-level concepts are learned independently from other user-provided data. Specifically, Testing with CAV (TCAV) uses directional derivatives 
\begin{equation}
S_{C,k,l}(x) = \nabla h_{l,k}(f_l(x)) \cdot v_C^l 
\end{equation}
to rank a user-defined concept with respect to a classification result, according to TCAV score
\begin{equation}
{\rm TCAV}_{Q_{C,k,l}} = \frac{|\{x \in X_k: S_{C,k,l}(x)>0 \}|}{|X_k|}
\end{equation}
where $C$ is a concept,  $l$ is a layer, $k$ is a class label, $h_{l,k}:\real^{s_l}\rightarrow \real$  maps the activations at layer $l$ to the logit output (i.e., the layer $K-1$) of a class $k$, $v_C^l \in\real^{s_l}$ is a unit CAV vector for a concept at layer $l$, and $f_l(x)$ is the activation for input $x$ at layer $l$.

\subsubsection{Learning to Explain (L2X)}
Learning to explain (L2X) \cite{CSWJ2018} utilises an instancewise feature selection for model interpretation. Roughly speaking, for a given input $\inputPoint$, among a set of $n$ features, it is to choose $k$ most informative features, which are of most relevance to the output $y=f(x)$. Let $\mathscr{P}_k=\{S\subset 2^{s_1}~|~|S| = k\}$ be the set of all subsets of size $k$. An explainer $\explain(f,\cdot)$ of size $k$ is a mapping from the feature space $\real^{s_1}$ to the power set $\mathscr{P}_k$.  Then, by modeling the input and the output as random variables $X$ and $Y$ respectively, the explanation problem is to find a solution to the following optimsiation problem: 
\begin{equation}
\max_{e} I(X_S;Y) \text{   subject to } S\sim e(X)
\end{equation}
where $I(X_S;Y)$ represents the mutual information between the selected features $X_S$ and the output $Y$, and $e(\cdot)$ is the specific feature selection explainer. Intuitively, the explainer is to find a set of features that maximise the statistical dependence. between selected features and the output. 

%



\subsection{Instancewise Explanation by Saliency Maps}\label{sec:saliencymaps}

While the ranking based methods are often used to generate saliency maps for visualisation purpose, the methods surveyed in this subsection is to compute saliency maps without computing a ranking score. 

\subsubsection{Gradient-based Methods}\label{sec:gradientBasedMethod}

{\bf Gradient} \cite{SVZ2013}.
Given an image classification model, let $S_c(x)$ be a score function for an image $x$ with respect to the class $c$. By back-propagation method, it aims to find an input locally optimising
\begin{equation}
\max_x S_c(x) - \lambda ||x||_2^2
\end{equation}
where $S_c(x)$ can be approximated to a linear function $S_c(x)=w^T x + b$ by first-order Taylor expansion in the neighbourhood of a reference image $x_0$, so that $w_c(x_0) = \frac{\partial S_c}{\partial x} \big |_{x_0}$ serves as an explanation map. Such gradient indicates the difference that a tiny change of each pixel of the input $x$ affects the classification score $c$.

To sharpen the sensitive maps, {\bf SmoothGrad} \cite{STKVW2017} randomly perturbs the input $x$ with a small noise and averages the resulting explanation maps, i.e., $\hat{w}_c(x) = \frac{1}{n} \sum_{i=1}^n w_c(x + g_i)$,
where $g_i \sim \mathcal{N}(0,\sigma^2)$ is a Gaussian noise enabling random sampling around $x$. 

Moreover, the {\bf Grad $\odot$ Input} \cite{SGK2017} method yields another explanation $\hat{w}_c(x)=x \odot w_c(x)$ to reduce visual diffusion.
{\bf DeConvNet} \cite{ZF2013} highlights the portions of a particular image that are responsible for the firing of each neural unit. 
{\bf Guided Backpropagation} \cite{SDB2014} builds upon DeConvNet and sets negative gradients to zero during backpropagation.


\subsubsection{Perturbation-based Methods}

\cite{DG2017}  proposes an accurate saliency detection method that manipulates the classification output by masking certain regions of the input image. Two concepts of saliency maps were considered: a smallest sufficient region (SSR) that allows for a confident classification, and a smallest destroying region (SDR) that prevents a confidence classification if removed. For instance, it applies a mask (e.g.,  a binary matrix) to the image data matrix by e.g., element-wise product, so as to set certain regions of the image to zero. A masking model can be trained to find such a mask for any input image in a single feed-forward pass.

\cite{FV2017} proposes a general framework that inspects how the output of a black-box model is affected by an input perturbation. Given a mask function $m: \Lambda \mapsto [0,1]$, a meaningful perturbation by constant mapping, adding noise, and blurring over a pixel $u \in \Lambda$ can be respectively defined by
\begin{align}
\Phi(u) =\left\{ \begin{array}{ll} m(u)x(u)+(1-m(u))\mu, & \text{ Constant mapping}\\
m(u)x(u)+(1-m(u))\eta(u), & \text{ Noising}\\
\int g(v-u)x(v) {\rm d}v, & \text{ Blurring}
\end{array} \right.
\end{align}
where the constant $\mu$ is the average colour, $\eta(u)$ is a random Gaussian noise i.i.d. over pixels, and $g(\cdot)$ is a Gaussian blur kernel.

\subsection{Model Explanation by Influence Functions}\label{sec:influencefunction}

The empirical influence function, a classic technique from robust statistics, is a measure of the dependence of the estimator on the value of one of the points in the sample. It has been utilised to interpret  the effects of training data to the model.  \cite{KL2017} considers influence functions to trace a model’s prediction through the
learning algorithm and back to its training data, without retraining the model. Basically, it can be done by e.g., upweighting a training point or perturbing a training input. 
For instance, given the training points $\{z_i=(x_i,y_i)\}_{i=1}^n$ and the loss function $\lossfunction(z,\theta)$ w.r.t. parameters $\theta \in \Theta$, the idea of unweighting is to compute the parameter change with respect to an infinitesimal $\epsilon$,
\begin{equation}
\hat{\theta}_{\epsilon,z} = \arg \min_{\theta \in \Theta} \frac{1}{n} \sum_{i=1}^n \lossfunction(z_i,\theta) + \epsilon \lossfunction(z,\theta)
\end{equation}
and evaluate the influence by 
\begin{equation}
\mathcal{I}(z) = \frac{d \hat{\theta}_{\epsilon,z}}{d \epsilon} \Big |_{\epsilon=0} = -H_{\hat{\theta}}^{-1} \nabla_{\theta} \lossfunction(z, \hat{\theta})
\end{equation}
where $H_{\hat{\theta}}=\frac{1}{n} \sum_{i=1}^n \nabla_{\theta}^2 L(z_i, \hat{\theta})$ is the Hessian.
The influence of the removal of one specific training point $z$ from the training set can be directly evaluated by calculating the parameter change
\begin{equation}
\hat{\theta}_{-z} - \hat{\theta} \approx -\frac{1}{n} \mathcal{I}(z)
\end{equation}
where $\hat{\theta}_{-z}=\arg \min_{\theta \in \Theta} \frac{1}{n} \sum_{z_i \ne z} L(z_i,\theta)$ is the new parameter due to the removal of the training point $z$.

\subsection{Model Explanation  by Simpler Models}\label{sec:simplermodels}

Interpretability can be achieved by approximating the neural network (either a feedforward neural network or a recurrent neural network) with a simpler model (or a set of simpler models), on which an intuitive explanation can be obtained. We remark that, some local interpretability methods in Section~\ref{sec:ranking} -- such as LIME and SHAP -- computes a simpler model, e.g., an additive linear moel, for every input sample. Because the simpler models are different for different input samples, we do not consider them as model explanation methods, which intend to use a single simpler model to explain the neural network (and its decision on all input samples).


\subsubsection{Rule Extraction}

\cite{Anchors} ties a prediction locally to a decision rule (represented as a set of predicates), based on the assumption that while the model is globally too complex to be explained succinctly, ``zooming in'' on individual predictions makes the explanation task feasible. Let $A$ be a rule, such that $A(x)=1$ if all its feature predicates are true for the input $x$. $A$ is an anchor (i.e., a valid explanation) if 
\begin{equation}
\expectation_{{\cal D}(z|A)}[{f(x)=f(z)}] > 1-\epsilon \text{ and } A(x)=1
\end{equation}
where ${\cal D}(\cdot|A)$ is the conditional distribution when the rule $A$ applies, and $f(x)=f(z)$ represents that the inputs $x$ and $z$ have the same label. Intuitively, it requires that for inputs on which the anchor holds, the prediction is (almost) always the same. 

\subsubsection{Decision Tree Extraction}

Extraction of decision tree from complex models is a popular thread of research. An example of this is \cite{DeepRED}, which interprets DNN by a decision tree representation, in which decision rules are extracted from each layer of DNN by decomposition. It is based on Continuous/discrete Rule Extractor via Decision tree induction (CRED) \cite{CRED} and C4.5 decision tree learning algorithm. 

\subsubsection{Linear Classifiers to Approximate Piece-wise Linear Neural Networks}

\cite{CHHWP2018} interprets piecewise linear neural networks (PLNN) by a set of locally linear classifiers.
In PLNN, activation functions are piecewise linear such as ReLU. Given an input $x$, its corresponding activation pattern, i.e., the states of all hidden neurons, is fixed and equivalent to a set of linear inequalities. Each inequality represents a decision feature, and therefore the interpretation of $x$ includes all the decision features. Moreover, because all inputs sharing the same activation pattern form a unique convex polytope, the interpretation of an input $x$ can also include the the decision features of the polytope boundaries.

\subsubsection{Automata Extraction from Recurrent Neural Networks}

\cite{WGY2017} extends the L* automata learning algorithm \cite{Angluin1987} to work with recurrent neural networks (RNNs). L* algorithm requires a mechanism to handle two types of queries: membership query and equivalence query. Let $A$ be the learning automaton. For membership query, the RNN classifier is used to check whether an input is correctly classified. For equivalence query, it takes an automaton $A^p$ generated by \cite{OG1996} and check if $A$ and $A^p$ are equivalent. If there is a disagreement between the two automata, i.e., an input has different classes, it will determine whether to return the input as a counterexample or to refine the automata $A$ and restart the comparison. This process iterates until it converges (i.e., $A$ and $A^p$ are equivalent) or a specific  limit has been reached.

\subsection{Information-flow Explanation by Information Theoretical Methods}\label{sec:informationtheory}

Information theory is increasingly believed to be one of the most promising tools to open the black-box of DNN. The key building block is an information bottleneck method originally proposed by \cite{IB}.
\subsubsection{Information Bottleneck Method}
Given two discrete real-valued variables $X$ and $Y$ with joint distribution $p(x,y)$, the objective of the information bottleneck (IB) method \cite{IB} is to figure out the stochastic encoder $p(t | x)$, i.e.,
\begin{subequations}
\begin{align}
\text{IB: } \min_{p(t|x)} &\quad \mathcal{L}_{\rm{IB}} = I(X; T) - \beta I(T;Y) \\
\text{   subject to } &\quad T \leftrightarrow X \leftrightarrow Y
\end{align}
\end{subequations}
for $\beta > 1$, where $T \leftrightarrow X \leftrightarrow Y$ forms a Markov chain and $I(X;T)$ represents the mutual information between variables $X$ and $T$. Otherwise if $\beta \le 1$, the optimal solution is degenerated $I(X; T) = I(T;Y)=0$.
The information-flow explanation $\explain(\mathcal{F})$ is the solution to the above optimisation problem, stochastic encoder $p(t|x)$ and decoder $p(y|t)$, which can be given by
\begin{align}
p(t|x) &= \frac{p(t)}{Z(x,\beta)} \exp[-\beta D_{KL}[p(y|x) \; | \; p(y|t)]] \\
p(y|t) &= \frac{1}{p(t)} \sum_x p(t|x) p(x,y)
\end{align}
where the normalisation factor $Z(x,\beta) $ is given by
\begin{align}
Z(x,\beta) = \exp[\frac{\lambda(x)}{p(x)} - \beta \sum_y p(y|x) \log \frac{p(y|x)}{p(y)}]
\end{align}
and $\lambda(x)$ is a Lagrange multiplier. The computation of the above quantities can be obtained by applying the iterative Blahut-Arimoto algorithm, which however becomes computationally intractable when $p(x,y)$ is high-dimensional and/or non-Gaussian distributed.

\subsubsection{Information Plane}
Given a $K$-layer deterministic DNN with $T_k$ being a multivariate random variables representing the output of $k$-th hidden layer, \cite{openbox-IB} introduced the concept of information plane with coordinates $(I(X;T_k),I(T_k,Y))$.\footnote{According to \cite{GBGMNKP2018},
the quantity computed and plotted in \cite{openbox-IB} is actually $I(X; {\rm Bin}(T_k))$ rather than $I(X;T_k)$, where ${\rm Bin}(\cdot)$ is ``a per-neuron discretisation of each hidden activity of $T_k$ into a user-selected number of bins.''} 

By assuming the Markov chain of $K$-layer DNN, the information bottleneck method turns to
\begin{subequations}
\begin{align}
\text{IB: } \min_{p(t_k|x)} &\quad \mathcal{L}_{\rm{IB}} = I(X; T_k) - \beta I(T_k;Y) \\
\text{   subject to } &\quad Y \leftrightarrow X \leftrightarrow T_1 \leftrightarrow \dots \leftrightarrow T_K \leftrightarrow \hat{Y}
\end{align}
\end{subequations}
where the coordinates of the information planes follow
\begin{align}
I(X;Y) \ge I(T_1;Y) \ge \dots \ge I(T_K;Y) \ge I(\hat{Y};Y)\\
H(X) \ge I(X;T_1) \ge \dots \ge I(X;T_K) \ge I(X;\hat{Y}).
\end{align}

\begin{figure}[h]
\center
\includegraphics[width=3.2in]{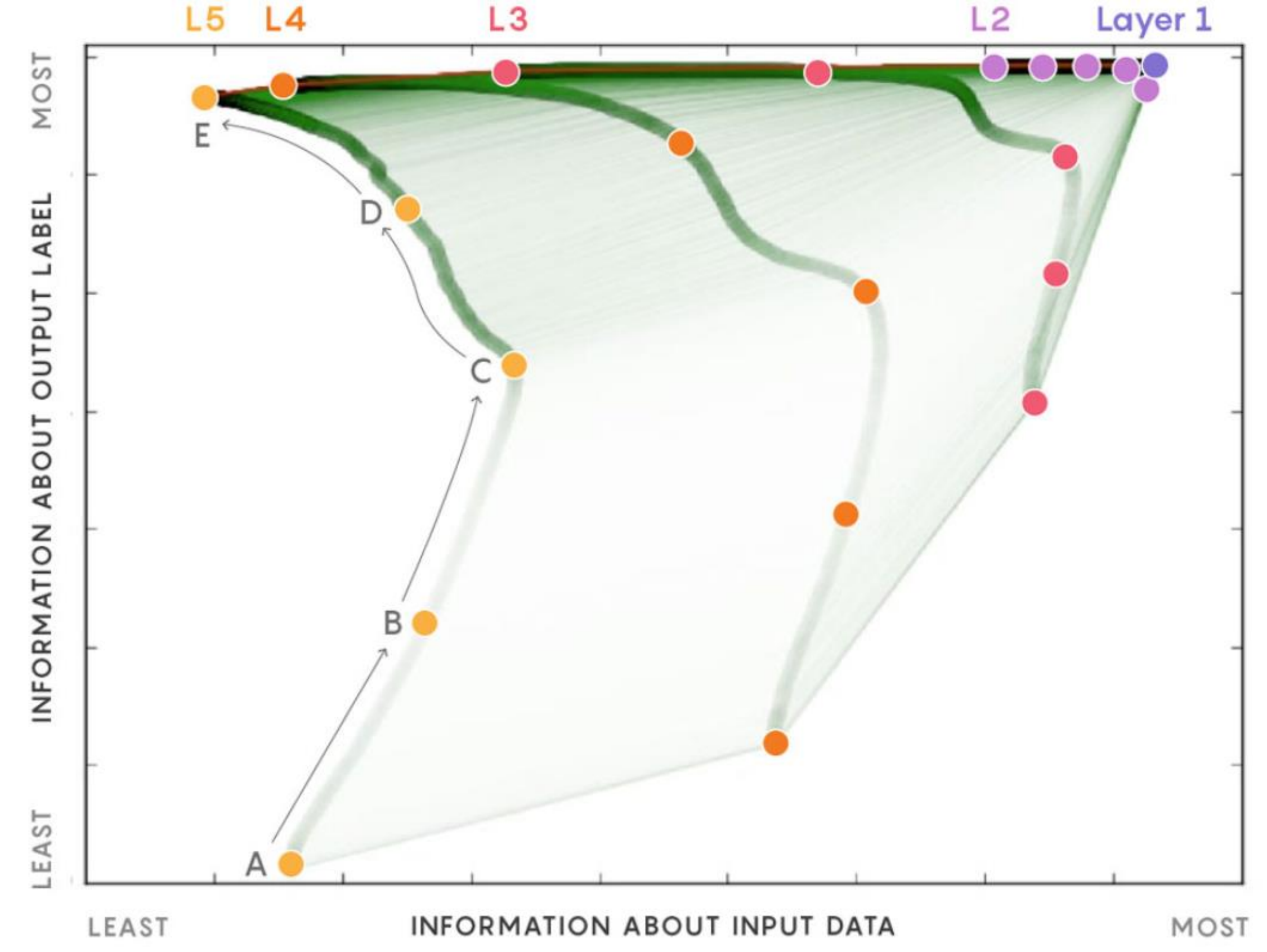}
\caption{The evolution of Information Plane $(I(X;T_k),I(T_k;Y))$ during the training procedure of a 5-layer DNN \cite{quantamagazine},
where A-B-C indicates the fitting phase, and C-D-E is the compression phase.}
\label{fig:inforplane}
\end{figure}

In doing so, DNN training procedure can be interpreted by visualising the information planes with respect to the SGD layer dynamics of the training. It was suggested by \cite{openbox-IB} that the training procedure consists of two phases:
\begin{itemize}
\item Fitting Phase: The neurons in higher layers learn as much information as possible about the inputs, and try to fit it to the outputs. In this phase, both $I(X;T_k)$ and $I(T_k;Y)$ increase, indicating that neurons are learning.
\item Compression Phase: Higher layer neurons try to compress their representation of the input, while keep the most relevant information to predict the output. In this phase, while $I(T_k;Y)$ still keeps increasing, $I(X;T_k)$ is decreasing, which indicates the neurons try to forget what they have learnt in the fitting phase but only retain the most relevant information for the output prediction.
\end{itemize}

It was also claimed in \cite{openbox-IB} that the compression phase is responsible for the generalisation performance of DNN.
Recently, such a fresh perspective for interpreting DNN training process has attracted a lot of attention, and also caused the ongoing debate on whether or not it is possible to translate theoretical insights into DNN optimisation in practice.

\subsubsection{From Deterministic to Stochastic DNNs}

It is argued by several papers (e.g.,  \cite{Saxe2018,Amjad2018,GBGMNKP2018}
) that the IB method suffers from some issues on explaining the training process of deterministic DNNs. 
The main issue is that, the quantity $I(X,T_k)$, whose decrease leads to compression, will not change because it is either a constant value (when $X$ is discrete) or infinite (when $X$ is continuous), given a deterministic DNN.

To resolve this issue, several stochastic DNN frameworks have been proposed to transform deterministic DNNs to stochastic ones and make the compression term $I(X,T_k)$ more meaningful.
For instance, \cite{Saxe2018} suggested injecting an independent Gaussian noise with user-defined variance to $T_k$ followed by quantisation.
\cite{Amjad2018} suggested including a (hard or soft) decision rule to increase the dynamicity or replacing $I(X,T_k)$ by other similar yet well-behaved cost functions.
\cite{GBGMNKP2018} suggested relating the mapping $X \mapsto T_k$ to the input and the output of a stochastic parameterised communication channel with parameters being DNN's weights and biases.

This is still an active ongoing research to interpret how DNNs get trained from an information-theoretic perspective, in the hope to guide DNN practice.

\subsection{Summary}
In this section, we reviewed the main approaches for DNN interpretability from data, model, and information perspectives. They are not mutually-exclusive but complement, and some techniques can be combined to provide instance-wise, model, and/or information-flow explanation in different dimensions.

Table~\ref{tab:interpretability} summarises some papers according to three aspects, i.e., explanation subject, explanation scope, and whether the method is model agnostic or not. Explanation scope can be either local or global. For local explanation, it only considers the function $f$ working on the local region close to an instance. For global explanation, it is to explain the entire input domain. 

\begin{table*}[h]
\scriptsize
    \caption{Comparison between different interpretability techniques}
    \label{tab:interpretability} 
    \centering
    \scalebox{0.9}{
    \def\arraystretch{1.2}
    \begin{tabular}{|p{40mm}||p{30mm}|p{25mm}|p{20mm}|p{15mm}|p{20mm}|}
    \toprule
    
     &  explanation subject & explanation scope & model agnostic \\ 
     \hline
\cite{EBCV2009} & instancewise & local & No\\ 
\cite{MV2014} &instancewise & local & No \\
\cite{DB2015} &instancewise & local & No \\
\hline
\cite{LIME} &instancewise & local & Yes \\
\cite{STY2017} & instancewise & local & No \\
\cite{LRP} & instancewise & local  & No \\
\cite{SGK2017} & instancewise & local & No \\
\cite{CAM} & instancewise &  local & No \\
\cite{LL2017} & instancewise & local & Yes \\
\cite{PDA} & instancewise & local & Yes \\
\cite{KWGCWVS2018} & instancewise & local & No \\
\cite{CSWJ2018} & instancewise & local & No \\
\hline
\cite{SVZ2013} & instancewise & local & No \\
\cite{STKVW2017} & instancewise & local  & No \\
\cite{SGK2017}  & instancewise & local & No \\
\cite{ZF2013} & instancewise & local & No \\
\cite{SDB2014}  & instancewise & local & No \\
\cite{DG2017} & instancewise & local & Yes \\
\cite{FV2017}  & instancewise & local & Yes \\
\hline
\cite{KL2017} & Model & global & No \\
\hline
\cite{Anchors} & Model & local & Yes \\
\cite{DeepRED} &Model & global & No \\
\cite{CHHWP2018} & Model & local & No \\
\cite{WGY2017}  & Model & global & Yes \\
\hline
\cite{IB} & Information flow & global & Yes \\
\cite{openbox-IB} & Information flow &global  & Yes \\
\cite{Saxe2018}  & Information flow & global  & Yes \\
\cite{Amjad2018} & Information flow &global  & Yes \\
\cite{GBGMNKP2018} & Information flow & global  & Yes \\

    \bottomrule
   \end{tabular}
   }
\end{table*}
\newpage
\section{Future Challenges}
\label{sec:challenges}

Research on enhancing the safety and trustworthiness of DNNs  is still in its infancy. In the following, we discuss a few prominent challenges.

\subsection{Distance Metrics closer to Human Perception}

Distance metrics are key building blocks of the techniques we surveyed, and are mainly used to measure the similarity between two inputs. 
Ideally, two inputs with smaller distance should be more similar with respect to human perception ability. Given the fact that it is hard to measure human perception ability, 
well-known metrics including $L_0$, $L_1$, $L_2$, and $L_\infty$-norms, are applied. While most techniques are orthogonal to the consideration of distance metrics (under the condition that the distances can be efficiently computed), it has been argued that better metrics may be needed. For example, \cite{NOneedtoworry} argued that adversarial examples found by a few existing methods (mostly based on $L_2$ or $L_\infty$) are not necessarily physical, and \cite{YCNFL2015} claims that  gradient-based approaches (where the computation of gradients are usually based on certain norm distance) for interpretability do not produce images that resemble natural images.

\cite{Lipton2016} states that the demand for interpretability arises when
there is a mismatch between the formal objectives of supervised
learning (test set predictive performance) and the real
world costs in a deployment setting. Distance metrics are key components in various training objectives of supervised learning. Actually, for DNNs on perception tasks, it is the mismatch of distance metrics used in the training objective and used by human to differentiate objects that hinders an intuitive explanation of a given decision.

For pattern recognition and image classification, there are other image similarity distances proposed, such as the structure similarity measure SSIM \cite{WSB2003}, but they are restricted to the computational efficiency problem and the existence of analytical solution when computing gradients. It will be interesting to understand if the application of such metrics become more practical for the tasks we are concerned with when more efficient computational methods such as \cite{BV2017} become available. 

\subsection{Improvement to Robustness} 

Traditional verification techniques aim to not only verify (software and hardware) systems when they are correct, but also find counterexamples whenever they are incorrect. These counterexamples can be used to either directly improve the system (e.g.,  counterexample-guided abstract refinement \cite{CEGAR}, etc) or to provide useful information to the system designers for them to improve the system. This is similar for the software testing, where bugs are utilised to repair implementation errors (e.g., automatic repair techniques \cite{KB2013}, etc) and for the programmers to improve their implementation. While similar expectation looks reasonable since existing DNN verification and testing tools are able to find e.g., counterexamples to local robustness (i.e., adversarial examples), it is relatively unclear how these counterexamples can be utilised to improve the correctness (such as the local robustness) of the DNN, other than the straightforward method of adding the adversarial examples into the training dataset, which may lead to  bias towards those input subspaces with adversarial examples. Section~\ref{sec:defence} reviews a few techniques such as adversarial training.

Another pragmatic way can be to design a set of quantitative metrics to compare the robustness of  DNNs with different architectures (for example different numbers and types of hidden layers) so that a DNN designer is able to diagnose the DNN and figure out a good architecture for a particular problem. Relevant study has been started in e.g., \cite{sun2018testing-b,sun2018concolic}, with some preliminary results reported in the experiments. A significant next step will be to automate the procedure and synthesise an architecture according to some given criteria. 

\subsection{Other Machine Learning Models}

Up to now, most efforts are spent on feedforward DNNs, with image classification as one of the main tasks. Research is needed to consider other types of neural networks, such as deep reinforcement learning models \cite{MKSRVBG2015,SQAS2015,WdFL2015,vHGS2015} and recursive neural networks, and other types of machine learning algorithms, such as SVM, k-NN, naive Bayesian classifier, etc. 
%
Most deep reinforcement learning models use feedforward DNNs to store their learned policies, and therefore for a single observation (i.e., an input) similar safety analysis techniques can be applied. However, reinforcement learning optimises over the objectives which may base on the rewards of  multiple, or even an infinite number of, time steps. Therefore, other than  the DNN, the subject of study includes a sequence of, instead of one, observations. A few attack techniques, such as \cite{HPGDA2017,PTLBC2017,LHLSLS2017,TJF2018}, have been proposed, with the key techniques based on generalising the idea of FGSM~\cite{FGSM}. For recurrent neural networks, there are a few attack techniques such as \cite{WZS2018}. Less work has been done for verification, testing, and interpretability. 

\subsection{Verification Completeness}

Additional to the properties we mentioned in Section~\ref{sec:safetyproblem}, the correctness of a DNN may include other properties. More importantly, the properties in Section~\ref{sec:safetyproblem} are expressed in different ways, and for each property, a set of ad hoc analysis techniques are developed to work with them.  See e.g., Table~\ref{tab:verification} for a comparison of verification technqiues. 
%
%
%
Similar to the logic languages LTL (linear time logic) and CTL (computational tree logic) which can express various temporal properties related to the safety and liveness properties of a concurrent system, research is needed to develop a high-level language that can express a set of properties related to the correctness of a DNN. Such a language will enable the development of general, instead of ad hoc, formal analysis techniques to work with various properties  expressible with that language.  
The development of a high-level specification language for DNNs is hindered by the lack of specifications for data-driven machine learning techniques, which learn the model directly from a set of data samples. A possible direction can be to obtain specifications from the training data, e.g., by studying how the data samples are scattered in the input high-dimensional space. 

\subsection{Scalable Verification with Tighter Bounds}

Existing verification approaches  either cannot scale to work with state-of-the-art networks (e.g., for constraint-solving based approaches) or cannot achieve a tight bound (e.g., for over-approximation approaches). After the initial efforts on conducting exact computation, such as \cite{HKWW2017,katz2017reluplex,LM2017,Rudiger2017}, recent efforts have been on approximate techniques to alleviate the computational problem while still achieve tangible bounds, e.g., \cite{wicker2018feature,RHK2018,GMDTCV2018}. Significant research efforts are required  to achieve tight bounds with approximate techniques for state-of-the-art DNNs. It can be hard to work with real-world models which usually contain complex structures and lots of real-valued parameters. A possible direction will be to develop an abstraction and refinement framework, as \cite{CGJLV2000} did for concurrent systems, and it will be interesting to see how it is related to the layer-by-layer refinement \cite{HKWW2017}. 

\subsection{Validation of Testing Approaches}\label{sec:validationTesting}

Up to now, testing DNNs is mainly on coverage-based testing, trying to emulate the structural coverage testing developed in software testing. However, different from traditional (sequential) software in which every input is associated with a single activated path and eventually leads to an output, in DNNs every input is associated with a large set of activated paths of neurons and the output is collectively determined by these paths, i.e., activation pattern \cite{sun2018testing,sun2018testingjournal}. Such semantic difference suggests that, a careful validation of the coverage-based testing is needed to make sure that the extended methods can work effectively in the context of DNNs. 

In particular, for most proposals up to now, neurons are treated as the most basic elements in the coverage definitions and are regarded as the variables in the traditional software. However, unlike a variable which usually holds certain weight in determining the execution path, a single neuron in most cases cannot solely determine, or change, the activation path of an input. Therefore, the testing coverage based on neurons does not examine the actual functionality of DNNs. It can be reasonable to consider
emulating variables in traditional software with a set of neurons (instead of a single neuron) and therefore let 
 paths be the sequences of sets of neurons. A preliminary study appears  in  \cite{sun2018testing-b} with more sophisticated design on the coverage metrics required. 
The lift of the most basic element from neuron to a set of neurons will also affect the design of test case generation algorithms. Moreover, it is expected that interpretability techniques  can be employed as building blocks for test case generation algorithms. 

\subsection{Learning-Enabled Systems}

As suggested in e.g., \cite{Sifakis2018}, real-world autonomous systems contain both logic components -- to handle e.g., high-level goal management, planning, etc. -- and data-drive learning components -- to handle e.g., perception tasks, etc. To analyse such learning-enabled systems, methods are needed to interface the analysis techniques for individual components. Compositional and assume-guarantee reasoning can be applied in this context. Significant efforts are needed to be on a few topics  including e.g., how to utilise logic components to identify safety properties of DNN components (e.g., a set of constraints DNN components need to satisfy, a subspace of the input domain needed to be considered, etc), how the safety assurance cases of DNN components can be streamlined into the assurance cases of the whole system, etc.

Verification techniques have been developed to work with learning-enabled systems. In \cite{dreossi2017systematic}, a compositional framework is developed for the falsification of temporal logic properties of cyber physical systems with ML components. Through experiments on a dynamic tracking system based on high resolution imagery input, \cite{SZMSH2020} studies the uncertainties eliminated and introduced from the interaction between components. Based on this, \cite{HZMSMH2020} suggests a formal verification technique to work with robustness and resilience of learning-enabled state estimation systems. 

For the safety assurance of learning-enabled systems, it would be useful to understand how low level evidences -- obtained via e.g.,  verification, testing, etc.  -- can be utilised to reason about high level assurance goals such as the system can operate correctly in the next 100 days with probability more than 99\%. Research on this has started in \cite{2020arXiv200305311Z}, where a Bayesian inference approach is taken.

\subsection{Distributional Shift, Out-of-Distribution Detection,  and Run-time Monitoring}\label{sec:distributionshif}

DNNs are trained over a set of inputs sampled from the real distribution. However, due to its high-dimensionality, the training dataset may not be able to cover the input space. Therefore, although it is reasonable to believe that the resulting trained models can perform well on new inputs close to the training data, it is also understandable that the trained models might not perform correctly in those inputs where there is no neighbouring training data. 
While techniques are being requested to achieve better generalisability for DNN training algorithm including various regularisation techniques (see e.g., \cite{Goodfellow2016} for a comprehensive overview), as suggested in e.g.,~\cite{AOSCSM2016,box-clever, Moreno-Torres2012}, it is also meaningful (particularly for the certification of safety critical systems) to be able to identify those inputs on which the trained models should not have high confidence. Technically, such inputs can be formally defined as both  topologically far away from training data in the input space and being classified with high probability by the trained models. Moreover, \cite{2018arXiv180808282A} suggests having an extra class ``dustbin'' for such inputs. 

The distributional shift has been studied from the perspective of out-of-distribution detection, which checks whether an input 
is on the distribution from which the training data is sampled. Research on this direction has been active, see e.g.,  \cite{liang2018enhancing,2018arXiv180204865D}.

The ubiquity of experiencing distributional shift in the application of deep neural networks, and the difficulty of having the verification completeness, suggest the importance of developing run-time monitoring techniques to enable the detection of safety problems on-the-fly. Research is needed to understand which quantity or property is to monitor, although some suggestions have been made including e.g., the activation patterns \cite{2018arXiv180906573C}.

\subsection{Unifying Formulation of Interpretability} 

There is a lack of consistent definitions of interpretability, although a number of approaches have been surveyed (e.g., in \cite{Lipton2016}) from different angles. While the existing research is able  to provide various partial information about the DNNs, it is hard to compare them under a systematic framework. 
Recently, there were some attempts to accommodate several similar approaches to a single unified framework, to facilitate the comparison. For example, \cite{olah2017feature} suggested that some visualisation methods could be fitted in the optimisation problem with different regularisation terms, \cite{LL2017} used Shapley value to explain a few attribute-based approaches with an additive model, and \cite{ACOG2018} used a modified gradient function to accommodate a few gradient-based approaches. Although it may be difficult, if not impossible, to have a unified definition for interpretability, it is necessary to define it  in a consistent way for the purpose of comparison. It is hoped that our attempt in defining interpretability from different dimensions of data, model, and information could shed light on such a unifying formulation. 

\subsection{Application of Interpretability to other Tasks}

Except for the size of DNNs, the key difficulty of verifying, testing, or attacking DNNs is due to the fact that DNNs are black-box. It is therefore reasonable to expect that, the interpretation results  will be able to enable better verification and testing approaches. For testing, interpretability can potentially enhance, or refine, the design of coverage metrics and enable more efficient test case generation algorithms. For verification, it is expected that interpretability can help identify more specifications to enhance the verification completeness. The applications of interpretability in attack techniques have been seen in e.g., \cite{JSMA} and \cite{ruan2019global}, where a ranking over the input dimensions provides heuristics to find good adversarial examples.

\subsection{Human-in-the-Loop}\label{sec:humanintheloop}

All the techniques reviewed are to improve the trust of human users on the DNNs through the angles of certification and explanation. Certification techniques improve the confidence of the users on the correctness of the DNNs, and the explanation techniques increase human users' understanding about the DNNs and thus improve the trust. These can be seen as a one-way enhancement of confidence from the DNNs to the human users. The other direction, i.e., how  human users can help improve the trustworthiness of the DNNs, is less explored. There are only a few works such as  \cite{TKDB2017}, where  a visual analytic interface is presented to enable expert user by interactively exploring a set of instance-level explanations.

Trust is a complex notion, 
viewed as a belief, attitude, intention or behaviour, and is most generally understood as {a subjective evaluation of a truster on a trustee about something in particular}, e.g., the completion of a task \cite{Hardin:Trust:2002}.  A classical definition from organisation theory \cite{Mayer:AMR:1995} defines trust as {the willingness of a party to be vulnerable to the actions of another party based on the expectation that the other will perform a particular action important to the trustor, irrespective of the ability to monitor or control that party}. It is therefore reasonable to assume that the interaction between the DNNs and their human users can significantly affect the trust, and the trust is not a constant value and can be fluctuated. The formulation of the changes of trust in terms of the interaction has been started in \cite{HK2017} with a comprehensive reasoning framework.


\section{Conclusions}
\label{sec:concl}

In this survey, we reviewed  techniques to determine and improve the safety and trustworthiness of deep neural networks, based on the assumption that trustworthiness is determined by two major concepts: certification and explanation. This is a new, and exciting, area requiring expertise and close collaborations from multidisciplinary fields including formal verification, software testing, machine learning, and logic reasoning.

{\footnotesize
\setlength{\parskip}{1em}
DSTL/JA122942 This document is an overview of UK MOD (part) sponsored research and is released for informational purposes only. The contents of this document should not be interpreted as representing the views of the UK MOD, nor should it be assumed that they reflect any current or future UK MOD policy. The information contained in this document cannot supersede any statutory or contractual requirements or liabilities and is offered without prejudice or commitment.
\newline \indent
Content includes material subject to \textcopyright~Crown copyright (2020), Dstl. This material is licensed under the terms of the Open Government Licence except where otherwise stated. To view this licence, visit http://www.nationalarchives.gov.uk/doc/open-government-licence/version/3 or write to the Information Policy Team, The National Archives, Kew, London TW9 4DU, or email: psi@nationalarchives.gsi.gov.uk.
\par
}

\newpage
\bibliographystyle{apalike}
\bibliography{references}

\begin{thebibliography}{}

\bibitem[GDR, 2016]{GDRP}
 (2016).
\newblock General data protection regulation.
  {\url{http://data.europa.eu/eli/reg/2016/679/oj}}.

\bibitem[exp, 2018]{explainableAI}
 (2018).
\newblock Explainable artificial intelligence.
  {\url{https://www.darpa.mil/program/explainable-artificial-intelligence}}.

\bibitem[Tes, 2018]{TeslaIncident}
 (2018).
\newblock {NTSB} releases preliminary report on fatal {Tesla} crash on
  autopilot.
  {https://electrek.co/2018/06/07/tesla-fatal-crash-autopilot-ntsb-releases-preliminary-report/}.

\bibitem[Ube, 2018]{UberIncident}
 (2018).
\newblock Why {U}ber’s self-driving car killed a pedestrian.
  {\url{https://www.economist.com/the-economist-explains/2018/05/29/why-ubers-self-driving-car-killed-a-pedestrian}}.

\bibitem[{Abbasi} et~al., 2018]{2018arXiv180808282A}
{Abbasi}, M., {Rajabi}, A., {Sadat Mozafari}, A., {Bobba}, R.~B., and {Gagne},
  C. (2018).
\newblock {Controlling Over-generalization and its Effect on Adversarial
  Examples Generation and Detection}.
\newblock {\em ArXiv e-prints}.

\bibitem[Agarwal et~al., 2018]{agarwal2018automated}
Agarwal, A., Lohia, P., Nagar, S., Dey, K., and Saha, D. (2018).
\newblock Automated test generation to detect individual discrimination in ai
  models.
\newblock {\em arXiv preprint arXiv:1809.03260}.

\bibitem[Amjad and Geiger, 2018]{Amjad2018}
Amjad, R.~A. and Geiger, B.~C. (2018).
\newblock How (not) to train your neural network using the information
  bottleneck principle.
\newblock {\em CoRR}, abs/1802.09766.

\bibitem[Ammann and Offutt, 2008]{AO2008}
Ammann, P. and Offutt, J. (2008).
\newblock {\em Introduction to Software Testing}.
\newblock Cambridge University Press.

\bibitem[Amodei et~al., 2016]{AOSCSM2016}
Amodei, D., Olah, C., Steinhardt, J., Christiano, P., Schulman, J., and
  Man{\'e}, D. (2016).
\newblock Concrete problems in ai safety.
\newblock {\em arXiv preprint arXiv:1606.06565}.

\bibitem[Ancona et~al., 2018]{ACOG2018}
Ancona, M., Ceolini, E., Öztireli, C., and Gross, M. (2018).
\newblock Towards better understanding of gradient-based attribution methods
  for deep neural networks.
\newblock In {\em International Conference on Learning Representations}.

\bibitem[Angluin, 1987]{Angluin1987}
Angluin, D. (1987).
\newblock Learning regaular sets from queries and counterexamples.
\newblock {\em Information and Computation}, 75:87–106.

\bibitem[Ashmore and Hill, 2018]{box-clever}
Ashmore, R. and Hill, M. (2018).
\newblock Boxing clever: Practical techniques for gaining insights into
  training data and monitoring distribution shift.
\newblock In {\em First International Workshop on Artificial Intelligence
  Safety Engineering}.

\bibitem[Bach et~al., 2015]{LRP}
Bach, S., Binder, A., Montavon, G., Klauschen, F., Müller, K.-R., and Samek,
  W. (2015).
\newblock On pixel-wise explanations for non-linear classifier decisions by
  layer-wise relevance propagation.
\newblock {\em PLoS ONE}, 10(7).

\bibitem[Bastani et~al., 2016]{BILVNC2016}
Bastani, O., Ioannou, Y., Lampropoulos, L., Vytiniotis, D., Nori, A., and
  Criminisi, A. (2016).
\newblock Measuring neural net robustness with constraints.
\newblock In {\em Advances in neural information processing systems}, pages
  2613--2621.

\bibitem[Bhagoji et~al., 2017]{BCM2017}
Bhagoji, A.~N., Cullina, D., and Mittal, P. (2017).
\newblock Dimensionality reduction as a defense against evasion attacks on
  machine learning classifiers.
\newblock {\em CoRR}, abs/1704.02654.

\bibitem[Biggio et~al., 2013]{biggio2013evasion}
Biggio, B., Corona, I., Maiorca, D., Nelson, B., {\v{S}}rndi{\'c}, N., Laskov,
  P., Giacinto, G., and Roli, F. (2013).
\newblock Evasion attacks against machine learning at test time.
\newblock In {\em Joint European conference on machine learning and knowledge
  discovery in databases}, pages 387--402. Springer.

\bibitem[Bruni and Vitulano, 2017]{BV2017}
Bruni, V. and Vitulano, D. (2017).
\newblock An entropy based approach for ssim speed up.
\newblock {\em Signal Processing}, 135:198--209.

\bibitem[Buckman et~al., 2018]{BRRG2018}
Buckman, J., Roy, A., Raffel, C., and Goodfellow, I. (2018).
\newblock Thermometer encoding: One hot way to resist adversarial examples.
\newblock In {\em International Conference on Learning Representations}.

\bibitem[Bunel et~al., 2017]{bunel2017piecewise}
Bunel, R., Turkaslan, I., Torr, P.~H., Kohli, P., and Kumar, M.~P. (2017).
\newblock Piecewise linear neural network verification: A comparative study.
\newblock {\em arXiv preprint arXiv:1711.00455}.

\bibitem[Carlini and Wagner, 2017a]{carlini2017adversarial}
Carlini, N. and Wagner, D. (2017a).
\newblock Adversarial examples are not easily detected: Bypassing ten detection
  methods.
\newblock In {\em Proceedings of the 10th ACM Workshop on Artificial
  Intelligence and Security}, pages 3--14. ACM.

\bibitem[Carlini and Wagner, 2017b]{carlini2017magnet}
Carlini, N. and Wagner, D. (2017b).
\newblock Magnet and" efficient defenses against adversarial attacks" are not
  robust to adversarial examples.
\newblock {\em arXiv preprint arXiv:1711.08478}.

\bibitem[Carlini and Wagner, 2017c]{CW2016}
Carlini, N. and Wagner, D. (2017c).
\newblock Towards evaluating the robustness of neural networks.
\newblock In {\em Security and Privacy (SP), IEEE Symposium on}, pages 39--57.

\bibitem[Chen et~al., 2018]{CSWJ2018}
Chen, J., Song, L., Wainwright, M.~J., and Jordan, M.~I. (2018).
\newblock Learning to explain: An information-theoretic perspective on model
  interpretation.
\newblock In {\em International Conference on Machine Learning}.

\bibitem[Cheng et~al., 2018a]{cheng2018nn}
Cheng, C.-H., Huang, C.-H., and N{\"u}hrenberg, G. (2018a).
\newblock nn-dependability-kit: Engineering neural networks for safety-critical
  systems.
\newblock {\em arXiv preprint arXiv:1811.06746}.

\bibitem[Cheng et~al., 2018b]{huang-atva18}
Cheng, C.-H., Huang, C.-H., and Yasuoka, H. (2018b).
\newblock Quantitative projection coverage for testing {ML}-enabled autonomous
  systems.
\newblock In {\em International Symposium on Automated Technology for
  Verification and Analysis}. Springer.

\bibitem[Cheng et~al., 2018c]{huang18b}
Cheng, C.-H., N{\"u}hrenberg, G., Huang, C.-H., and Yasuoka, H. (2018c).
\newblock Towards dependability metrics for neural networks.
\newblock In {\em Proceedings of the 16th ACM-IEEE International Conference on
  Formal Methods and Models for System Design}.

\bibitem[Cheng et~al., 2017]{CNR2017}
Cheng, C.-H., N{\"u}hrenberg, G., and Ruess, H. (2017).
\newblock Maximum resilience of artificial neural networks.
\newblock In D'Souza, D. and Narayan~Kumar, K., editors, {\em Automated
  Technology for Verification and Analysis}, pages 251--268. Springer.

\bibitem[{Cheng} et~al., 2018]{2018arXiv180906573C}
{Cheng}, C.-H., {N{\"u}hrenberg}, G., and {Yasuoka}, H. (2018).
\newblock {Runtime Monitoring Neuron Activation Patterns}.
\newblock {\em arXiv e-prints}, page arXiv:1809.06573.

\bibitem[Cheng et~al., 2018]{cheng2018manifesting}
Cheng, D., Cao, C., Xu, C., and Ma, X. (2018).
\newblock Manifesting bugs in machine learning code: An explorative study with
  mutation testing.
\newblock In {\em International Conference on Software Quality, Reliability and
  Security (QRS)}, pages 313--324. IEEE.

\bibitem[Chu et~al., 2018]{CHHWP2018}
Chu, L., Hu, X., Hu, J., Wang, L., and Pei, J. (2018).
\newblock Exact and consistent interpretation for piecewise linear neural
  networks: {A} closed form solution.
\newblock In {\em ACM SIGKDD International Conference on Knowledge Discovery \&
  Data Mining}, pages 1244--1253. ACM.

\bibitem[Clarke et~al., 2000]{CGJLV2000}
Clarke, E., Grumberg, O., Jha, S., Lu, Y., and Veith, H. (2000).
\newblock Counterexample-guided abstraction refinement.
\newblock In Emerson, E.~A. and Sistla, A.~P., editors, {\em Computer Aided
  Verification}, pages 154--169, Berlin, Heidelberg. Springer Berlin
  Heidelberg.

\bibitem[Clarke et~al., 2003]{CEGAR}
Clarke, E., Grumberg, O., Jha, S., Lu, Y., and Veith, H. (2003).
\newblock Counterexample-guided abstraction refinement for symbolic model
  checking.
\newblock {\em J. ACM}, 50(5):752--794.

\bibitem[Clarke~Jr et~al., 2018]{clarkebook}
Clarke~Jr, E.~M., Grumberg, O., Kroening, D., Peled, D., and Veith, H. (2018).
\newblock {\em Model checking}.
\newblock The MIT Press.

\bibitem[Collobert et~al., 2011]{CWBKKK2011}
Collobert, R., Weston, J., Bottou, L., Karlen, M., Kavukcuoglu, K., and Kuksa,
  P. (2011).
\newblock Natural language processing (almost) from scratch.
\newblock {\em J. Mach. Learn. Res.}, 12:2493--2537.

\bibitem[Cousot and Cousot, 1977]{CC1977}
Cousot, P. and Cousot, R. (1977).
\newblock Abstract interpretation: A unified lattice model for static analysis
  of programs by construction or approximation of fixpoints.
\newblock In {\em Fourth ACM Symposium on Principles of Programming Languages
  (POPL)}, pages 238--252.

\bibitem[Dabkowski and Gal, 2017]{DG2017}
Dabkowski, P. and Gal, Y. (2017).
\newblock Real time image saliency for black box classifiers.
\newblock In {\em NIPS}.

\bibitem[{DeVries} and {Taylor}, 2018]{2018arXiv180204865D}
{DeVries}, T. and {Taylor}, G.~W. (2018).
\newblock {Learning Confidence for Out-of-Distribution Detection in Neural
  Networks}.
\newblock {\em arXiv e-prints}, page arXiv:1802.04865.

\bibitem[Dhillon et~al., 2018]{dhillon2018stochastic}
Dhillon, G.~S., Azizzadenesheli, K., Lipton, Z.~C., Bernstein, J., Kossaifi,
  J., Khanna, A., and Anandkumar, A. (2018).
\newblock Stochastic activation pruning for robust adversarial defense.
\newblock In {\em International Conference on Learning Representations}.

\bibitem[Dosovitskiy and Brox, 2015]{DB2015}
Dosovitskiy, A. and Brox, T. (2015).
\newblock Inverting convolutional networks with convolutional networks.
\newblock {\em CoRR}, abs/1506.02753.

\bibitem[Dreossi et~al., 2017]{dreossi2017systematic}
Dreossi, T., Ghosh, S., Sangiovanni-Vincentelli, A., and Seshia, S.~A. (2017).
\newblock Systematic testing of convolutional neural networks for autonomous
  driving.
\newblock {\em arXiv preprint arXiv:1708.03309}.

\bibitem[Du et~al., 2019]{du2019deepstellar}
Du, X., Xie, X., Li, Y., Ma, L., Liu, Y., and Zhao, J. (2019).
\newblock Deepstellar: model-based quantitative analysis of stateful deep
  learning systems.
\newblock In {\em Proceedings of the 27th ACM Joint Meeting on European
  Software Engineering Conference and Symposium on the Foundations of Software
  Engineering}, pages 477--487.

\bibitem[Dutta et~al., 2018]{dutta2017output}
Dutta, S., Jha, S., Sanakaranarayanan, S., and Tiwari, A. (2018).
\newblock Output range analysis for deep neural networks.
\newblock {\em arXiv preprint arXiv:1709.09130}.

\bibitem[Dvijotham et~al., 2018]{DSGMK2018}
Dvijotham, K., Stanforth, R., Gowal, S., Mann, T.~A., and Kohli, P. (2018).
\newblock A dual approach to scalable verification of deep networks.
\newblock In {\em Conference on Uncertainty in Artificial Intelligence (UAI)},
  pages 550--559.

\bibitem[Ebrahimi et~al., 2018]{ebrahimi2018hotflip}
Ebrahimi, J., Rao, A., Lowd, D., and Dou, D. (2018).
\newblock Hotflip: White-box adversarial examples for text classification.
\newblock In {\em Proceedings of the 56th Annual Meeting of the Association for
  Computational Linguistics (Volume 2: Short Papers)}, volume~2, pages 31--36.

\bibitem[Ehlers, 2017]{Rudiger2017}
Ehlers, R. (2017).
\newblock Formal verification of piece-wise linear feed-forward neural
  networks.
\newblock In D'Souza, D. and Narayan~Kumar, K., editors, {\em Automated
  Technology for Verification and Analysis}, pages 269--286, Cham. Springer
  International Publishing.

\bibitem[Engstrom et~al., 2017]{engstrom2017rotation}
Engstrom, L., Tsipras, D., Schmidt, L., and Madry, A. (2017).
\newblock A rotation and a translation suffice: Fooling cnns with simple
  transformations.
\newblock {\em arXiv preprint arXiv:1712.02779}.

\bibitem[Erhan et~al., 2009]{EBCV2009}
Erhan, D., Bengio, Y., Courville, A., and Vincent, P. (2009).
\newblock {Visualizing Higher-Layer Features of a Deep Network}.
\newblock Technical report, University of Montreal.

\bibitem[{Feinman} et~al., 2017]{FCSG2017}
{Feinman}, R., {Curtin}, R.~R., {Shintre}, S., and {Gardner}, A.~B. (2017).
\newblock {Detecting Adversarial Samples from Artifacts}.
\newblock {\em arXiv e-prints}, page arXiv:1703.00410.

\bibitem[Finlayson et~al., 2018]{finlayson2018adversarial}
Finlayson, S.~G., Kohane, I.~S., and Beam, A.~L. (2018).
\newblock Adversarial attacks against medical deep learning systems.
\newblock {\em arXiv preprint arXiv:1804.05296}.

\bibitem[Fong and Vedaldi, 2017]{FV2017}
Fong, R. and Vedaldi, A. (2017).
\newblock Interpretable explanations of black boxes by meaningful perturbation.
\newblock {\em 2017 IEEE International Conference on Computer Vision (ICCV)},
  pages 3449--3457.

\bibitem[Gehr et~al., 2018]{GMDTCV2018}
Gehr, T., Mirman, M., Drachsler-Cohen, D., Tsankov, P., Chaudhuri, S., and
  Vechev, M. (2018).
\newblock {AI2}: Safety and robustness certification of neural networks with
  abstract interpretation.
\newblock In {\em 2018 IEEE Symposium on Security and Privacy (S\&P2018)},
  pages 948--963.

\bibitem[{Goldfeld} et~al., 2018]{GBGMNKP2018}
{Goldfeld}, Z., {van den Berg}, E., {Greenewald}, K., {Melnyk}, I., {Nguyen},
  N., {Kingsbury}, B., and {Polyanskiy}, Y. (2018).
\newblock {Estimating Information Flow in Neural Networks}.
\newblock {\em arXiv e-prints}, page arXiv:1810.05728.

\bibitem[Goodfellow et~al., 2016]{Goodfellow2016}
Goodfellow, I., Bengio, Y., and Courville, A. (2016).
\newblock {\em Deep Learning}.
\newblock MIT Press.

\bibitem[Goodfellow et~al., 2014a]{goodfellow2014generative}
Goodfellow, I., Pouget-Abadie, J., Mirza, M., Xu, B., Warde-Farley, D., Ozair,
  S., Courville, A., and Bengio, Y. (2014a).
\newblock Generative adversarial nets.
\newblock In {\em Advances in neural information processing systems}, pages
  2672--2680.

\bibitem[Goodfellow, 2018]{Goodfellow2018}
Goodfellow, I.~J. (2018).
\newblock Gradient masking causes {CLEVER} to overestimate adversarial
  perturbation size.
\newblock {\em CoRR}, abs/1804.07870.

\bibitem[Goodfellow et~al., 2014b]{FGSM}
Goodfellow, I.~J., Shlens, J., and Szegedy, C. (2014b).
\newblock Explaining and harnessing adversarial examples.
\newblock {\em CoRR}, abs/1412.6572.

\bibitem[Gopinath et~al., 2018]{gopinath2018symbolic}
Gopinath, D., Wang, K., Zhang, M., Pasareanu, C.~S., and Khurshid, S. (2018).
\newblock Symbolic execution for deep neural networks.
\newblock {\em arXiv preprint arXiv:1807.10439}.

\bibitem[Gowal et~al., 2019]{Gowal_2019_ICCV}
Gowal, S., Dvijotham, K.~D., Stanforth, R., Bunel, R., Qin, C., Uesato, J.,
  Arandjelovic, R., Mann, T., and Kohli, P. (2019).
\newblock Scalable verified training for provably robust image classification.
\newblock In {\em The IEEE International Conference on Computer Vision (ICCV)}.

\bibitem[Guo et~al., 2017]{guo2017countering}
Guo, C., Rana, M., Cisse, M., and van~der Maaten, L. (2017).
\newblock Countering adversarial images using input transformations.
\newblock {\em arXiv preprint arXiv:1711.00117}.

\bibitem[Guo et~al., 2018]{dlfuzz}
Guo, J., Jiang, Y., Zhao, Y., Chen, Q., and Sun, J. (2018).
\newblock {DLFuzz}: Differential fuzzing testing of deep learning systems.
\newblock In {\em Proceedings of the 2018 12nd Joint Meeting on Foundations of
  Software Engineering, {ESEC/FSE} 2018}.

\bibitem[Hardin, 2002]{Hardin:Trust:2002}
Hardin, R. (2002).
\newblock {\em Trust and trustworthiness}.
\newblock Russell Sage Foundation.

\bibitem[Hayes and Danezis, 2018]{hayes2018learning}
Hayes, J. and Danezis, G. (2018).
\newblock Learning universal adversarial perturbations with generative models.
\newblock In {\em 2018 IEEE Security and Privacy Workshops (SPW)}, pages
  43--49. IEEE.

\bibitem[Hayhurst et~al., 2001]{HVCR2001}
Hayhurst, K., Veerhusen, D., Chilenski, J., and Rierson, L. (2001).
\newblock A~practical tutorial on modified condition/decision coverage.
\newblock Technical report, NASA.

\bibitem[He et~al., 2016]{he2016deep}
He, K., Zhang, X., Ren, S., and Sun, J. (2016).
\newblock Deep residual learning for image recognition.
\newblock In {\em Proceedings of the IEEE conference on computer vision and
  pattern recognition}, pages 770--778.

\bibitem[Hein and Andriushchenko, 2017]{HA2017}
Hein, M. and Andriushchenko, M. (2017).
\newblock Formal guarantees on the robustness of a classifier against
  adversarial manipulation.
\newblock In {\em NIPS}.

\bibitem[Hendrik~Metzen et~al., 2017]{hendrik2017universal}
Hendrik~Metzen, J., Chaithanya~Kumar, M., Brox, T., and Fischer, V. (2017).
\newblock Universal adversarial perturbations against semantic image
  segmentation.
\newblock In {\em Proceedings of the IEEE International Conference on Computer
  Vision}, pages 2755--2764.

\bibitem[Hendrycks and Gimpel, 2016]{hendrycks2016early}
Hendrycks, D. and Gimpel, K. (2016).
\newblock Early methods for detecting adversarial images.
\newblock {\em arXiv preprint arXiv:1608.00530}.

\bibitem[Heusel et~al., 2017]{10.5555/3295222.3295408}
Heusel, M., Ramsauer, H., Unterthiner, T., Nessler, B., and Hochreiter, S.
  (2017).
\newblock Gans trained by a two time-scale update rule converge to a local nash
  equilibrium.
\newblock In {\em Proceedings of the 31st International Conference on Neural
  Information Processing Systems}, NIPS’17, page 6629–6640, Red Hook, NY,
  USA. Curran Associates Inc.

\bibitem[{Hinton} et~al., 2015]{HVD2015}
{Hinton}, G., {Vinyals}, O., and {Dean}, J. (2015).
\newblock {Distilling the Knowledge in a Neural Network}.
\newblock {\em arXiv e-prints}, page arXiv:1503.02531.

\bibitem[Houle, 2017]{Houle2017}
Houle, M.~E. (2017).
\newblock Local intrinsic dimensionality i: An extreme-value-theoretic
  foundation for similarity applications.
\newblock In Beecks, C., Borutta, F., Kr{\"o}ger, P., and Seidl, T., editors,
  {\em Similarity Search and Applications}, pages 64--79, Cham. Springer
  International Publishing.

\bibitem[Huang et~al., 2017a]{HPGDA2017}
Huang, S.~H., Papernot, N., Goodfellow, I.~J., Duan, Y., and Abbeel, P.
  (2017a).
\newblock Adversarial attacks on neural network policies.
\newblock {\em CoRR}, abs/1702.02284.

\bibitem[Huang et~al., 2019a]{huang2019testrnn}
Huang, W., Sun, Y., Huang, X., and Sharp, J. (2019a).
\newblock testrnn: Coverage-guided testing on recurrent neural networks.
\newblock {\em arXiv preprint arXiv:1906.08557}.

\bibitem[Huang et~al., 2019b]{huang2019test}
Huang, W., Sun, Y., Sharp, J., and Huang, X. (2019b).
\newblock Test metrics for recurrent neural networks.
\newblock {\em arXiv preprint arXiv:1911.01952}.

\bibitem[Huang et~al., 2020]{HZMSMH2020}
Huang, W., Zhou, Y., Meng, J., Sharp, J., Maskell, S., and Huang, X. (2020).
\newblock Formal verification of robustness and resilience of learning-enabled
  state estimation system for robotics.
\newblock {\em Technical Report}.

\bibitem[Huang and Kwiatkowska, 2017]{HK2017}
Huang, X. and Kwiatkowska, M. (2017).
\newblock Reasoning about cognitive trust in stochastic multiagent systems.
\newblock In {\em AAAI2017}, pages 3768--3774.

\bibitem[Huang et~al., 2017b]{HKWW2017}
Huang, X., Kwiatkowska, M., Wang, S., and Wu, M. (2017b).
\newblock Safety verification of deep neural networks.
\newblock In Majumdar, R. and Kun{\v{c}}ak, V., editors, {\em Computer Aided
  Verification}, pages 3--29, Cham. Springer International Publishing.

\bibitem[Jacob~Steinhardt, 2017]{SKL2017}
Jacob~Steinhardt, Pang Wei~Koh, P.~L. (2017).
\newblock Certified defenses for data poisoning attacks.
\newblock In {\em Advances in Neural Information Processing Systems 30}.

\bibitem[Jia and Harman, 2011]{JH2011}
Jia, Y. and Harman, M. (2011).
\newblock An analysis and survey of the development of mutation testing.
\newblock {\em IEEE Transactions on Software Engineering}, 37(5):649--678.

\bibitem[Johnson et~al., 2016]{johnson2016perceptual}
Johnson, J., Alahi, A., and Fei-Fei, L. (2016).
\newblock Perceptual losses for real-time style transfer and super-resolution.
\newblock In {\em European Conference on Computer Vision}, pages 694--711.
  Springer.

\bibitem[Katz et~al., 2017]{katz2017reluplex}
Katz, G., Barrett, C., Dill, D.~L., Julian, K., and Kochenderfer, M.~J. (2017).
\newblock Reluplex: An efficient {SMT} solver for verifying deep neural
  networks.
\newblock In {\em International Conference on Computer Aided Verification},
  pages 97--117. Springer.

\bibitem[Kim et~al., 2018a]{KWGCWVS2018}
Kim, B., Wattenberg, M., Gilmer, J., Cai, C., Wexler, J., Viegas, F., and
  sayres, R. (2018a).
\newblock Interpretability beyond feature attribution: Quantitative testing
  with concept activation vectors ({TCAV}).
\newblock In Dy, J. and Krause, A., editors, {\em Proceedings of the 35th
  International Conference on Machine Learning}, volume~80 of {\em Proceedings
  of Machine Learning Research}, pages 2668--2677, Stockholmsmässan, Stockholm
  Sweden. PMLR.

\bibitem[Kim et~al., 2018b]{kim2018guiding}
Kim, J., Feldt, R., and Yoo, S. (2018b).
\newblock Guiding deep learning system testing using surprise adequacy.
\newblock {\em arXiv preprint arXiv:1808.08444}.

\bibitem[Kingma and Ba, 2014]{kingma2014adam}
Kingma, D.~P. and Ba, J. (2014).
\newblock Adam: A method for stochastic optimization.
\newblock {\em arXiv preprint arXiv:1412.6980}.

\bibitem[{Koh} and {Liang}, 2017]{KL2017}
{Koh}, P.~W. and {Liang}, P. (2017).
\newblock {Understanding Black-box Predictions via Influence Functions}.
\newblock In {\em International Conference on Machine Learning}.

\bibitem[K{\"o}nighofer and Bloem, 2013]{KB2013}
K{\"o}nighofer, R. and Bloem, R. (2013).
\newblock Repair with on-the-fly program analysis.
\newblock In Biere, A., Nahir, A., and Vos, T., editors, {\em Hardware and
  Software: Verification and Testing}, pages 56--71, Berlin, Heidelberg.
  Springer Berlin Heidelberg.

\bibitem[Kurakin et~al., 2016]{KGB2016}
Kurakin, A., Goodfellow, I.~J., and Bengio, S. (2016).
\newblock Adversarial examples in the physical world.
\newblock {\em CoRR}, abs/1607.02533.

\bibitem[Li et~al., 2019]{li2019universal}
Li, J., Ji, R., Liu, H., Hong, X., Gao, Y., and Tian, Q. (2019).
\newblock Universal perturbation attack against image retrieval.
\newblock In {\em Proceedings of the IEEE International Conference on Computer
  Vision}, pages 4899--4908.

\bibitem[Li et~al., 2018]{LLYCH2018}
Li, J., Liu, J., Yang, P., Chen, L., and Huang, X. (2018).
\newblock Analyzing deep neural networks with symbolic propagation: Towards
  higher precision and faster verification.
\newblock {\em Submitted}.

\bibitem[Liang et~al., 2018]{liang2018enhancing}
Liang, S., Li, Y., and Srikant, R. (2018).
\newblock Enhancing the reliability of out-of-distribution image detection in
  neural networks.
\newblock In {\em International Conference on Learning Representations}.

\bibitem[Lin et~al., 2017]{LHLSLS2017}
Lin, Y., Hong, Z., Liao, Y., Shih, M., Liu, M., and Sun, M. (2017).
\newblock Tactics of adversarial attack on deep reinforcement learning agents.
\newblock In {\em International Joint Conference on Artificial Intelligence}.

\bibitem[Lipton, 2018]{Lipton2016}
Lipton, Z.~C. (2018).
\newblock The mythos of model interpretability.
\newblock {\em Commun. ACM}, 61:36--43.

\bibitem[Lomuscio and Maganti, 2017]{LM2017}
Lomuscio, A. and Maganti, L. (2017).
\newblock An approach to reachability analysis for feed-forward {ReLU} neural
  networks.
\newblock {\em arXiv preprint arXiv:1706.07351}.

\bibitem[{Lu} et~al., 2017]{NOneedtoworry}
{Lu}, J., {Sibai}, H., {Fabry}, E., and {Forsyth}, D. (2017).
\newblock {NO Need to Worry about Adversarial Examples in Object Detection in
  Autonomous Vehicles}.
\newblock In {\em Conference on Computer Vision and Pattern Recognition,
  Spotlight Oral Workshop}.

\bibitem[Lundberg and Lee, 2017]{LL2017}
Lundberg, S. and Lee, S. (2017).
\newblock A unified approach to interpreting model predictions.
\newblock {\em NIPS}.

\bibitem[Ma et~al., 2018a]{ma2018deepgauge}
Ma, L., Juefei{-}Xu, F., Sun, J., Chen, C., Su, T., Zhang, F., Xue, M., Li, B.,
  Li, L., Liu, Y., Zhao, J., and Wang, Y. (2018a).
\newblock {DeepGauge}: Comprehensive and multi-granularity testing criteria for
  gauging the robustness of deep learning systems.
\newblock In {\em Automated Software Engineering (ASE), 33rd IEEE/ACM
  International Conference on}.

\bibitem[Ma et~al., 2018b]{deepmutation}
Ma, L., Zhang, F., Sun, J., Xue, M., Li, B., Juefei-Xu, F., Xie, C., Li, L.,
  Liu, Y., Zhao, J., et~al. (2018b).
\newblock {DeepMutation}: Mutation testing of deep learning systems.
\newblock In {\em Software Reliability Engineering, IEEE 29th International
  Symposium on}.

\bibitem[Ma et~al., 2018c]{ma2018combinatorial}
Ma, L., Zhang, F., Xue, M., Li, B., Liu, Y., Zhao, J., and Wang, Y. (2018c).
\newblock Combinatorial testing for deep learning systems.
\newblock {\em arXiv preprint arXiv:1806.07723}.

\bibitem[Ma et~al., 2018d]{mode}
Ma, S., Liu, Y., Zhang, X., Lee, W.-C., and Grama, A. (2018d).
\newblock {MODE}: Automated neural network model debugging via state
  differential analysis and input selection.
\newblock In {\em Proceedings of the 12nd Joint Meeting on Foundations of
  Software Engineering}. ACM.

\bibitem[Ma et~al., 2018e]{ma2018characterizing}
Ma, X., Li, B., Wang, Y., Erfani, S.~M., Wijewickrema, S., Houle, M.~E.,
  Schoenebeck, G., Song, D., and Bailey, J. (2018e).
\newblock Characterizing adversarial subspaces using local intrinsic
  dimensionality.
\newblock {\em arXiv preprint arXiv:1801.02613}.

\bibitem[Madry et~al., 2017]{madry2017towards}
Madry, A., Makelov, A., Schmidt, L., Tsipras, D., and Vladu, A. (2017).
\newblock Towards deep learning models resistant to adversarial attacks.
\newblock {\em arXiv preprint arXiv:1706.06083}.

\bibitem[Mahendran and Vedaldi, 2015]{MV2014}
Mahendran, A. and Vedaldi, A. (2015).
\newblock Understanding deep image representations by inverting them.
\newblock {\em 2015 IEEE Conference on Computer Vision and Pattern Recognition
  (CVPR)}, pages 5188--5196.

\bibitem[Mayer et~al., 1995]{Mayer:AMR:1995}
Mayer, R.~C., Davis, J.~H., and Schoorman, F.~D. (1995).
\newblock An integrative model of organizational trust.
\newblock {\em Academy of management review}, 20(3):709--734.

\bibitem[Meng and Chen, 2017a]{meng2017magnet}
Meng, D. and Chen, H. (2017a).
\newblock Magnet: a two-pronged defense against adversarial examples.
\newblock In {\em Proceedings of the 2017 ACM SIGSAC Conference on Computer and
  Communications Security}, pages 135--147. ACM.

\bibitem[Meng and Chen, 2017b]{MC2017}
Meng, D. and Chen, H. (2017b).
\newblock Magnet: a two-pronged defense against adversarial examples.
\newblock In {\em ACM Conference on Computer and Communications Security}.

\bibitem[Metzen et~al., 2017]{metzen2017detecting}
Metzen, J.~H., Genewein, T., Fischer, V., and Bischoff, B. (2017).
\newblock On detecting adversarial perturbations.
\newblock {\em arXiv preprint arXiv:1702.04267}.

\bibitem[Mirman et~al., 2018]{mirman2018differentiable}
Mirman, M., Gehr, T., and Vechev, M. (2018).
\newblock Differentiable abstract interpretation for provably robust neural
  networks.
\newblock In {\em International Conference on Machine Learning}, pages
  3575--3583.

\bibitem[Mnih et~al., 2015]{MKSRVBG2015}
Mnih, V., Kavukcuoglu, K., Silver, D., Rusu, A.~A., Veness, J., Bellemare,
  M.~G., Graves, A., Riedmiller, M., Fidjeland, A.~K., Ostrovski, G., Petersen,
  S., Beattie, C., Sadik, A., Antonoglou, I., King, H., Kumaran, D., Wierstra,
  D., Legg, S., and Hassabis, D. (2015).
\newblock Human-level control through deep reinforcement learning.
\newblock {\em Nature}, 518:529 EP --.

\bibitem[Moosavi-Dezfooli et~al., 2017]{moosavi2017universal}
Moosavi-Dezfooli, S.-M., Fawzi, A., Fawzi, O., and Frossard, P. (2017).
\newblock Universal adversarial perturbations.
\newblock {\em 2017 IEEE Conference on Computer Vision and Pattern Recognition
  (CVPR)}, pages 86--94.

\bibitem[Moosavi-Dezfooli et~al., 2016]{moosavi2016deepfool}
Moosavi-Dezfooli, S.-M., Fawzi, A., and Frossard, P. (2016).
\newblock Deepfool: a simple and accurate method to fool deep neural networks.
\newblock In {\em Proceedings of the IEEE Conference on Computer Vision and
  Pattern Recognition}, pages 2574--2582.

\bibitem[Mopuri et~al., 2018]{mopuri2018generalizable}
Mopuri, K.~R., Ganeshan, A., and Babu, R.~V. (2018).
\newblock Generalizable data-free objective for crafting universal adversarial
  perturbations.
\newblock {\em IEEE transactions on pattern analysis and machine intelligence},
  41(10):2452--2465.

\bibitem[Moreno-Torres et~al., 2012]{Moreno-Torres2012}
Moreno-Torres, J.~G., Raeder, T., Alaiz-Rodr{\'i}guez, R., Chawla, N.~V., and
  Herrera, F. (2012).
\newblock A unifying view on dataset shift in classification.
\newblock {\em Pattern Recogn.}, 45(1):521--530.

\bibitem[Na et~al., 2018]{na2017cascade}
Na, T., Ko, J.~H., and Mukhopadhyay, S. (2018).
\newblock Cascade adversarial machine learning regularized with a unified
  embedding.
\newblock In {\em International Conference on Learning Representations (ICLR)}.

\bibitem[Nair and Hinton, 2010]{relu}
Nair, V. and Hinton, G.~E. (2010).
\newblock Rectified linear units improve restricted {Boltzmann} machines.
\newblock In {\em Proceedings of the 27th International Conference on Machine
  Learning (ICML)}, pages 807--814.

\bibitem[Narodytska, 2018]{narodytska2018formal}
Narodytska, N. (2018).
\newblock Formal analysis of deep binarized neural networks.
\newblock In {\em Proceedings of the 27th International Joint Conference on
  Artificial Intelligence}, pages 5692--5696. AAAI Press.

\bibitem[Narodytska et~al., 2018]{NKPSW2017}
Narodytska, N., Kasiviswanathan, S.~P., Ryzhyk, L., Sagiv, M., and Walsh, T.
  (2018).
\newblock Verifying properties of binarized deep neural networks.
\newblock In {\em Proceedings of the Thirty-Second {AAAI} Conference on
  Artificial Intelligence}, pages 6615--6624.

\bibitem[Neekhara et~al., 2019]{neekhara2019universal}
Neekhara, P., Hussain, S., Pandey, P., Dubnov, S., McAuley, J., and Koushanfar,
  F. (2019).
\newblock Universal adversarial perturbations for speech recognition systems.
\newblock {\em Proc. Interspeech 2019}, pages 481--485.

\bibitem[Neumaier and Shcherbina, 2004]{NS04}
Neumaier, A. and Shcherbina, O. (2004).
\newblock Safe bounds in linear and mixed-integer linear programming.
\newblock {\em Math. Program.}, 99(2):283--296.

\bibitem[Noller et~al., 2020]{nollerhydiff}
Noller, Y., P{\u{a}}s{\u{a}}reanu, C.~S., B{\"o}hme, M., Sun, Y., Nguyen,
  H.~L., and Grunske, L. (2020).
\newblock {HyDiff}: Hybrid differential software analysis.
\newblock In {\em Proceedings of the 42nd International Conference on Software
  Engineering, {ICSE}}.

\bibitem[Odena and Goodfellow, 2018]{odena2018tensorfuzz}
Odena, A. and Goodfellow, I. (2018).
\newblock {TensorFuzz}: Debugging neural networks with coverage-guided fuzzing.
\newblock {\em arXiv preprint arXiv:1807.10875}.

\bibitem[Olah et~al., 2017]{olah2017feature}
Olah, C., Mordvintsev, A., and Schubert, L. (2017).
\newblock Feature visualization.
\newblock {\em Distill}.
\newblock https://distill.pub/2017/feature-visualization.

\bibitem[Omlin and Giles, 1996]{OG1996}
Omlin, C.~W. and Giles, C.~L. (1996).
\newblock Extraction of rules from discrete-time recurrent neural networks.
\newblock {\em Neural Networks}, 9(1):41--52.

\bibitem[OSearcoid, 2006]{OSearcoid2006}
OSearcoid, M. (2006).
\newblock {\em Metric Spaces}.
\newblock Springer Science \& Business Media.

\bibitem[Papernot et~al., 2018]{papernot2018cleverhans}
Papernot, N., Faghri, F., Carlini, N., Goodfellow, I., Feinman, R., Kurakin,
  A., Xie, C., Sharma, Y., Brown, T., Roy, A., Matyasko, A., Behzadan, V.,
  Hambardzumyan, K., Zhang, Z., Juang, Y.-L., Li, Z., Sheatsley, R., Garg, A.,
  Uesato, J., Gierke, W., Dong, Y., Berthelot, D., Hendricks, P., Rauber, J.,
  and Long, R. (2018).
\newblock Technical report on the cleverhans v2.1.0 adversarial examples
  library.
\newblock {\em arXiv preprint arXiv:1610.00768}.

\bibitem[Papernot et~al., 2016a]{papernot2016distillation}
Papernot, N., McDaniel, P., Wu, X., Jha, S., and Swami, A. (2016a).
\newblock Distillation as a defense to adversarial perturbations against deep
  neural networks.
\newblock In {\em Security and Privacy (SP), 2016 IEEE Symposium on}, pages
  582--597. IEEE.

\bibitem[Papernot et~al., 2016b]{PMWJS2016}
Papernot, N., McDaniel, P., Wu, X., Jha, S., and Swami, A. (2016b).
\newblock Distillation as a defense to adversarial perturbations against deep
  neural networks.
\newblock In {\em 2016 IEEE Symposium on Security and Privacy (SP)}, pages
  582--597.

\bibitem[Papernot et~al., 2016c]{JSMA}
Papernot, N., McDaniel, P.~D., Jha, S., Fredrikson, M., Celik, Z.~B., and
  Swami, A. (2016c).
\newblock The limitations of deep learning in adversarial settings.
\newblock In {\em Security and Privacy (EuroS\&P), 2016 IEEE European Symposium
  on}, pages 372--387. IEEE.

\bibitem[Pattanaik et~al., 2018]{PTLBC2017}
Pattanaik, A., Tang, Z., Liu, S., Bommannan, G., and Chowdhary, G. (2018).
\newblock Robust deep reinforcement learning with adversarial attacks.
\newblock In {\em International Conference on Autonomous Agents and Multiagent
  Systems (AAMAS)}.

\bibitem[Peck et~al., 2017]{PRGS2017}
Peck, J., Roels, J., Goossens, B., and Saeys, Y. (2017).
\newblock Lower bounds on the robustness to adversarial perturbations.
\newblock In Guyon, I., Luxburg, U.~V., Bengio, S., Wallach, H., Fergus, R.,
  Vishwanathan, S., and Garnett, R., editors, {\em Advances in Neural
  Information Processing Systems 30}, pages 804--813. Curran Associates, Inc.

\bibitem[Pei et~al., 2017a]{PCYJ2017}
Pei, K., Cao, Y., Yang, J., and Jana, S. (2017a).
\newblock {DeepXplore}: Automated whitebox testing of deep learning systems.
\newblock In {\em Proceedings of the 26th Symposium on Operating Systems
  Principles}, pages 1--18. ACM.

\bibitem[Pei et~al., 2017b]{pei2017towards}
Pei, K., Cao, Y., Yang, J., and Jana, S. (2017b).
\newblock Towards practical verification of machine learning: The case of
  computer vision systems.
\newblock {\em arXiv preprint arXiv:1712.01785}.

\bibitem[Poursaeed et~al., 2018]{poursaeed2017generative}
Poursaeed, O., Katsman, I., Gao, B., and Belongie, S.~J. (2018).
\newblock Generative adversarial perturbations.
\newblock In {\em Conference on Computer Vision and Pattern Recognition}.

\bibitem[Pulina and Tacchella, 2010]{PT2010}
Pulina, L. and Tacchella, A. (2010).
\newblock An abstraction-refinement approach to verification of artificial
  neural networks.
\newblock In {\em International Conference on Computer Aided Verification},
  pages 243--257. Springer.

\bibitem[Raghunathan et~al., 2018]{raghunathan2018certified}
Raghunathan, A., Steinhardt, J., and Liang, P. (2018).
\newblock Certified defenses against adversarial examples.
\newblock {\em arXiv preprint arXiv:1801.09344}.

\bibitem[Ribeiro et~al., 2016]{LIME}
Ribeiro, M.~T., Singh, S., and Guestrin, C. (2016).
\newblock "why should {I} trust you?": Explaining the predictions of any
  classifier.
\newblock In {\em HLT-NAACL Demos}.

\bibitem[Ribeiro et~al., 2018]{Anchors}
Ribeiro, M.~T., Singh, S., and Guestrin, C. (2018).
\newblock Anchors: High-precision model-agnostic explanations.
\newblock In {\em Proceedings of the Thirty-Second {AAAI} Conference on
  Artificial Intelligence}.

\bibitem[Ronneberger et~al., 2015]{ronneberger2015u}
Ronneberger, O., Fischer, P., and Brox, T. (2015).
\newblock U-net: Convolutional networks for biomedical image segmentation.
\newblock In {\em International Conference on Medical image computing and
  computer-assisted intervention}, pages 234--241. Springer.

\bibitem[RTCA, 2011]{do178}
RTCA (2011).
\newblock Do-178c, software considerations in airborne systems and equipment
  certification.

\bibitem[Ruan et~al., 2018a]{RHK2018}
Ruan, W., Huang, X., and Kwiatkowska, M. (2018a).
\newblock Reachability analysis of deep neural networks with provable
  guarantees.
\newblock In {\em Proceedings of the 27th International Joint Conference on
  Artificial Intelligence}, pages 2651--2659. AAAI Press.

\bibitem[Ruan et~al., 2018b]{ruan2018reachability}
Ruan, W., Huang, X., and Kwiatkowska, M. (2018b).
\newblock Reachability analysis of deep neural networks with provable
  guarantees.
\newblock In {\em Proceedings of the Twenty-Seventh International Joint
  Conference on Artificial Intelligence, {IJCAI-18}}, pages 2651--2659.
  International Joint Conferences on Artificial Intelligence Organization.

\bibitem[Ruan et~al., 2019]{ruan2019global}
Ruan, W., Wu, M., Sun, Y., Huang, X., Kroening, D., and Kwiatkowska, M. (2019).
\newblock Global robustness evaluation of deep neural networks with provable
  guarantees for the {H}amming distance.
\newblock In {\em Proceedings of the Twenty-Eighth International Joint
  Conference on Artificial Intelligence, {IJCAI-19}}, pages 5944--5952.
  International Joint Conferences on Artificial Intelligence Organization.

\bibitem[Rushby, 2015]{Rushby2015}
Rushby, J. (2015).
\newblock The interpretation and evaluation of assurance cases.
\newblock Technical report, SRI International.

\bibitem[Russakovsky et~al., 2015]{ILSVRC15}
Russakovsky, O., Deng, J., Su, H., Krause, J., Satheesh, S., Ma, S., Huang, Z.,
  Karpathy, A., Khosla, A., Bernstein, M., Berg, A.~C., and Fei-Fei, L. (2015).
\newblock {ImageNet Large Scale Visual Recognition Challenge}.
\newblock {\em International Journal of Computer Vision (IJCV)},
  115(3):211--252.

\bibitem[Salay and Czarnecki, 2018]{salay2018using}
Salay, R. and Czarnecki, K. (2018).
\newblock Using machine learning safely in automotive software: An assessment
  and adaption of software process requirements in iso 26262.
\newblock {\em arXiv preprint arXiv:1808.01614}.

\bibitem[Samangouei et~al., 2018]{samangouei2018defense}
Samangouei, P., Kabkab, M., and Chellappa, R. (2018).
\newblock Defense-gan: Protecting classifiers against adversarial attacks using
  generative models.
\newblock In {\em International Conference on Learning Representations (ICLR)}.

\bibitem[Sato and Tsukimoto, 2001]{CRED}
Sato, M. and Tsukimoto, H. (2001).
\newblock Rule extraction from neural networks via decision tree induction.
\newblock In {\em IJCNN'01. International Joint Conference on Neural Networks.
  Proceedings (Cat. No.01CH37222)}, volume~3, pages 1870--1875 vol.3.

\bibitem[Saxe et~al., 2018]{Saxe2018}
Saxe, A.~M., Bansal, Y., Dapello, J., Advani, M., Kolchinsky, A., Tracey,
  B.~D., and Cox, D.~D. (2018).
\newblock On the information bottleneck theory of deep learning.
\newblock In {\em International Conference on Learning Representations}.

\bibitem[Schaul et~al., 2015]{SQAS2015}
Schaul, T., Quan, J., Antonoglou, I., and Silver, D. (2015).
\newblock Prioritized experience replay.
\newblock {\em CoRR}, abs/1511.05952.

\bibitem[Selvaraju et~al., 2016]{CAM}
Selvaraju, R.~R., Das, A., Vedantam, R., Cogswell, M., Parikh, D., and Batra,
  D. (2016).
\newblock Grad-cam: Why did you say that? visual explanations from deep
  networks via gradient-based localization.
\newblock In {\em NIPS 2016 Workshop on Interpretable Machine Learning in
  Complex Systems}.

\bibitem[Shen et~al., 2018]{shenmunn}
Shen, W., Wan, J., and Chen, Z. (2018).
\newblock {MuNN}: Mutation analysis of neural networks.
\newblock In {\em International Conference on Software Quality, Reliability and
  Security Companion, {QRS-C}}. IEEE.

\bibitem[Shrikumar et~al., 2017]{SGK2017}
Shrikumar, A., Greenside, P., and Kundaje, A. (2017).
\newblock Learning important features through propagating activation
  differences.
\newblock In {\em Proceedings of Machine Learning Research 70:3145-3153}.

\bibitem[Shwartz{-}Ziv and Tishby, 2017]{openbox-IB}
Shwartz{-}Ziv, R. and Tishby, N. (2017).
\newblock Opening the black box of deep neural networks via information.
\newblock {\em CoRR}, abs/1703.00810.

\bibitem[Sifakis, 2018]{Sifakis2018}
Sifakis, J. (2018).
\newblock Autonomous systems - an architectural characterization.
\newblock {\em CoRR}, abs/1811.10277.

\bibitem[Silver et~al., 2017]{alphaGoZero}
Silver, D., Schrittwieser, J., Simonyan, K., Antonoglou, I., Huang, A., Guez,
  A., Hubert, T., Baker, L., Lai, M., Bolton, A., Chen, Y., Lillicrap, T., Hui,
  F., Sifre, L., van~den Driessche, G., Graepel, T., and Hassabis, D. (2017).
\newblock Mastering the game of {Go} without human knowledge.
\newblock {\em Nature}, 550(354--359).

\bibitem[Simonyan et~al., 2013]{SVZ2013}
Simonyan, K., Vedaldi, A., and Zisserman, A. (2013).
\newblock Deep inside convolutional networks: Visualising image classification
  models and saliency maps.
\newblock {\em CoRR}, abs/1312.6034.

\bibitem[Sinha et~al., 2018]{sinha2018certifiable}
Sinha, A., Namkoong, H., and Duchi, J. (2018).
\newblock Certifiable distributional robustness with principled adversarial
  training.
\newblock In {\em International Conference on Learning Representations}.

\bibitem[Smilkov et~al., 2017]{STKVW2017}
Smilkov, D., Thorat, N., Kim, B., Vi{\'{e}}gas, F.~B., and Wattenberg, M.
  (2017).
\newblock Smoothgrad: removing noise by adding noise.
\newblock {\em CoRR}, abs/1706.03825.

\bibitem[Song et~al., 2018]{song2017pixeldefend}
Song, Y., Kim, T., Nowozin, S., Ermon, S., and Kushman, N. (2018).
\newblock Pixeldefend: Leveraging generative models to understand and defend
  against adversarial examples.
\newblock In {\em International Conference on Learning Representations}.

\bibitem[Springenberg et~al., 2014]{SDB2014}
Springenberg, J.~T., Dosovitskiy, A., Brox, T., and Riedmiller, M.~A. (2014).
\newblock Striving for simplicity: The all convolutional net.
\newblock {\em CoRR}, abs/1412.6806.

\bibitem[Su et~al., 2017]{SWMPHS2017}
Su, T., Wu, K., Miao, W., Pu, G., He, J., Chen, Y., and Su, Z. (2017).
\newblock A~survey on data-flow testing.
\newblock {\em ACM Computing Surveys}, 50(1):5:1--5:35.

\bibitem[Sun et~al., 2018a]{sun2018testing}
Sun, Y., Huang, X., and Kroening, D. (2018a).
\newblock Testing deep neural networks.
\newblock {\em arXiv preprint arXiv:1803.04792}.

\bibitem[Sun et~al., 2019a]{sun2018concolicb}
Sun, Y., Huang, X., Kroening, D., Sharp, J., Hill, M., and Ashmore, R. (2019a).
\newblock Deepconcolic: testing and debugging deep neural networks.
\newblock In {\em 41st International Conference on Software Engineering:
  Companion Proceedings (ICSE-Companion)}, pages 111--114. IEEE.

\bibitem[Sun et~al., 2019b]{sun2018testing-b}
Sun, Y., Huang, X., Kroening, D., Sharp, J., Hill, M., and Ashmore, R. (2019b).
\newblock Structural test coverage criteria for deep neural networks.
\newblock In {\em 41st International Conference on Software Engineering:
  Companion Proceedings (ICSE-Companion)}, pages 320--321. IEEE.

\bibitem[Sun et~al., 2019c]{sun2018testingjournal}
Sun, Y., Huang, X., Kroening, D., Sharp, J., Hill, M., and Ashmore, R. (2019c).
\newblock Structural test coverage criteria for deep neural networks.
\newblock {\em ACM Trans. Embed. Comput. Syst.}, 18(5s).

\bibitem[Sun et~al., 2018b]{sun2018concolic}
Sun, Y., Wu, M., Ruan, W., Huang, X., Kwiatkowska, M., and Kroening, D.
  (2018b).
\newblock Concolic testing for deep neural networks.
\newblock In {\em Automated Software Engineering (ASE), 33rd IEEE/ACM
  International Conference on}.

\bibitem[Sun et~al., 2020]{SZMSH2020}
Sun, Y., Zhou, Y., Maskell, S., Sharp, J., and Huang, X. (2020).
\newblock Reliability validation of learning enabled vehicle tracking.
\newblock In {\em ICRA}.

\bibitem[Sundararajan et~al., 2017]{STY2017}
Sundararajan, M., Taly, A., and Yan, Q. (2017).
\newblock Axiomatic attribution for deep networks.
\newblock In {\em International Conference on Machine Learning}.

\bibitem[Szegedy et~al., 2014]{szegedy2014intriguing}
Szegedy, C., Zaremba, W., Sutskever, I., Bruna, J., Erhan, D., Goodfellow, I.,
  and Fergus, R. (2014).
\newblock Intriguing properties of neural networks.
\newblock In {\em In ICLR}. Citeseer.

\bibitem[Tamagnini et~al., 2017]{TKDB2017}
Tamagnini, P., Krause, J., Dasgupta, A., and Bertini, E. (2017).
\newblock Interpreting black-box classifiers using instance-level visual
  explanations.
\newblock In {\em Proceedings of the 2Nd Workshop on Human-In-the-Loop Data
  Analytics}, HILDA'17, pages 6:1--6:6, New York, NY, USA. ACM.

\bibitem[Tian et~al., 2018]{tian2017deeptest}
Tian, Y., Pei, K., Jana, S., and Ray, B. (2018).
\newblock {DeepTest}: Automated testing of deep-neural-network-driven
  autonomous cars.
\newblock {\em 2018 IEEE/ACM 40th International Conference on Software
  Engineering (ICSE)}, pages 303--314.

\bibitem[{Tishby} et~al., 2000]{IB}
{Tishby}, N., {Pereira}, F.~C., and {Bialek}, W. (2000).
\newblock {The information bottleneck method}.
\newblock {\em arXiv e-prints}, page physics/0004057.

\bibitem[{Tram{\`e}r} et~al., 2018]{TKPGBM2017}
{Tram{\`e}r}, F., {Kurakin}, A., {Papernot}, N., {Goodfellow}, I., {Boneh}, D.,
  and {McDaniel}, P. (2018).
\newblock {Ensemble Adversarial Training: Attacks and Defenses}.
\newblock In {\em International Conference on Learning Representations}.

\bibitem[Tretschk et~al., 2018]{TJF2018}
Tretschk, E., Oh, S.~J., and Fritz, M. (2018).
\newblock Sequential attacks on agents for long-term adversarial goals.
\newblock {\em CoRR}, abs/1805.12487.

\bibitem[Udeshi et~al., 2018]{udeshi2018automated}
Udeshi, S., Arora, P., and Chattopadhyay, S. (2018).
\newblock Automated directed fairness testing.
\newblock In {\em Automated Software Engineering (ASE), 33rd IEEE/ACM
  International Conference on}.

\bibitem[van~den Oord et~al., 2016]{vdOKVEGK2016}
van~den Oord, A., Kalchbrenner, N., Vinyals, O., Espeholt, L., Graves, A., and
  Kavukcuoglu, K. (2016).
\newblock Conditional image generation with pixelcnn decoders.
\newblock In {\em NIPS}.

\bibitem[van Hasselt et~al., 2015]{vHGS2015}
van Hasselt, H., Guez, A., and Silver, D. (2015).
\newblock Deep reinforcement learning with double q-learning.
\newblock In {\em Association for the Advancement of Artificial Intelligence}.

\bibitem[Voosen, 2017]{interpretability}
Voosen, P. (2017).
\newblock How {AI} detectives are cracking open the black box of deep learning.
\newblock {\em Science}.

\bibitem[Wand and Jones, 1994]{WJ1994}
Wand, M.~P. and Jones, M.~C. (1994).
\newblock {\em Kernel smoothing}.
\newblock Chapman and Hall/CRC.

\bibitem[Wang et~al., 2018]{reluval}
Wang, S., Pei, K., Whitehouse, J., Yang, J., and Jana, S. (2018).
\newblock Formal security analysis of neural networks using symbolic intervals.
\newblock In {\em USENIX Security Symposium}. USENIX Association.

\bibitem[Wang et~al., 2016]{WdFL2015}
Wang, Z., de~Freitas, N., and Lanctot, M. (2016).
\newblock Dueling network architectures for deep reinforcement learning.
\newblock In {\em International Conference on Machine Learning}.

\bibitem[Wang et~al., 2003]{WSB2003}
Wang, Z., Simoncelli, E.~P., and Bovik, A.~C. (2003).
\newblock Multiscale structural similarity for image quality assessment.
\newblock In {\em Signals, Systems and Computers, Conference Record of the
  Thirty-Seventh Asilomar Conference on}.

\bibitem[Wei et~al., 2018]{WZS2018}
Wei, X., Zhu, J., and Su, H. (2018).
\newblock Sparse adversarial perturbations for videos.
\newblock {\em CoRR}, abs/1803.02536.

\bibitem[Weiss et~al., 2018]{WGY2017}
Weiss, G., Goldberg, Y., and Yahav, E. (2018).
\newblock Extracting automata from recurrent neural networks using queries and
  counterexamples.
\newblock In {\em International Conference on Machine Learning}.

\bibitem[Weng et~al., 2018]{WZCSHBDD2018}
Weng, T.-W., Zhang, H., Chen, H., Song, Z., Hsieh, C.-J., Boning, D., Dhillon,
  I.~S., and Daniel, L. (2018).
\newblock Towards fast computation of certified robustness for relu networks.
\newblock In {\em Proceedings of the 35th International Conference on Machine
  Learning (ICML)}, pages 5273--5282.

\bibitem[{Weng} et~al., 2018]{WZCYSGHD2018}
{Weng}, T.-W., {Zhang}, H., {Chen}, P.-Y., {Yi}, J., {Su}, D., {Gao}, Y.,
  {Hsieh}, C.-J., and {Daniel}, L. (2018).
\newblock {Evaluating the Robustness of Neural Networks: An Extreme Value
  Theory Approach}.
\newblock In {\em International Conference on Learning Representations (ICLR)}.

\bibitem[Wicker et~al., 2018]{wicker2018feature}
Wicker, M., Huang, X., and Kwiatkowska, M. (2018).
\newblock Feature-guided black-box safety testing of deep neural networks.
\newblock In {\em International Conference on Tools and Algorithms for the
  Construction and Analysis of Systems}, pages 408--426. Springer.

\bibitem[Wolchover, 2017]{quantamagazine}
Wolchover, N. (2017).
\newblock New theory cracks open the black box of deep learning.
  {https://www.quantamagazine.org/new-theory-cracks-open-the-black-box-of-deep-learning-20170921/}.

\bibitem[Wong and Kolter, 2018]{wong2018provable}
Wong, E. and Kolter, Z. (2018).
\newblock Provable defenses against adversarial examples via the convex outer
  adversarial polytope.
\newblock In {\em International Conference on Machine Learning}, pages
  5283--5292.

\bibitem[Wu et~al., 2020]{wu2019game}
Wu, M., Wicker, M., Ruan, W., Huang, X., and Kwiatkowska, M. (2020).
\newblock A game-based approximate verification of deep neural networks with
  provable guarantees.
\newblock {\em Theoretical Computer Science}, 807:298 -- 329.
\newblock In memory of Maurice Nivat, a founding father of Theoretical Computer
  Science - Part II.

\bibitem[Xiang et~al., 2018]{xiang2017output}
Xiang, W., Tran, H.-D., and Johnson, T.~T. (2018).
\newblock Output reachable set estimation and verification for multi-layer
  neural networks.
\newblock {\em IEEE Transactions on Neural Networks and Learning Systems},
  29:5777--5783.

\bibitem[Xiao et~al., 2018]{xiao2018spatially}
Xiao, C., Zhu, J.-Y., Li, B., He, W., Liu, M., and Song, D. (2018).
\newblock Spatially transformed adversarial examples.
\newblock {\em arXiv preprint arXiv:1801.02612}.

\bibitem[Xie et~al., 2017]{xie2017mitigating}
Xie, C., Wang, J., Zhang, Z., Ren, Z., and Yuille, A. (2017).
\newblock Mitigating adversarial effects through randomization.
\newblock In {\em International Conference on Learning Representations (ICLR)}.

\bibitem[Xie et~al., 2018]{xie2018coverage}
Xie, X., Ma, L., Juefei-Xu, F., Chen, H., Xue, M., Li, B., Liu, Y., Zhao, J.,
  Yin, J., and See, S. (2018).
\newblock Coverage-guided fuzzing for deep neural networks.
\newblock {\em arXiv preprint arXiv:1809.01266}.

\bibitem[Xu et~al., 2018]{XEQ2017}
Xu, W., Evans, D., and Qi, Y. (2018).
\newblock Feature squeezing: Detecting adversarial examples in deep neural
  networks.
\newblock In {\em Network and Distributed System Security Symposium (NDSS)}.

\bibitem[Yosinski et~al., 2015]{YCNFL2015}
Yosinski, J., Clune, J., Nguyen, A.~M., Fuchs, T.~J., and Lipson, H. (2015).
\newblock Understanding neural networks through deep visualization.
\newblock In {\em International Conference on Machine Learning (ICML) Deep
  Learning Workshop}.

\bibitem[Zeiler and Fergus, 2014]{ZF2013}
Zeiler, M.~D. and Fergus, R. (2014).
\newblock Visualizing and understanding convolutional networks.
\newblock In {\em European Conference on Computer Vision (ECCV)}.

\bibitem[Zhang et~al., 2018a]{zhang2018crown}
Zhang, H., Weng, T.-W., Chen, P.-Y., Hsieh, C.-J., and Daniel, L. (2018a).
\newblock Efficient neural network robustness certification with general
  activation functions.
\newblock In {\em Advances in neural information processing systems}, pages
  4939--4948.

\bibitem[Zhang et~al., 2019]{zhang2018recurjac}
Zhang, H., Zhang, P., and Hsieh, C.-J. (2019).
\newblock Recurjac: An efficient recursive algorithm for bounding jacobian
  matrix of neural networks and its applications.
\newblock In {\em AAAI Conference on Artificial Intelligence (AAAI), arXiv
  preprint arXiv:1810.11783}.

\bibitem[Zhang et~al., 2018b]{deeproad}
Zhang, M., Zhang, Y., Zhang, L., Liu, C., and Khurshid, S. (2018b).
\newblock {DeepRoad}: {GAN}-based metamorphic autonomous driving system
  testing.
\newblock In {\em Automated Software Engineering (ASE), 33rd IEEE/ACM
  International Conference on}.

\bibitem[{Zhao} et~al., 2020]{2020arXiv200305311Z}
{Zhao}, X., {Banks}, A., {Sharp}, J., {Robu}, V., {Flynn}, D., {Fisher}, M.,
  and {Huang}, X. (2020).
\newblock {A Safety Framework for Critical Systems Utilising Deep Neural
  Networks}.
\newblock {\em arXiv e-prints}, page arXiv:2003.05311.

\bibitem[Zhu et~al., 1997]{ZHM1997}
Zhu, H., Hall, P.~A., and May, J.~H. (1997).
\newblock Software unit test coverage and adequacy.
\newblock {\em ACM Computing Surveys}, 29(4):366--427.

\bibitem[Zilke et~al., 2016]{DeepRED}
Zilke, J.~R., Loza~Menc{\'i}a, E., and Janssen, F. (2016).
\newblock Deepred -- rule extraction from deep neural networks.
\newblock In Calders, T., Ceci, M., and Malerba, D., editors, {\em Discovery
  Science}, pages 457--473, Cham. Springer International Publishing.

\bibitem[Zintgraf et~al., 2017]{PDA}
Zintgraf, L.~M., Cohen, T.~S., Adel, T., and Welling, M. (2017).
\newblock Visualizing deep neural network decisions: Prediction difference
  analysis.
\newblock In {\em International Conference on Machine Learning (ICML)}.

\end{thebibliography}

\end{document}